\documentclass[10pt]{article}
\usepackage{amsmath}
\usepackage{amsfonts}
\usepackage{amssymb}
\usepackage{mathrsfs}                  
\usepackage{amsthm}
\usepackage{enumitem}
\usepackage{amscd}
\usepackage{xcolor}
\usepackage[hyperfootnotes=false]{hyperref}
\usepackage{mathtools}
\usepackage{geometry}
\usepackage{bbm}
\usepackage{bm}
\usepackage{bbm}

\numberwithin{equation}{section}

\theoremstyle{plain}
\newtheorem{theorem}{Theorem}[section]
\newtheorem{proposition}[theorem]{Proposition}

\newtheorem{corollary}[theorem]{Corollary}

\theoremstyle{definition}
\newtheorem{definition}[theorem]{Definition}

\theoremstyle{remark}
\newtheorem{remark}[theorem]{Remark}

\newcommand{\E}{\mathbb{E}}
\newcommand{\R}{\mathbb{R}}

\newcommand{\Z}{\mathbb{Z}}
\newcommand{\N}{\mathbb{N}}

\renewcommand{\P}{\mathbb{P}}

\newcommand{\bfi}{\bfseries\itshape}
\newcommand{\vertiii}[1]{{\left\vert\kern-0.25ex\left\vert\kern-0.25ex\left\vert #1 
    \right\vert\kern-0.25ex\right\vert\kern-0.25ex\right\vert}}

\begin{document}

\title{{\bf Infinite-dimensional reservoir computing}}
\author{Lukas Gonon$^{1}$, Lyudmila Grigoryeva$^{2, 3}$, and Juan-Pablo Ortega$^{4}$}
\date{}
\maketitle

\begin{abstract}
Reservoir computing approximation and generalization bounds are proved for a new concept class of input/output systems that extends the so-called generalized Barron functionals to a dynamic context. This new class is characterized by the readouts with a certain integral representation built on infinite-dimensional state-space systems. It is shown that this class is very rich and possesses useful features and universal approximation properties. The reservoir architectures used for the approximation and estimation of elements in the new class are randomly generated echo state networks with either linear or ReLU activation functions. Their readouts are built using randomly generated neural networks in which only the output layer is trained (extreme learning machines or random feature neural networks). The results in the paper yield a fully implementable recurrent neural network-based learning algorithm with provable convergence guarantees that do not suffer from the curse of dimensionality.
\end{abstract}

\bigskip


\makeatletter
\addtocounter{footnote}{1} \footnotetext{%
Imperial College.  Department of Mathematics. London. United Kingdom.
{\texttt{l.gonon@imperial.ac.uk}}}
\addtocounter{footnote}{1} \footnotetext{%
Universit\"at Sankt Gallen. Faculty of Mathematics and Statistics. Sankt Gallen. Switzerland.  {\texttt{lyudmila.grigoryeva@unisg.ch} }}
\addtocounter{footnote}{1} \footnotetext{%
University of Warwick. Department of Statistics. United Kingdom.
{\texttt{lyudmila.grigoryeva@warwick.ac.uk} }}
\addtocounter{footnote}{1} \footnotetext{%
Nanyang Technological University. Division of Mathematical Sciences. Singapore.   {\texttt{Juan-Pablo.Ortega@ntu.edu.sg} } }
\makeatother

\tableofcontents

\newpage

\section{Introduction}

{\bfi  Reservoir computing} (RC) \cite{jaeger2001, maass1, Jaeger04, maass2} and in particular {\bfi   echo state networks} (ESNs)  \cite{Matthews:thesis, Matthews1993, Jaeger04} have gained much popularity in recent years due to their excellent performance in the forecasting of dynamical systems \cite{GHLO2014, Jaeger04, pathak:chaos, Pathak:PRL, Ott2018, wikner2021using, arcomano2022hybrid} and due to the ease of their implementation. RC aims at approximating nonlinear input/output systems using randomly generated state-space systems (called {\bfi   reservoirs}) in which only a linear readout is estimated. It has been theoretically established that this is indeed possible in a variety of deterministic and stochastic contexts \cite{RC6, RC7, RC8, RC20, RC12} in which RC systems have been shown to have universal approximation properties.  

In this paper, we focus on deriving error bounds for a variant of the architectures that we just cited and consider as approximants randomly generated linear systems with readouts given by randomly generated neural networks in which only the output layer is trained. Thus, from a learning perspective, we combine linear echo state networks and what is referred to in the literature as {\bfi random features} \cite{Rahimi2007} /{\bfi extreme learning machines} (ELMs) \cite{Huang2006}. We develop explicit and readily computable approximation and estimation bounds for a newly introduced concept class whose elements we refer to as {\bfi recurrent (generalized) Barron functionals} since they can be viewed as a dynamical analog of the {\bfi (generalized) Barron functions} introduced in \cite{barron1992neural, Barron1993} and extended later in \cite{EWojtowytsch2020, EMaWWu2020, EMaWu2019}. The main novelty in this concept class with respect to others already available in the literature, like the {\bfi fading memory class} \cite{Boyd1985} in deterministic setups or  $L ^p $ functionals in the stochastic case, is that it consists of elements that admit an {\it explicit infinite-dimensional state-space representation}, which makes them analytically tractable. As we shall see later on, many interesting families of input/output systems belong to this class that, additionally, {\it is universal in the $L ^p $-sense}.

From an approximation-theoretical point of view, the universality properties of linear systems with readouts belonging to dense families have been known for a long time both in deterministic \cite{Boyd1985, RC6} and in stochastic \cite{RC8} setups. Their corresponding dynamical and memory properties have been extensively studied (see, for instance, \cite{Hermans2010, linearESN, tino:symmetric, Tino2019, RC15, li2022approximation} and references therein). Our contribution in this paper can hence be considered as an {\it extension of those works to the recurrent Barron class}. 

All the dynamic learning works cited so far use exclusively finite-dimensional state spaces. Hence, one of the main novelties of our contribution is the infinite-dimensional component in the concept class that we propose. It is worth mentioning that, even though in static setups, there exist numerous neural functional approximation results (see, for instance, \cite{chen1995approximation, stinchcombe1999neural, anastasis:neurips, benth2022neural, cuchiero2022universal, neufeld2022chaotic}, in addition to the works on Barron functions cited above), the use of infinite-dimensional state-space systems has not been much exploited, and it is only very recently that it is being seriously developed. See \cite{hermans:rkhs, Bouvrie2017, bouvrie2017kernel, KO:19, li2020fourier, kovachki2021neural, acciaio2022metric, galimberti2022designing, hu2022neural, salvi2022neural} for a few examples. Dedicated hardware realizations of RC systems using quantum systems are a potential application of these extensions for which, in the absence of adequate tools, most of the theoretical developments have been carried out so far in exclusively finite-dimensional setups \cite{chen2019learning, chen2020temporal, tran2020higher, tran2021learning, chen2022nonlinear, QRC1}.

In order to introduce our contributions in more detail, we start by recalling that an echo state network (ESN) is an input/output system defined by the two equations: 
\begin{align}
{\bf x}_t &= {\sigma}(A{\bf x}_{t-1} + C {\bf z}_t + \bm{\zeta}),\label{prescription esn}\\
{\bf  y}_t &= W {\bf x}_t, 
\label{readout}
\end{align}
for $t \in \Z_-$, where $d, m, N \in \N$, ${\bf z} \in (\R^d)^{\Z_-}$, ${\bf x}_t \in \R^N$, the matrix $W \in \mathbb{M}_{m, N}$ is trainable, $\sigma$ denotes the componentwise application of a given activation function $\sigma \colon \R \to \R$, and $A\in \mathbb{M}_{N}$, $C\in \mathbb{M}_{N, d}$, and $\bm{\zeta} \in \R^N$ are randomly generated.
If for each $ {\bf z} \in (\R^d)^{\Z_-}$ there exists a unique solution ${\bf x} = ({\bf x}_t)_{t \in \Z_-} \in (\R^N)^{\Z_-}$ to \eqref{prescription esn}, then \eqref{prescription esn}-\eqref{readout} define a mapping $H_W \colon (\R^d)^{\Z_-} \to   \R^m$ via $H_W({\bf z})={\bf y}_0$, with ${\bf y}\in  (\R^m)^{\Z_-}$ given by \eqref{readout} for {\bf x} the unique solution to \eqref{prescription esn} associated with ${\bf z} \in (\R^d)^{\Z_-}$. Given a (typically unknown) functional $H \colon (\R^d)^{\Z_-} \to   \R^m$ to be learned, the readout $W$ is then trained so that $H_W$ is a good approximation of $H$.

For echo state networks \eqref{prescription esn}--\eqref{readout}, approximation bounds were given in \cite{RC12} for maps $H$ which have the property that their restrictions to input sequences of finite length $T$ lie in a certain Sobolev space, for each $T \in \N$, and with weights $A$, $C$, $\bm{\zeta}$ in \eqref{prescription esn} sampled from a generic distribution with a certain structure. Here we consider a novel architecture for \eqref{prescription esn}--\eqref{readout}, where, instead of applying a linear function in \eqref{readout} we apply a random feedforward  neural network, that is, \ \eqref{readout} is replaced by 
\begin{equation}\label{eq:alternativereadout}
{\bf y}_t =
\sum_{i=1}^N W_i \sigma({\bf a}^{(i)} \cdot  \mathbf{x}_{t-1} + {\bf c}^{(i)} \cdot {\bf z}_{t} + b_i)
\end{equation}
for $t \in \Z_-$ and  with randomly generated coefficients ${\bf a}^{(i)}$, ${\bf c}^{(i)} $, $b_i$ valued in $\R^N$, $\R^d$, and $\R$, respectively. In most cases, the activation function $\sigma $ in \eqref{prescription esn} will be just the identity or, eventually, a rectified linear unit.

In order to derive learning error bounds for this architecture, we shall proceed in several steps. First, we examine the new concept class $\mathcal{C}$ of {\bfi recurrent (generalized) Barron functionals} consisting of reservoir functionals that admit a certain integral representation.
This class $\mathcal{C}$ turns out to possess very useful properties: first, we show that a large class of functionals  $H \colon (\R^d)^{\Z_-} \to \R^m$, including -- but not restricted to -- functionals with ``sufficient smoothness'' are in $\mathcal{C}$ and, under mild conditions, $\mathcal{C}$ is dense in the $L^2$-sense. We then examine the approximation of elements in $\mathcal{C}$ by reservoir computing systems and derive approximation error bounds. This shows that a large class of functionals can be approximated by recurrent neural networks whose hidden weights are randomly sampled from generic distributions, and explicit approximation rates can be derived for them.

The second step consists in obtaining generalization error bounds that match the parameter restrictions emerging from the approximation results for these systems. The key challenge here is that the observational data is non-IID and so classical statistical learning techniques \cite{Boucheron2013} can not be employed directly, see \cite{RC10}.  By combining these bounds, we then obtain learning error bounds for such random recurrent neural networks for learning a general class of maps $\mathcal{C}$. As a by-product, we obtain new universality results for reservoir systems, see  \cite{RC6, RC7, RC8, RC12, RC20},  with generically randomly generated coefficients.
It is worth emphasizing that the construction that we describe in this paper yields a fully implementable recurrent neural network-based learning algorithm with provable convergence guarantees not suffering from the curse of dimensionality.

\section{A dynamic analog of the generalized Barron functions}

In this section, we introduce the class $\mathcal{C}$ of recurrent (generalized) Barron functionals, which constitute the concept class that we study and approximate in this paper.  We prove elementary properties of $\mathcal{C}$ and identify important constituents in it. In particular, we show that $\mathcal{C}$ is a vector space that is dense in the set of square-integrable functionals and contains many important classes of maps $H \colon (\R^d)^{\Z_-} \to \R$ such as linear or ``sufficiently smooth'' maps. 

\subsection{Notation}
\label{Notation}
Let $D_d \subset \mathbb{R}^d$, $p \in [1,\infty]$, and let $q$ be the H\"older conjugate of $p$. Recall the notation $\ell^p=\ell^p( \mathbb{R})$ for the space of sequences $(x_n)_{n \in \N} \in \R^\N$ with $\sum_{i \in \N} |x_i|^p < \infty$, in the case $p<\infty$, and $\sup_{i \in \N} |x_i| < \infty$, in the case $p=\infty$. 
The symbol $\ell^p_-=\ell^p_-( \mathbb{R})$ denotes the space of sequences $(x_n)_{n \in \mathbb{Z}_{-}} \in \R^{\mathbb{Z}_{-}}$ that satisfy the same summability conditions as those in $\ell^p $, with $\mathbb{Z}_{-}  $ the negative integers including zero. 
We let $\ell^\infty_-(\ell^q)$ denote the set of sequences $(\mathbf{x}_t)_{t \in \Z_-} \in (\ell^q)^{\Z_-}$ with the property that $\sup_{t \in \Z_-} \|\mathbf{x}_t\|_q < \infty$.
For $\mathbf{x} \in \ell^p$ and ${\bf y} \in \ell^q$ we write $\mathbf{x} \cdot {\bf y} = \sum_{i \in \N} x_i y_i$. 
Let $\sigma_1,\sigma_2 \colon \R \to \R$ denote two functions, called activation functions. For both activation functions we denote using the same symbol their componentwise application to sequences or vectors. Furthermore, we assume that $\sigma_2$ is Lipschitz-continuous with constant $L_{\sigma_2}$ and $\sigma_1(0)=0$. Given $N \in \mathbb{N} $, we write $\mathcal{X} = \R \times \ell^p \times \R^d \times \R$ and $\mathcal{X}^N = \R \times \R^N \times \R^d \times \R$. We will consider inputs in $\mathcal{I}_d \subset (D_d)^{\Z_-}$. 

\subsection{Definition and properties of recurrent Barron functionals}
\label{subsec:Definition and properties of recurrent Barron functionals}
In this section we define the class of recurrent generalized Barron functionals and study some properties of these functionals.
\begin{definition} A functional $H \colon \mathcal{I}_d \to \R$ is said to be a {\bfi  recurrent (generalized) Barron functional}, if there exist a (Borel) probability measure $\mu$ on $\mathcal{X}$, ${\bf B} \in \ell^q$ and linear maps $A \colon \ell^q \to \ell^q$, $C \colon \R^d \to \ell^q$ such that for each ${\bf z} \in  \mathcal{I}_d$ the system 
\begin{equation}
\label{eq:system}
 \mathbf{x}_t = \sigma_1(A \mathbf{x}_{t-1} + C {\bf z}_t + {\bf B}), \quad t \in \Z_-,
 \end{equation}
   admits a unique solution $(\mathbf{x}_t)_{t \in \Z_-} \in \ell^\infty_-(\ell^q)$, $\mu$ satisfies that $$I_{\mu,p}:=\int_{\mathcal{X}} |w| (\|{\bf a}\|_p + \|{\bf c}\|  +  |b|) \mu(dw,d{\bf a},d {\bf c},d b) < \infty$$ and, writing  $\mathbf{x}_t = \mathbf{x}_t({\bf z})$,
\begin{equation} 
\label{eq:Hrepres} H({\bf z}) = \int_{\mathcal{X}} w \sigma_2({\bf a} \cdot \mathbf{x}_{-1}({\bf z}) + {\bf c} \cdot {\bf z}_0 +  b) \mu(dw,d{\bf a},d {\bf c},d b), \quad  {\bf z} \in  \mathcal{I}_d.
\end{equation}
We denote by $\mathcal{C}$ the {\bfi class} of all {\bfi recurrent generalized Barron functionals} or just {\bfi recurrent  Barron functionals}. 
\end{definition}

\medskip

\noindent Note that the readout \eqref{eq:Hrepres} is built in such a way that the output is constructed out of $\mathbf{x} _{-1}  $  and ${\bf z}_0 $ instead of exclusively $\mathbf{x} _0 $, like it is customary in reservoir computing (see \eqref{readout}). The reason for this choice is that it will allow us later on in Section \ref{Special case: static situation} to recover the static situation as a particular case of the recurrent setup in a straightforward manner.


\medskip

\noindent {\bf The unique solution property (USP).} The existence and uniqueness of solutions property of state equations of the type \eqref{eq:system}, which is part of the definition of the recurrent Barron functionals, is a well-studied problem. This property is  referred in the literature as the {\bfi  echo state property} (ESP) \cite{jaeger2001, Yildiz2012, Manjunath:Jaeger} or the {\bfi  unique solution property} (USP) \cite{manjunath:prsl1, manjunath2022embedding} and it has been much tackled in deterministic (see references above and \cite{RC7}, for instance), and stochastic setups \cite{RC27}. The following proposition is an infinite-dimensional adaptation of a commonly used sufficient condition for the USP to hold in the case of  \eqref{eq:system}, as well as a full characterization of it when the activation function $\sigma_1  $ is the identity, and hence \eqref{eq:system} is a linear system. In both cases, the inputs are assumed to be bounded; possible generalizations to the unbounded case can be found in \cite{RC9}.

\begin{proposition}
\label{usp infinite-dimensional}
Consider the state equation given by  \eqref{eq:system} and the two following cases:
\begin{description}
\item [(i)] Suppose that the inputs are defined in $\mathcal{I}_d \subset (D_d)^{\Z_-}$, with $D _d $ a compact subset of ${\Bbb R}^d$. Furthermore, assume that $\sigma_1$ is Lipschitz-continuous with constant $L_{\sigma_1}$ and that $L_{\sigma_1}\vertiii{A} <1$ with $\vertiii{A} $ the operator norm of the linear map $A \colon \ell^q \to \ell^q$. Then \eqref{eq:system} has the USP, i.e., for each ${\bf z} \in  \mathcal{I}_d$ there exists a unique $(\mathbf{x}_t)_{t \in \Z_-} \in \ell^\infty_-(\ell^q)$ satisfying \eqref{eq:system}.
\item [(ii)]  Suppose that $\sigma_1  $ is the identity and that $A \colon \ell^q \to \ell^q$, $C \colon \R^d \to \ell^q$ are bounded linear operators. 
If the spectral radius $\rho(A) $ of the operator $A$ is such that $\rho(A) <1$, then the linear state equation \eqref{eq:system} has the USP with respect to inputs in $\ell^{\infty}_-({\Bbb R}^d) $. In that case the unique solution of \eqref{eq:system} is given by
\begin{equation}
\label{eq:auxEq10}
\mathbf{x}_{t}({\bf z}) = \sum_{k=0}^\infty A^k (C {\bf z}_{t-k} + {\bf B}), \quad \mbox{$t \in \Bbb Z$}. 
\end{equation}
\end{description}
\end{proposition}

\begin{proof}
The proof of part {\bf (i)} is a straightforward generalization of \cite[Theorem 3.1(ii)]{RC7} together with the observation that the hypothesis $L_{\sigma_1}\vertiii{A} <1$ implies that the state equation is a contraction on the state entry. Part {\bf (ii)} can be obtained by mimicking the proof of \cite[Proposition 4.2(i)]{RC16}; a key element in that proof is the use of Gelfand's formula for the spectral radius, which holds for operators between infinite-dimensional Banach spaces \cite[Theorem 4, page 195]{lax:functional:analysis}.
\end{proof}

\medskip

\noindent {\bf The concept class $\mathcal{C}$ of recurrent Barron functionals is a vector space. }
\begin{proposition}
\label{lem:linearspace}
Suppose $H_1,H_2 \in \mathcal{C}$ and $\lambda_1,\lambda_2 \in \R$. Then $\lambda_1 H_1 + \lambda_2 H_2 \in \mathcal{C}$. 
\end{proposition}
\begin{proof}
Without loss of generality, we may assume $\lambda_i \neq 0$ for $i=1,2$. 
For $i=1,2$ let $\mu^{(i)}$ be (Borel) probability measures on $\mathcal{X}$, let ${\bf B}^{(i)} \in \ell^q$, let $A^{(i)} \colon \ell^q \to \ell^q$, $C^{(i)} \colon \R^d \to \ell^q$ be linear maps such that for each ${\bf z} \in  \mathcal{I}_d$ the system 
\begin{equation}\label{eq:systemi}
\mathbf{x}_t^{(i)} = \sigma_1(A^{(i)} \mathbf{x}_{t-1}^{(i)} + C^{(i)} {\bf z}_t + {\bf B}^{(i)}), \quad t \in \Z_-,
\end{equation}
admits a unique solution $(\mathbf{x}_t^{(i)})_{t \in \Z_-} \in \ell^\infty_-(\ell^q)$, $\mu^{(i)}$ satisfies $I_{\mu^{(i)},p}< \infty$, and
\begin{equation} \label{eq:Hrepresi} H_i({\bf z}) = \int_{\mathcal{X}} w \sigma_2({\bf a} \cdot \mathbf{x}_{-1}^{(i)}({\bf z}) + {\bf c} \cdot {\bf z}_0 +  b) \mu^{(i)}(dw,d{\bf a},d {\bf c},d b), \quad  {\bf z} \in  \mathcal{I}_d.
\end{equation}
Define $\pi_1 \colon \R^\N \to \R^\N$ by $\pi_1((y_n)_{n \in \N}) = (y_{\frac{n+1}{2}} \mathbbm{1}_{\N}(\frac{n+1}{2}) )_{n \in \N} $ and $\pi_2 \colon \R^\N  \to \R^\N $ by $\pi_2((y_n)_{n \in \N}) = (y_{\frac{n}{2}} \mathbbm{1}_{\N}(\frac{n}{2}))_{n \in \N}$ and note that $\pi_1, \pi_2$ are linear, they map both $\ell^p$ and $\ell^q$ into themselves, $\pi_1+\pi_2= \mathbb{I} $, and $\pi _i \circ \sigma _1= \sigma _1\circ \pi _i $,  $i=1,2 $. Define ${\bf B} \in \ell^q$ and linear maps $A \colon \ell^q \to \ell^q$, $C \colon \R^d \to \ell^q$ by setting  
$A {\bf x} = \pi_1(A^{(1)} {\bf x}) + \pi_2(A^{(2)} {\bf x})$, ${\bf B} = \pi_1({\bf B}^{(1)}) + \pi_2({\bf B}^{(2)})$ and $C {\bf z} = \pi_1(C^{(1)} {\bf z}) + \pi_2(C^{(2)} {\bf z})$.

Suppose now that $(\mathbf{x}_t)_{t \in \Z_-} \in \ell^\infty_-(\ell^q)$ is a solution to \eqref{eq:system} for the input ${\bf z} \in  \mathcal{I}_d$ and the parameters $A,{\bf B},C$ that we just defined. For each $t \in \Z_-$ denote by $\mathbf{x}_t^{(1)} = (\mathbf{x}_{t,2n-1})_{n \in \N}$ the odd components of $\mathbf{x} _t  $ and by $\mathbf{x}_t^{(2)} = (\mathbf{x}_{t,2n})_{n \in \N}$ the even components. Then $\mathbf{x}_t^{(i)} \in \ell^q$ and by construction of $A,{\bf B},C$, it follows that $(\mathbf{x}_t^{(i)})_{t \in \Z_-}$ is a solution of  \eqref{eq:systemi} for $i=1,2$. By the uniqueness of the solutions of \eqref{eq:systemi}, it follows that there is at most one solution of \eqref{eq:system}.  On the other hand, if we now denote by $(\mathbf{x}^{(i)}_t)_{t \in \Z_-} \in \ell^\infty_-(\ell^q)$, $i=1,2$, the unique solution to \eqref{eq:systemi}  for the input ${\bf z} \in  \mathcal{I}_d$, then setting $\mathbf{x}_t = \pi_1(\mathbf{x}_t^{(1)}) + \pi_2(\mathbf{x}_t^{(2)})$ defines a sequence $(\mathbf{x}_t)_{t \in \Z_-} \in \ell^\infty_-(\ell^q)$ which is a solution to \eqref{eq:system}, again by the construction of $A,{\bf B},C$.

Next, define $\phi_i \colon \mathcal{X} \to \mathcal{X}$ as
$\phi_i(w,{\bf a},{\bf c},b) = (w(\lambda_1+\lambda_2),\pi_i({\bf a}),{\bf c},b)$ for $i=1,2$ and set
$\mu = \frac{\lambda_1}{\lambda_1+\lambda_2} (\mu^{(1)} \circ \phi_1^{-1}) + \frac{\lambda_2}{\lambda_1+\lambda_2} (\mu^{(2)} \circ \phi_2^{-1})$, where $\mu^{(i)} \circ \phi_i^{-1}$ denotes the pushforward of $\mu^{(i)}$ under $\phi_i$. Then the integral transformation theorem shows that 
\[
 \int_{\mathcal{X}} |w| (\|{\bf a}\|_p + \|{\bf c}\|  +  |b|) \mu(dw,d{\bf a},d {\bf c},d b) = \sum_{i=1}^2 \lambda_i \int_{\mathcal{X}} |w| (\|\pi_i({\bf a})\|_p + \|{\bf c}\|  +  |b|) \mu^{(i)}(dw,d{\bf a},d {\bf c},d b) < \infty
\]
and, by \eqref{eq:Hrepresi}, for all ${\bf z} \in \mathcal{I}_d$, 
\begin{align*} 
\int_{\mathcal{X}} w \sigma_2({\bf a} \cdot \mathbf{x}_{-1} + {\bf c} \cdot {\bf z}_0 +  b) \mu(dw,d{\bf a},d {\bf c},d b) & = \sum_{i=1}^2 \lambda_i \int_{\mathcal{X}} w \sigma_2(\pi_i({\bf a}) \cdot \mathbf{x}_{-1} + {\bf c} \cdot {\bf z}_0 +  b) \mu^{(i)}(dw,d{\bf a},d {\bf c},d b)
\\
& = \sum_{i=1}^2 \lambda_i \int_{\mathcal{X}} w \sigma_2({\bf a} \cdot \mathbf{x}_{-1}^{(i)} + {\bf c} \cdot {\bf z}_0 +  b) \mu^{(i)}(dw,d{\bf a},d {\bf c},d b)
\\
& = \lambda_1 H_1({\bf z}) + \lambda_2 H_2({\bf z}). 
\end{align*}
Hence, $\lambda_1 H_1 + \lambda_2 H_2 \in \mathcal{C}$, as claimed. 
\end{proof}

\noindent {\bf Finite-dimensional recurrent Barron functionals.} For $N \in \N$, denote by $\mathcal{C}_N$ the set of functionals $H^{{N}} \colon \mathcal{I}_d \subset (D_d)^{\Z_-} \to \R$ with the following property: there exist a (Borel) probability measure $\mu^N$ on $\mathcal{X}^N$, ${A} \in \mathbb{M}_{N , N}$, ${\bf B} \in \R^N$, ${C}  \in \mathbb{M}_{N , d}$, such that for each ${\bf z} \in  \mathcal{I}_d$ the system 
\begin{equation}
\label{eq:systemN}
\mathbf{x}_t^N = \sigma_1({A} \mathbf{x}_{t-1}^N + {C} {\bf z}_t + {\bf B}), \quad t \in \Z_-,
\end{equation}
admits a unique solution $(\mathbf{x}_t^N)_{t \in \Z_-} \in \ell^\infty_-(\R^N)$, $\mu^N$ satisfies $$\int_{\mathcal{X}^N} |w| (\|{\bf a}\| + \|{\bf c}\|  +  |b|) \mu^N(dw,d{\bf a},d {\bf c},d b) < \infty$$ and, writing  $\mathbf{x}_t^N = \mathbf{x}_t^N({\bf z})$, 
\begin{equation} 
\label{eq:HrepresN} 
H^N({\bf z}) = \int_{\mathcal{X}^N} w \sigma_2({\bf a} \cdot \mathbf{x}_{-1}^N({\bf z}) + {\bf c} \cdot {\bf z}_0 +  b) \mu^N(dw,d{\bf a},d {\bf c},d b), \quad  {\bf z} \in  \mathcal{I}_d.
\end{equation}
We shall refer to the elements in $\mathcal{C}_N$ as {\bfi finite-dimensional recurrent Barron functionals}. In the following proposition, we show that   the set of finite-dimensional recurrent Barron functionals can be naturally seen as a subspace of $\mathcal{C} $. In the statement, we use the notion and properties of system morphisms that are recalled, for instance, in \cite{RC16}.

\medskip

\begin{proposition} 
\label{lem:finite} 
For $N \in \N$ arbitrary, let $\iota \colon \R^N \to \ell^q$ be the natural injection $(x _1, \ldots, x _N)^{\top} \longmapsto (x _1, \ldots, x _N, 0, \ldots)$. Given an element $H ^N \in \mathcal{C}_N $, there exists a system of type \eqref{eq:system}-\eqref{eq:Hrepres} that realizes $H ^N$ and for which the map $\iota $ is a system morphism. This allows us to conclude that 
\begin{equation}
\label{inclusionfininf}
\mathcal{C}_N \subset\mathcal{C}, 
\end{equation}
that is, every finite-dimensional recurrent Barron functional admits a realization as a recurrent Barron functional.
\end{proposition}

\begin{proof} 
Let $H^N \in \mathcal{C}_N$ be such that \eqref{eq:HrepresN} holds with $\mathbf{x}^N\in \ell^\infty_-(\R^N)$ satisfying \eqref{eq:systemN}.
First, define $\bar{A} \colon \ell^q \to \ell^q$, $\bar{{\bf B}} \in \ell^q$, $\bar{C} \colon \R^d \to \ell^q$ by setting  $(\bar{A} \mathbf{x})_{i} = \mathbbm{1}_{\{i\leq N\}}\sum_{j=1}^N {A}_{i j} x_j$, $\bar{B}_i = \mathbbm{1}_{\{i\leq N\}} {B}_i$, $(\bar{C} {\bf z})_{i} = \mathbbm{1}_{\{i\leq N\}}\sum_{j=1}^d { C}_{i j} z_j$ for $\mathbf{x} \in \ell^q$, ${\bf z} \in \R^d$, $i \in \N$. Furthermore,  if $\iota$ is the canonical injection, we denote by the same symbol the injection of $\mathcal{X} ^N$ into ${\cal X}$ and by $\mu = \mu^N \circ \iota^{-1}$ the pushforward measure of $\mu^N$ under $\iota$. 

Consider now the system \eqref{eq:systemN}--\eqref{eq:HrepresN} and the one given by  \eqref{eq:system} and \eqref{eq:Hrepres} with the parameters $\bar{A},\bar{{\bf B}},\bar{C}$ that we just defined, as well as the readout given by the measure $\mu = \mu^N \circ \iota^{-1}$. We shall prove that the map $\iota \colon \R^N \to \ell^q$ is a system morphism between these systems. This requires showing that   $\iota  $ is system equivariant and readout invariant (see \cite{RC16} for this terminology).

First, we notice that $\iota  $ is system equivariant because
\begin{equation*}
\iota\left(\sigma_1({A} \mathbf{x}^N + {C} {\bf z} + {\bf B})\right) = \sigma_1(\iota({A} \mathbf{x}^N + {C} {\bf z} + {\bf B})) = \sigma_1( \bar{A} \iota(\mathbf{x}^N) +  \bar{C} {\bf z} + \bar{{\bf B}}), \  \mbox{for any $\mathbf{x}^N \in \mathbb{R}^N$, ${\bf z} \in {\Bbb R}^d $,}
\end{equation*}
as required. As a consequence of this fact (see \cite[Proposition 2.2]{RC16}) if $ (\mathbf{x}_t ^N) _{t \in \mathbb{Z}_{-}} \in \ell^\infty_-(\R^N)$ is a solution of the system determined by \eqref{eq:systemN} and \eqref{eq:HrepresN} for ${\bf z} \in {\cal I}_d $, then so is $ (\mathbf{x}_t) _{t \in \mathbb{Z}_{-}} \in \ell^\infty_-(\ell^q)$ with $\mathbf{x}_t=\iota(\mathbf{x}_t^N) $ for the one given by \eqref{eq:system} and \eqref{eq:Hrepres} with the parameters $\bar{A},\bar{{\bf B}},\bar{C} $. This solution is unique because if $(\overline{\mathbf{x}}_t)_{t \in \Z_-} \in \ell^\infty_-(\ell^q)$ is another solution for it and we denote by $\overline{\mathbf{x}}_t^N \in \R^N$ its first $N$ components then from \eqref{eq:system} we obtain for the components $i=1,\ldots,N$ that $\overline{{x}}_{t,i}^N = \sigma_1((\bar{A} \overline{\mathbf{x}}_{t-1})_i + (\bar{C} {\bf z}_t)_i + \bar{B}_i) = \sigma_1(({A} \overline{\mathbf{x}}_{t-1}^N)_i + ({C} {\bf z}_t)_i + {B}_i)$, which shows that $\overline{\mathbf{x}}^N$ is a solution to \eqref{eq:systemN} and thus, by uniqueness, $\overline{\mathbf{x}}^N = \mathbf{x}^N$ necessarily. For components $i>N$ we get from \eqref{eq:system} that $\overline{x}_{t,i} = \sigma_1(0) = 0$. 
Altogether, this proves that for each ${\bf z} \in  \mathcal{I}_d$ the system given by \eqref{eq:system} with the parameters $\bar{A},\bar{{\bf B}},\bar{C}$ has a unique solution $\mathbf{x}$ and the first $N$ entries of each term  $\mathbf{x}_t$  are given by $ \mathbf{x}_t^N$. 

Notice now that the change-of-variables theorem and the equivalence of norms on $\R^N$ yields that \[\int_{\mathcal{X}} |w| (\|{\bf a}\|_p + \|{\bf c}\|  +  |b|) \mu(dw,d{\bf a},d {\bf c},d b) = \int_{\mathcal{X}^N} |w| (\|\iota({\bf a})\|_p + \|{\bf c}\|  +  |b|) \mu^N(dw,d{\bf a},d {\bf c},d b) < \infty,
\]
which proves that $H$ in \eqref{eq:Hrepres} defines an element in $\mathcal{C}$. Finally, we show that the map $\iota $ is readout invariant using that the first $N$ components of each solution $\mathbf{x}_t$ are given by $ \mathbf{x}_t^N$ and applying again the change-of-variables theorem:
\begin{multline*}  
H({\bf z}) = \int_{\mathcal{X}} w \sigma_2({\bf a} \cdot \mathbf{x}_{-1}({\bf z}) + {\bf c} \cdot {\bf z}_0 +  b) \mu(dw,d{\bf a},d {\bf c},d b)
= \int_{\mathcal{X}^N} w \sigma_2(\iota({\bf a}) \cdot \mathbf{x}_{-1}({\bf z}) + {\bf c} \cdot {\bf z}_0 +  b) \mu^N(dw,d{\bf a},d {\bf c},d b) \\
= \int_{\mathcal{X}^N} w \sigma_2({\bf a} \cdot \mathbf{x}_{-1}^N({\bf z}) + {\bf c} \cdot {\bf z}_0 +  b) \mu^N(dw,d{\bf a},d {\bf c},d b)=H ^N({\bf z})
\end{multline*}
for ${\bf z} \in  \mathcal{I}_d$, as required. 
In particular, $H^N$ and $H$ are in fact the same functional, which proves the inclusion \eqref{inclusionfininf}.
\end{proof}

\subsection{Examples of recurrent Barron functionals}

We now show that large classes of input/output systems naturally belong to the recurrent Barron class.

\medskip

\noindent {\bf Finite-memory functionals.} The following proposition shows that finite-memory causal and time-invariant systems that have certain regularity properties can be written as finite-dimensional recurrent Barron functionals with a linear state equation. 

\begin{proposition} 
Assume $\sigma_1(x)=x$, $\sigma_2(x)=\max(x,0)$, $D_d$ is bounded,
let $T \in \N$, and suppose that $h \colon \R^{Td} \to \R$ is in the Sobolev space $H^{s}(\R^{Td})$ for $s > \frac{Td}{2}+2$. Then, the functional $H({\bf z}) = h({\bf z}_0,\ldots,{\bf z}_{-T+1})$, ${\bf z} \in \mathcal{I}_d$, is a finite-dimensional recurrent Barron functional.  
\end{proposition}

\begin{proof}
Choose $N = d T$, ${\bf B}=\mathbf{0}$, 
\begin{equation} \label{eq:shiftMatrix1}
{A}=  \left(
\begin{array}{cc}
\mathbf{0}_{d,d(T-1)}&\mathbf{0}_{d,d}\\
\mathbb{I}_{d(T-1)}&\mathbf{0}_{d(T-1),d}
\end{array}
\right) \quad \mbox{and} \quad
{ C}= \left(
\begin{array}{c}
\mathbb{I}_{d}\\
\mathbf{0}_{d(T-1),d}\\
\end{array}
\right).
\end{equation}  Then the system 
\begin{equation}\label{eq:auxEq4}
\mathbf{x}_t^N = \sigma_1({A} \mathbf{x}_{t-1}^N + {C} {\bf z}_t + {\bf B}), \quad t \in \Z_-,
\end{equation}
admits as unique solution $\mathbf{x}_t^N = ({\bf z}_t^\top,\ldots,{\bf z}_{t-T+1}^\top)^\top$, $t \in \Z_-$.  Next, by \cite[Theorem~3.1]{EW2020} (which is based on \cite[Proposition~1 and Section~IX.15)]{Barron1993}; alternatively the result also follows using the representation obtained in the proof of \cite[Theorem~1]{RC12}), there exists a (Borel) probability measure $\tilde{\mu}$ on $\R \times \R^{Td+1}$ such that $\int_{\R \times \R^{Td+1}} |w| \|{\bf v}\|  \tilde{\mu}(dw,d{\bf v}) < \infty$ and for Lebesgue-a.e.\ ${\bf u} \in (D_d)^{T}$, 
\begin{equation}
\label{eq:auxEq5}
h({\bf u}) = \int_{\R \times \R^{Td+1}} w\sigma_2({\bf v} \cdot ({\bf u}^\top,1)^\top) \tilde{\mu} (dw,d{\bf v}).
\end{equation}
But $h$ is continuous in $(D_d)^{T}$ by \cite[Theorem~6(ii) in Chapter~5.6]{EvansPDEs} and also the right-hand side in \eqref{eq:auxEq5} is continuous in ${\bf u}$, hence the representation \eqref{eq:auxEq5} holds for all ${\bf u} \in (D_d)^{T}$. 

Let $\pi \colon \R \times \R^{Td+1} \to \mathcal{X}^N$ be defined as follows: for $w \in \R$, ${\bf v} \in \R^{Td+1}$ we set $\pi(w,{\bf v}) = (w,({ v}_{d+1},\ldots,{ v}_{Td},0,\ldots,0), ({ v}_{1},\ldots,{ v}_{d}), {v}_{Td +1})$. Denote by $\mu^N = \tilde{\mu} \circ \pi^{-1}$ the pushforward measure of $\tilde{\mu}$ under $\pi$. Then by the change-of-variables formula $\mu^N$ satisfies 
\[ \begin{aligned}
\int_{\mathcal{X}^N} |w| (\|{\bf a}\| + \|{\bf c}\|  +  |b|) \mu^N(dw,d{\bf a},d {\bf c},d b) & \leq \sqrt{3}
\int_{\mathcal{X}^N} |w| (\|{\bf a}\|^2 + \|{\bf c}\|^2  +  |b|^2)^{1/2} \mu^N(dw,d{\bf a},d {\bf c},d b)
\\ & = \sqrt{3} \int_{\R \times \R^{Td+1}} |w| \|{\bf v}\|  \tilde{\mu}(dw,d{\bf v}) < \infty
\end{aligned}
\]
 and for all $ {\bf z} \in  (D_d)^{\Z_-}$, with $g_{\bf z}(w,{\bf a},{\bf c}, b) = w \sigma_2(({\bf c}^\top,{\bf a}^\top,b)  ({\bf z}_0^\top,{\bf z}_{-1}^\top,\ldots,{\bf z}_{-T}^\top,1)^\top)$, 
\begin{equation} \label{eq:auxEq3} 
\begin{aligned} H({\bf z}) & = h({\bf z}_0,\ldots,{\bf z}_{-T+1}) = \int_{\R \times \R^{Td+1}} w\sigma_2({\bf v} \cdot ({\bf z}_0^\top,\ldots,{\bf z}_{-T+1}^\top,1)^\top) \tilde{\mu} (dw,d{\bf v})
\\ & =  \int_{\R \times \R^{Td+1}} g_{\bf z}(\pi(w,{\bf v})) \tilde{\mu} (dw,d{\bf v}) =  \int_{\mathcal{X}^N} w \sigma_2(({\bf c}^\top,{\bf a}^\top,b)  ({\bf z}_0^\top,{\bf z}_{-1}^\top,\ldots,{\bf z}_{-T}^\top,1)^\top) \mu^N(dw,d{\bf a},d {\bf c},d b)
\\ & = \int_{\mathcal{X}^N} w \sigma_2({\bf a} \cdot \mathbf{x}_{-1}^N({\bf z}) + {\bf c} \cdot {\bf z}_0 +  b) \mu^N(dw,d{\bf a},d {\bf c},d b), 
\end{aligned}
\end{equation}
which shows that $H$ is a finite-dimensional recurrent Barron functional, that is $H\in \mathcal{C}_N$.	
\end{proof}

\medskip

\noindent {\bf Linear functionals.} The following propositions show that certain linear functionals are also in the recurrent Barron class $\mathcal{C}$. 

\begin{proposition} 
\label{lem:linear}
Suppose $\sigma_1(x) = x$ and either $ \sigma_2(x)= x$ or $\sigma_2(x)=\max(x,0)$. Let $\mathbf{w} \in \ell^1_-$, ${\bf a}_i \in \R^d$ for $i \in \Z_-$ and assume that $D_d$ is bounded, and $\sup_{i \in \Z_-} \|{\bf a}_i \| < \infty$. Assume that either 
\begin{description}
	\item[(i)] $\sum_{i \in \Z_-} |w_i|\lambda^{i}  < \infty$ for some $\lambda \in (0,1)$
	\item[(ii)] or $\mathcal{I}_d \subset \ell^q_-(D_d)$.
\end{description}   Then the functional
\[
H({\bf z}) = \sum_{i \in \Z_-} w_{i} {\bf a}_{i} \cdot {\bf z}_i , \quad {\bf z} \in \mathcal{I}_d,
\] satisfies $H \in \mathcal{C}$. 
\end{proposition}

\begin{proof} Let $\lambda \in (0,1)$ be as in {\bf (i)} or, in case {\bf  (ii)} holds, let $\lambda =1$.
		
Consider first the case $\sigma_2(x)=x$. 
Let ${\bf B} ={\bf 0}$ and define $A \colon \ell^q \to \ell^q$, $C \colon \R^d \to \ell^q$ by $(A \mathbf{x})_{i} = \mathbbm{1}_{\{i> d\}} \lambda x_{i-d}$, $(C {\bf z})_{i} = \mathbbm{1}_{\{i\leq d\}} { z}_i$ for $\mathbf{x} \in \ell^q$, ${\bf z} \in \R^d$, $i \in \N$. Let us now look at the $i$-th component of \eqref{eq:system}. Inserting the definitions yields
\begin{equation}\label{eq:auxEq1}
{x}_{t,i} = \mathbbm{1}_{\{i> d\}} \lambda{x}_{t-1,i-d} + \mathbbm{1}_{\{i\leq d\}} {z}_{t,i}, \quad t \in \Z_-.
\end{equation}
By iterating \eqref{eq:auxEq1} we see that for $k\in \N$, $j=1,\ldots,d$, we must have ${x}_{t,(k-1)d+j} = \lambda^{k-1} {z}_{t-k+1,j}$. Using this as the definition of $\mathbf{x}$, we hence obtain that \eqref{eq:system} has a unique solution. In addition, this solution satisfies the condition $(\mathbf{x}_t)_{t \in \Z_-} \in \ell^\infty_-(\ell^q)$: suppose that {\bf  (i)} holds, then in the case $q<\infty$ we verify for any $t \in \Z_-$ that $\sum_{k \in \N} |{x}_{t,k}|^q = \sum_{i \in \N} \lambda^{q i} \|{\bf z}_{t-i}\|_q^q \leq d M^q \lambda^q (1-\lambda^q)^{-1}$, where $M>0$ is chosen such that $\|{\bf u}\|_\infty \leq M$ for all ${\bf u} \in D_d$. In the case of $q=\infty$ we verify that $\sup_{k \in \N} | {x}_{t,k}|\leq \sup_{s,k \in \N} |{  z}_{s,k}| \leq M$. Similarly, if {\bf (ii)} holds, $(\mathbf{x}_t)_{t \in \Z_-} \in \ell^\infty_-(\ell^q)$ follows from ${\bf z} \in \mathcal{I}_d \subset \ell^q_-(D_d)$.

Define, for each $i \in \Z_- \setminus \{0\}$, $\bar{\mathbf{a}}_i \in \ell^p$ by setting $\bar{a}_{i,j+d(|i|-1)} = \lambda^{i+1} {a}_{i,j} $ for $j=1,\ldots,d$ and by setting the remaining components of $\bar{\mathbf{a}}_i$ equal to zero. 	
Set $\mu = \frac{1}{\|\mathbf{w}\|_1} \sum_{i \in \Z_-} |w_i| \delta_{(\mathrm{sign}(w_i)\|\mathbf{w}\|_1,\bar{{\bf a}}_i \mathbbm{1}_{\{i<0\}},{\bf a}_0 \mathbbm{1}_{\{i=0\}}, 0)} $, then $\mu$ is indeed a (Borel) probability measure on $\mathcal{X}$ and  
$\mu$ satisfies 
\[\begin{aligned} I_{\mu,p} = \int_{\mathcal{X}} |\bar{w}| (\|{\bf a}\|_p + \|{\bf c}\|  +  |b|) \mu(d\bar{w},d{\bf a},d {\bf c},d b) & = \sum_{i \in \Z_-} |w_i| (\|\bar{{\bf a}}_i \mathbbm{1}_{\{i<0\}}\|_p + \|{\bf a}_0 \mathbbm{1}_{\{i=0\}}\|)
 \\ & \leq \sqrt{d} \sum_{i \in \Z_-} |w_i| \lambda^{i+1} \|{\bf a}_i \| 
 \\ & \leq \sqrt{d} \sup_{i \in \Z_-} \lbrace\|{\bf a}_i \|\rbrace \sum_{i \in \Z_-} |w_i| \lambda^{i+1}  < \infty,
\end{aligned}\]
since $\|\cdot \|_p \leq \|\cdot\|$ for $p\geq 2$ and $\|\cdot \|_p \leq d^{\frac{1}{p}-\frac{1}{2}}\|\cdot\|$ for $p \in [1,2]$. Furthermore, for any ${\bf z} \in  \mathcal{I}_d$, 
\begin{equation} \label{eq:auxEq2} \begin{aligned} \int_{\mathcal{X}} \bar{w} \sigma_2({\bf a} \cdot \mathbf{x}_{-1}({\bf z}) + {\bf c} \cdot {\bf z}_0 +  b) \mu(d\bar{w},d{\bf a},d {\bf c},d b) & =  \sum_{i \in \Z_-} w_i (\bar{\mathbf{a}}_i \mathbbm{1}_{\{i<0\}} \cdot \mathbf{x}_{-1} + {\bf a}_0 \mathbbm{1}_{\{i=0\}} \cdot {\bf z}_0) 
\\
&  = {\bf a}_0  \cdot {\bf z}_0 w_0 + \sum_{i \in \Z_- \setminus \{0\}} w_i  \lambda^{i+1} \sum_{j=1}^d  {a}_{i,j} {x}_{-1,j+d(|i|-1)}  
\\
&  = {\bf a}_0  \cdot {\bf z}_0 w_0 + \sum_{i \in \Z_- \setminus \{0\}} w_i \sum_{j=1}^d  {a}_{i,j} {z}_{-|i|,j}
\\  
&  =  H({\bf z}),
\end{aligned}
\end{equation}
which proves that $H \in \mathcal{C}$ in the case $\sigma_2(x)=x$. 

For the case  $\sigma_2(x)=\max(x,0)$, let $\varphi \colon \mathcal{X} \to \mathcal{X}$ be defined by $\varphi(\bar{w},{\bf a},{\bf c},b) = (-\bar{w},-{\bf a},-{\bf c},-b)$ and consider $\bar{\mu} = \frac{1}{2} \mu + \frac{1}{2} (\mu \circ \varphi^{-1})$. Then the change-of-variables theorem and the fact that $I_{\mu,p} < \infty$ (as established above) show that $I_{\bar{\mu},p} < \infty$ and by \eqref{eq:auxEq2} and $\sigma_2(x)-\sigma_2(-x) = x$ we obtain
\begin{equation} \label{eq:auxEq30} \begin{aligned} & \int_{\mathcal{X}} \bar{w} \sigma_2({\bf a} \cdot \mathbf{x}_{-1}({\bf z}) + {\bf c} \cdot {\bf z}_0 +  b) \bar{\mu}(d\bar{w},d{\bf a},d {\bf c},d b) \\ &  = \frac{1}{2}\int_{\mathcal{X}} \bar{w} \sigma_2({\bf a} \cdot \mathbf{x}_{-1}({\bf z}) + {\bf c} \cdot {\bf z}_0 +  b) \mu(d\bar{w},d{\bf a},d {\bf c},d b) - \frac{1}{2}\int_{\mathcal{X}} \bar{w} \sigma_2(-{\bf a} \cdot \mathbf{x}_{-1}({\bf z}) - {\bf c} \cdot {\bf z}_0 -  b) \mu(d\bar{w},d{\bf a},d {\bf c},d b)
\\ 
&  =  \frac{1}{2} H({\bf z}),
\end{aligned}
\end{equation}
hence $\frac{1}{2} H \in \mathcal{C}$ and, by Proposition~\ref{lem:linearspace}, also $H \in \mathcal{C}$. 
\end{proof}

The next proposition also covers the linear case under slightly different hypotheses.

\begin{proposition} 
\label{lem:linear2}
	Suppose $\sigma_1(x) = x$ and either $ \sigma_2(x)= x$ or $\sigma_2(x)=\max(x,0)$. Let ${\bf h} \in \ell^p_-(\R^d)$  and  $\mathcal{I}_d \subset \ell^q_-(D_d)$.
   Then the functional
	\[
	H({\bf z}) = \sum_{i \in \Z_-} {\bf h}_{i} \cdot {\bf z}_i , \quad {\bf z} \in \mathcal{I}_d,
	\] satisfies $H \in \mathcal{C}$. 
\end{proposition}

\begin{proof}
The case $\sigma_2(x) = \max(x,0)$ can be deduced from the case $\sigma_2(x) = x$ precisely as in the proof of Proposition~\ref{lem:linear}, hence it suffices to consider the case $\sigma_2(x)=x$. Choosing $\lambda =1$ and $A,{\bf B},C$ as in the proof of Proposition~\ref{lem:linear} yields a unique solution $(\mathbf{x}_t)_{t \in \Z_-} \in \ell^\infty_-(\ell^q)$ to \eqref{eq:system} and for $k\in \N$, $j=1,\ldots,d$ we have ${x}_{t,(k-1)d+j} = {z}_{t-k+1,j}$. Define now $\mathbf{a} \in \ell^p$ by setting $a_{(k-1)d+j} = {h}_{-k,j}$ for  $k\in \N$, $j=1,\ldots,d$, and consider $\mu= \delta_{(1,\mathbf{a},{\bf h}_0,0)}$. Then $I_{\mu,p} < \infty$ and 
for any ${\bf z} \in  \mathcal{I}_d$,  
\begin{multline} 
\label{eq:auxEq31} 
 \int_{\mathcal{X}} \bar{w} \sigma_2({\bf a} \cdot \mathbf{x}_{-1}({\bf z}) + {\bf c} \cdot {\bf z}_0 +  b) \mu(d\bar{w},d{\bf a},d {\bf c},d b) =   \mathbf{a} \cdot \mathbf{x}_{-1}({\bf z}) + {\bf h}_0 \cdot {\bf z}_0 
= {\bf h}_0 \cdot {\bf z}_0 + \sum_{k \in \N} \sum_{j=1}^d {h}_{-k,j} {z}_{-k,j} =  H({\bf z}),
\end{multline}
which proves that $H \in \mathcal{C}$.
\end{proof}

\begin{remark}
This proposition covers, in particular, the functionals associated to {\bfi  convolutional filters}. Indeed, let ${\bf h} \in \ell^p_-(\R^d)$ and consider the convolutional filter $U_{\bf h} \colon \ell^q_-(D_d) \to \R^{\Z_-}$ defined by $U_{\bf h}({\bf z})_t = \sum_{j \in \Z_-} {\bf h}_j  \cdot {\bf z}_{t+j}$. Then its associated functional $H \colon \ell^q_-(D_d) \to \R$ defined by $H({\bf z}) = \sum_{j \in \Z_-} {\bf h}_j \cdot {\bf z}_{j}$ is an element of $\mathcal{C}$, as shown in Proposition~\ref{lem:linear2} above. 
\end{remark}

As a further example, we now see that any functional associated with a linear state-space system on a separable Hilbert space admits a realization as an element in $\mathcal{C}$ for $p=2$. 

\begin{proposition}
Let $\mathcal{Y}$ be a separable Hilbert space, let 
$\bar{A} \colon \mathcal{Y} \to \mathcal{Y}$, $\bar{C} \colon \R^d \to \mathcal{Y}$ be linear and $\bar{\bf B} \in \mathcal{Y}$ and assume that for each 
${\bf z} \in  \mathcal{I}_d$ the system 
\begin{equation}\label{eq:systemHilbert}
\bar{\mathbf{x}}_t = \bar{A} \bar{\mathbf{x}}_{t-1} + \bar{C} {\bf z}_t + \bar{\bf B}, \quad t \in \Z_-,
\end{equation}
admits a unique solution $(\bar{\mathbf{x}}_t)_{t \in \Z_-} \in \ell^\infty_-(\mathcal{Y})$, i.e., satisfying $\sup_{t \in \Z_-} \|\bar{\mathbf{x}}_t\|_{\mathcal{Y}}< \infty$. Let $\bar{\mu}$ be a (Borel) probability measure on $\R \times \mathcal{Y} \times \R^d \times \R$ with $\int_{\R \times \mathcal{Y} \times \R^d \times \R} |w| (\|{\bf a}\|_{\mathcal{Y}} + \|{\bf c}\|  +  |b|)  \bar{\mu}(dw,d{\bf a},d {\bf c},d b) < \infty$ and consider  
\begin{equation} \label{eq:HrepresHilbert} H({\bf z}) = \int_{\R \times \mathcal{Y} \times \R^d \times \R} w \sigma_2(\langle{\bf a}, \bar{\mathbf{x}}_{-1}({\bf z})\rangle_{\mathcal{Y}} + {\bf c} \cdot {\bf z}_0 +  b) \bar{\mu}(dw,d{\bf a},d {\bf c},d b), \quad  {\bf z} \in  \mathcal{I}_d.
\end{equation}
Then $H \in \mathcal{C}$ with $p=2$. 
\end{proposition}

\begin{proof}
Denote by $T \colon \mathcal{Y} \to \ell^2$ an isometric isomorphism between $\mathcal{Y}$ and $\ell^2$. Define $\mathbf{x}_t:=T \bar{\mathbf{x}}_t$, $A=T\bar{A}T^{-1}$, $C = T\bar{C}$, ${\bf B } =T \bar{\bf B}$, $\phi \colon \R \times \mathcal{Y} \times \R^d \times \R \to \mathcal{X}$, $\phi(w,{{\bf a}},{\bf c},b)=(w,T{\bf a},{\bf c},b)$ and let $\mu = \bar{\mu} \circ \phi^{-1}$ be the pushforward measure of $\bar{\mu}$ under $\phi$. 

Then $\mu$ is a probability measure  on $\R \times \ell^2 \times \R^d \times \R$ and $A \colon \ell^2 \to \ell^2$, $C \colon \R^d \to \ell^2$ are linear maps. Furthermore, by choice of $A,{\bf B},C$ we have that $\mathbf{x} \in \ell^\infty_-(\ell^2)$ is a solution \eqref{eq:system}. On the other hand, if $(\mathbf{x}_t)_{t \in \Z_-} \in \ell^\infty_-(\ell^2)$ is any solution to \eqref{eq:system}, then $\bar{\mathbf{x}}_t:=T^{-1}\mathbf{x}_t$ defines a solution to \eqref{eq:systemHilbert}. But the latter system has a unique solution by assumption, hence also the solution to \eqref{eq:system} is unique. 

Finally, the change-of-variables formula and the fact that $T$ is an isometry imply that $\mu$ satisfies $I_{\mu,2}=\int_{\R \times \mathcal{Y} \times \R^d \times \R} |w| (\|T{\bf a}\|_2 + \|{\bf c}\|  +  |b|) \bar{\mu}(dw,d{\bf a},d {\bf c},d b) < \infty$ and
\begin{equation} \label{eq:HrepresHilbertproof}
\begin{aligned} H({\bf z}) & = \int_{\R \times \mathcal{Y} \times \R^d \times \R} w \sigma_2(T{\bf a} \cdot T\bar{\mathbf{x}}_{-1}({\bf z}) + {\bf c} \cdot {\bf z}_0 +  b) \bar{\mu}(dw,d{\bf a},d {\bf c},d b)
\\ & = \int_{\R \times \ell^2 \times \R^d \times \R} w \sigma_2({\bf a} \cdot \mathbf{x}_{-1}({\bf z}) + {\bf c} \cdot {\bf z}_0 +  b) \mu(dw,d{\bf a},d {\bf c},d b).
\end{aligned}
\end{equation}
This shows that the functional $H$ can be represented as a readout \eqref{eq:Hrepres} of a system \eqref{eq:system}, hence $H \in \mathcal{C} $. 
It also proves that $T$ is a system isomorphism between the system determined by $\bar{A}$, $\bar{C}$,  $\bar{B}$, and $\bar{\mu} $, and the system in $\mathcal{C} $ determined by ${A}$, ${C}$,  ${B}$, and ${\mu} $. 
\end{proof}

The next proposition is a generalization of a universality result in \cite[Theorem~III.13]{RC8} to the context of recurrent Barron functionals.

\begin{proposition}
\label{lem:dense}
	 Assume $\sigma_1(x)=x$ and either $\sigma_2(x)=\max(x,0)$ or $\sigma_2$ is bounded, continuous and non-constant. Let $\bar{p} \in [1,\infty)$ and let $\gamma$ be a probability measure on $\mathcal{I}_d \subset \ell^\infty_-(D_d)$. Then $\mathcal{C} \cap L^{\bar{p}}(\mathcal{I}_d,\gamma)$ is dense in $L^{\bar{p}}(\mathcal{I}_d,\gamma)$. 
\end{proposition}

\begin{proof}
Let $H \in L^{\bar{p}}(\mathcal{I}_d,\gamma)$ and $\varepsilon > 0$. Consider first the case $\sigma_2(x)=\max(x,0)$. Define the activation function $\bar{\sigma}(x) = \sigma_2(x+1)-\sigma_2(x)$, then  $|\bar{\sigma}(x)| \leq 1$ for $x \in \R$. 
Hence, by \cite[Theorem~III.13]{RC8} there exists a reservoir functional $H^{RC}$ associated to a linear system with neural network readout such that $H^{RC} \in L^{\bar{p}}(\mathcal{I}_d,\gamma)$ and
$\|H-H^{RC}\|_{L^{\bar{p}}(\mathcal{I}_d,\gamma)} < \varepsilon$. More specifically, there exist $N \in \N$, ${A}^N \in \R^{N \times N}$, ${C}^N \in \R^{N\times d}$ such that the system 
\begin{equation}\label{eq:auxEqN}
\mathbf{x}_t^N = {A}^N \mathbf{x}_{t-1}^N + {C}^N \mathbf{z}_t, \quad t \in \Z_-
\end{equation}
has the echo state property and 
 $H^{RC}(\mathbf{z}) = h(\mathbf{x}_0^N({\bf z}))$ for some
 neural network $h \colon \R^N \to  \R$ given by 
\[
h(\mathbf{x}) = \sum_{j=1}^k \beta_j \bar{\sigma}(\bm{\alpha}_j \cdot \mathbf{x} - \theta_j)
\]
for some $k \in \N$, $\beta_j, \theta_j \in \R$ and $\bm{\alpha}_j \in \R^N$.
We claim that $H^{RC} \in \mathcal{C}_N$. To prove this, 
note that with the choices ${A} = {A}^N$, ${\bf B} = {\bf 0}$ and ${C}={C}^N$ the process $\mathbf{x}^N$ is the unique solution to the system \eqref{eq:systemN} due to the echo state property of \eqref{eq:auxEqN}. From the proof of \cite[Theorem~III.13]{RC8} and the assumption that $\mathcal{I}_d \subset \ell^\infty_-(D_d)$ we obtain that the unique solution satisfies $\mathbf{x}^N \in \ell^\infty_-(\R^N)$.
Set now $\mu^N = \frac{1}{2k} \sum_{j=1}^k \delta_{(2k\beta_j,{ A}^\top\bm{\alpha}_j,{ C}^\top \bm{\alpha}_j,-\theta_j+1)} + \delta_{(-2k\beta_j,{ A}^\top \bm{\alpha}_j,{ C}^\top \bm{\alpha}_j,-\theta_j)} $. Then   $\mu^N$ is a probability measure on $\mathcal{X}^N$ and  for ${\bf z} \in  \mathcal{I}_d$
\begin{equation*} 
\begin{aligned}
& \int_{\mathcal{X}^N} w \sigma_2({\bf a} \cdot \mathbf{x}_{-1}^N({\bf z}) + {\bf c} \cdot {\bf z}_0 +  b) \mu^N(dw,d{\bf a},d {\bf c},d b) 
\\ &  = \frac{1}{2k} \sum_{j=1}^k  2k\beta_j \sigma_2(\bm{\alpha}_j \cdot ({A}\mathbf{x}_{-1}^N({\bf z}) +  {C} {\bf z}_0)   -\theta_j+1)   -2k\beta_j \sigma_2(\bm{\alpha}_j \cdot ({A}\mathbf{x}_{-1}^N({\bf z}) +  {C} {\bf z}_0) -\theta_j) 
\\ &  = \sum_{j=1}^k  \beta_j \bar{\sigma}(\bm{\alpha}_j \cdot \mathbf{x}_{0}^N({\bf z})    -\theta_j) = h(\mathbf{x}_{0}^N({\bf z})) = H^{RC}(\mathbf{z}).
\end{aligned}
\end{equation*}
This shows that $H^{RC} \in \mathcal{C}_N$  and $\|H-H^{RC}\|_{L^{\bar{p}}(\mathcal{I}_d,\gamma)} < \varepsilon$. Combining this with Proposition~\ref{lem:finite} yields the claim.

In the case where $\sigma$ is bounded, continuous and non-constant we may directly work with $\sigma$ (instead of $\bar{\sigma}$) and proceed similarly.
\end{proof}

\medskip

\noindent {\bf Generalized convolutional functionals.} 
We conclude by showing that the class $\mathcal{C}$ contains functionals that generalize convolutional filters, an important family of transforms.
\begin{proposition}
\label{lem:infinite}
Assume $\sigma_1(x)=x$, $p \in (1,\infty)$ and suppose $\mathcal{I}_d \subset \ell^q_-(D_d)$.  Let $\mu$ be a probability measure on $\mathcal{X}$ with $I_{\mu,p} < \infty$. Then the functional $H \colon \mathcal{I}_d \to \R$,
\begin{equation} \label{eq:auxEq48} H({\bf z}) = \int_{\mathcal{X}} w \sigma_2({\bf a} \cdot {\bf z}_{-1:-\infty} + {\bf c} \cdot {\bf z}_0 +  b) \mu(dw,d{\bf a},d {\bf c},d b), \quad  {\bf z} \in  \mathcal{I}_d
\end{equation}
is in $\mathcal{C}$. 
\end{proposition}

\begin{proof} We only need to construct ${\bf B} \in \ell^q$ and linear maps $A \colon \ell^q \to \ell^q$, $C \colon \R^d \to \ell^q$ such that for each ${\bf z} \in  \mathcal{I}_d$ the system 
	\begin{equation}\label{eq:systemProof}
	\mathbf{x}_t = A \mathbf{x}_{t-1} + C {\bf z}_t + {\bf B}, \quad t \in \Z_-,
	\end{equation}
	admits the unique solution $(\mathbf{x}_t)_{t \in \Z_-} = ({\bf z}_{t:-\infty})_{t \in \Z_-} \in \ell^\infty_-(\ell^q)$. Then $\mathbf{x}_{-1} = {\bf z}_{-1:-\infty}$ and so $H$ has representation \eqref{eq:Hrepres}.  
	Choose ${\bf B}={\bf 0}$, let $C{\bf y}$ be the sequence with the components of ${\bf y}$ in the first $d$ entries and $0$ otherwise, and let $A$ be the operator that shifts the $i$-th entry of the input to the $i+d$-th entry and inserts $0$ in the first $d$ entries. Then  $A \colon \ell^q \to \ell^q$, $C \colon \R^d \to \ell^q$ are linear,  and indeed $\mathbf{x}_t = {\bf z}_{t:-\infty}$ for all $t \in \Z_-$ is the unique solution to \eqref{eq:systemProof} and  $\mathbf{x} \in \ell^\infty_-(\ell^q)$. 
\end{proof}

Notice that standard convolutional filters are a special case of the functional in \eqref{eq:auxEq48} when $\mu$ is a Dirac measure and $\sigma_2$ is the identity.

\section{Approximation}
\label{sec:approximation}

In this section, we establish approximation results for the concept class $ \mathcal{C}$ of recurrent Barron functionals. As approximating systems, we use a modification of the finite-dimensional echo state networks \eqref{prescription esn}--\eqref{readout} in which the linear readout $W$ in \eqref{readout} is replaced by an extreme learning machine (ELM) / random features neural network, that is, a  neural network of the type \eqref{eq:alternativereadout} whose parameters are all randomly generated, except for the output layer given by the weights $(W _i)_{i=1, \ldots, N} $ which are trainable. To derive such bounds, we proceed in several approximation steps and {\bf (a)} approximate the infinite system \eqref{eq:system} by a finite system of type \eqref{prescription esn}, {\bf (b)} apply a Monte Carlo approximation of the integral in \eqref{eq:Hrepres}, and {\bf (c)} use an importance sampling procedure to guarantee that the neural network weights can be generated from a generic distribution rather than the (unknown) measure $\mu$.

\subsection{Approximation by a finite-dimensional system}

We start by showing that the elements in $\mathcal{C} $ can be approximated by finite-dimensional truncations, that is, by finite-dimensional recurrent Barron functionals in $\mathcal{C} _N $, for any $N \in \mathbb{N} $. 

\medskip

\begin{proposition} \label{prop:truncation}
Assume $q \in [1,\infty)$ and $\mathcal{I}_d \subset \ell^\infty_-(\R^d)$.
Suppose $H \in \mathcal{C}$ has representation \eqref{eq:Hrepres} with probability measure $\mu$ on $\mathcal{X}$, ${\bf B} \in \ell^q$, and bounded, linear maps $A \colon \ell^q \to \ell^q$, $C \colon \R^d \to \ell^q$. Let $N \in \N$ and for $j \in \N$, $k=1,\ldots,d$ let $\boldsymbol{\epsilon}^j = (\delta_{ij})_{i \in \N} \in \R^\N$ and ${\bf e}^k = (\delta_{ik})_{i=1,\ldots,d} \in \R^d$. Let $\bar{\bf B} \in \R^N$, $\bar{A} \in \R^{N \times N}$, $\bar{C}  \in \R^{N \times d}$ be given as $\bar{B}_i = B_i$, $\bar{A}_{ij} = (A\boldsymbol{\epsilon}^j)_i$, $\bar{C}_{ik} = (C{\bf e}^k)_i$ for  $i,j=1,\ldots,N$, $k=1,\ldots,d$. Assume $\sigma_1$ is $L_{\sigma_1}$-Lipschitz and $L_{\sigma_1} \vertiii{A} <1$. Then the following statements hold:
\begin{description}
\item[(i)] For each ${\bf z} \in  \mathcal{I}_d$ the system 
\begin{equation}\label{eq:systemN2}
\mathbf{x}_t^N = \sigma_1(\bar{A} \mathbf{x}_{t-1}^N + \bar{C} {\bf z}_t + \bar{\bf B}), \quad t \in \Z_-,
\end{equation}
admits a unique solution $(\mathbf{x}_t^N)_{t \in \Z_-} \in \ell^\infty_-(\R^N)$.
\item[(ii)] Write $\mathbf{x}_t^N({\bf z}) = \mathbf{x}_t^N $ and define the functional $H^N \colon \mathcal{I}_d \to \R$ by 
\begin{equation} \label{eq:HapproxDef} H^N({\bf z}) = \int_{\mathcal{X}} w \sigma_2({\bf a} \cdot \iota(\mathbf{x}_{-1}^N({\bf z})) + {\bf c} \cdot {\bf z}_0 +  b) \mu(dw,d{\bf a},d {\bf c},d b), \quad  {\bf z} \in  \mathcal{I}_d.
\end{equation}
Then there exists a constant $C_{\mathrm{fin}} >0 $ not depending on $N$ such that for all ${\bf z} \in  \mathcal{I}_d$, 
\begin{equation}
\label{eq:truncationError}
|H({\bf z})-H^N({\bf z})| \leq  C_{\mathrm{fin}} \sum_{k=0}^\infty (L_{\sigma_1} \vertiii{A})^k \left(\sum_{i=N+1}^\infty |{x}_{-1-k,i}|^q\right)^{1/q}.
\end{equation}
In particular, $\lim_{N \to \infty} H^N({\bf z}) = H({\bf z})$. The constant is given by  $C_{\mathrm{fin}} =  L_{\sigma_2} \int_{\mathcal{X}} |w| \| {\bf a} \|_p   \mu(dw,d{\bf a},d {\bf c},d b)$. 
\item[(iii)] $H^N \in \mathcal{C}_N$. 
\end{description}
\end{proposition}

\begin{proof} Part {\bf (i)} is a consequence of Proposition \ref{usp infinite-dimensional} and of the inequality $\vertiii{\bar{A}}\leq \vertiii{A} $ which in turn implies that $L_{\sigma _1}\vertiii{\bar{A}}<1 $. Regarding {\bf (ii)}, note first that by our choice of $\bar{C}$, it follows  that $(C{\bf u})_i = \sum_{k=1}^d {u}_k (C{\bf e}^k)_i =  \sum_{k=1}^d {u}_k \bar{C}_{ik} = (\bar{C} {\bf u})_i$ for all ${\bf u} \in \R^d$, $i=1,\ldots,N$. Similarly, for ${\bf y} \in \R^N$, $i=1,\ldots,N$ we have that  $(\bar{A} {\bf y})_i = \sum_{j=1}^N {y}_j \bar{A}_{ij} = \sum_{j=1}^N {y}_j (A\boldsymbol{\epsilon}^j)_i = (A \iota({\bf y}))_i $.
Consequently, for $i=1,\ldots,N$: 
\begin{equation}
\label{eq:auxEq8}
\begin{aligned}
 |{x}_{t,i}-{x}_{t,i}^N| & \leq L_{\sigma_1}|(A \mathbf{x}_{t-1})_i + (C {\bf z}_t)_i + B_i- (\bar{A} \mathbf{x}_{t-1}^N)_i - (\bar{C} {\bf z}_t)_i - \bar{B}_i|
 \\ & = L_{\sigma_1}|(A (\mathbf{x}_{t-1} - \iota(\mathbf{x}_{t-1}^N)))_i |.
\end{aligned}
\end{equation}
Denote by $\pi \colon \ell^q \to \R^N$ the projection onto the first $N$ coordinates. Then
\begin{equation}
\label{eq:auxEq7}
\begin{aligned}
 \| \mathbf{x}_{t}({\bf z})  - \iota(\mathbf{x}_{t}^N({\bf z}))\|_q & \leq  \|\mathbf{x}_{t}({\bf z}) - \iota(\pi(\mathbf{x}_{t}({\bf z}))) \|_q +  \| \iota(\pi(\mathbf{x}_{t}({\bf z})))  - \iota(\mathbf{x}_{t}^N({\bf z}))\|_q
 \\ & = \left(\sum_{i=N+1}^\infty |{x}_{t,i}|^q\right)^{1/q} + \left(\sum_{i=1}^N |{x}_{t,i}-{x}_{t,i}^N|^q\right)^{1/q}
 \\ & \leq \left(\sum_{i=N+1}^\infty |{x}_{t,i}|^q\right)^{1/q} + L_{\sigma_1} \left(\sum_{i=1}^N |(A (\mathbf{x}_{t-1} - \iota(\mathbf{x}_{t-1}^N)))_i |^q\right)^{1/q}
  \\ & \leq \left(\sum_{i=N+1}^\infty |{x}_{t,i}|^q\right)^{1/q} + L_{\sigma_1} \|A (\mathbf{x}_{t-1} - \iota(\mathbf{x}_{t-1}^N)) \|_q
    \\ & \leq \left(\sum_{i=N+1}^\infty |{x}_{t,i}|^q\right)^{1/q} + L_{\sigma_1} \vertiii{A} \| \mathbf{x}_{t-1} - \iota(\mathbf{x}_{t-1}^N) \|_q.
\end{aligned}
\end{equation}
Iterating \eqref{eq:auxEq7} thus yields 
\begin{equation}
\label{almost there fininfi}
 \| \mathbf{x}_{t}({\bf z})  - \iota(\mathbf{x}_{t}^N({\bf z}))\|_q \leq \sum_{k=0}^\infty (L_{\sigma_1} \vertiii{A})^k \left(\sum_{i=N+1}^\infty |{x}_{t-k,i}|^q\right)^{1/q}.
\end{equation}
This can now be used to estimate the difference between $H$ and its truncation: 
\begin{equation}
\label{eq:auxEq6}
\begin{aligned}
|H({\bf z})-H^N({\bf z})| &  \leq \int_{\mathcal{X}} |w| | \sigma_2({\bf a} \cdot \mathbf{x}_{-1}({\bf z}) + {\bf c} \cdot {\bf z}_0 +  b) - \sigma_2({\bf a} \cdot \iota(\mathbf{x}_{-1}^N({\bf z})) + {\bf c} \cdot {\bf z}_0 +  b)| \mu(dw,d{\bf a},d {\bf c},d b)
\\ & \leq L_{\sigma_2} \int_{\mathcal{X}} |w| | {\bf a} \cdot \mathbf{x}_{-1}({\bf z})  - {\bf a} \cdot \iota(\mathbf{x}_{-1}^N({\bf z}))| \mu(dw,d{\bf a},d {\bf c},d b)
\\ & \leq L_{\sigma_2} \int_{\mathcal{X}} |w| \| {\bf a} \|_p   \mu(dw,d{\bf a},d {\bf c},d b) \| \mathbf{x}_{-1}({\bf z})  - \iota(\mathbf{x}_{-1}^N({\bf z}))\|_q
\\& \leq C_{\mathrm{fin}} \sum_{k=0}^\infty (L_{\sigma_1} \vertiii{A})^k \left(\sum_{i=N+1}^\infty |{x}_{-1-k,i}|^q\right)^{1/q}
\end{aligned}
\end{equation}
with $C_{\mathrm{fin}} =  L_{\sigma_2} \int_{\mathcal{X}} |w| \| {\bf a} \|_p   \mu(dw,d{\bf a},d {\bf c},d b)$. 

It remains to be shown that $H^N \in \mathcal{C}_N$. 
Recall that $\pi \colon \ell^p \to \R^N$ denotes the projection onto the first $N$ components and let $\phi \colon \mathcal{X} \to \mathcal{X}^N $ be the product map $\phi = \mathrm{Id}_{\R} \times \pi \times \mathrm{Id}_{\R^d} \times \mathrm{Id}_{\R}$, that is, $\phi(w,{\bf a},{\bf c},b)=(w,(a_i)_{i=1,\ldots,N},{\bf c},b)$. Denote by $\mu^N = \mu \circ \phi^{-1}$ the pushforward measure of $\mu$ under $\phi$. Then $\mu^N$ is a (Borel) probability measure on $\mathcal{X}^N$ and in $(i)$ we showed that for each ${\bf z} \in  \mathcal{I}_d$ the system 
\eqref{eq:systemN2}
admits a unique solution. In addition, the integral transformation theorem implies that
$\mu^N$ satisfies 
\[
\int_{\mathcal{X}^N} |w| (\|{\bf a}\| + \|{\bf c}\|  +  |b|) \mu^N(dw,d{\bf a},d {\bf c},d b) = \int_{\mathcal{X}} |w| (\|\pi({\bf a})\| + \|{\bf c}\|  +  |b|) \mu(dw,d{\bf a},d {\bf c},d b) < \infty
\]
 and, writing  $\mathbf{x}_t^N = \mathbf{x}_t^N({\bf z})$, 
\begin{equation} \begin{aligned} H^N({\bf z}) &  = 
\int_{\mathcal{X}} w \sigma_2({\bf a} \cdot \iota(\mathbf{x}_{-1}^N({\bf z})) + {\bf c} \cdot {\bf z}_0 +  b) \mu(dw,d{\bf a},d {\bf c},d b)
\\
& = 
\int_{\mathcal{X}} w \sigma_2(\pi({\bf a}) \cdot \mathbf{x}_{-1}^N({\bf z}) + {\bf c} \cdot {\bf z}_0 +  b) \mu(dw,d{\bf a},d {\bf c},d b)
\\ 
& = 
 \int_{\mathcal{X}^N} w \sigma_2({\bf a} \cdot \mathbf{x}_{-1}^N({\bf z}) + {\bf c} \cdot {\bf z}_0 +  b) \mu^N(dw,d{\bf a},d {\bf c},d b), \quad  {\bf z} \in  \mathcal{I}_d.
 \end{aligned}
\end{equation}
This proves that $H^N \in \mathcal{C}_N$. 
\end{proof}

\subsection{Normalized realizations} 
The next result shows that a large class of elements in the concept class $\mathcal{C}$ can be transformed into what we call a {\bfi normalized realization} in which the solution map for \eqref{eq:system} is a multiplicative scaling operator. 

\begin{proposition}
\label{lem:canonical}
Assume $\sigma_1(x)=x$, $p \in (1,\infty)$ and $D_d$ is bounded. Suppose $H \in \mathcal{C}$ has a realization of the type \eqref{eq:system}-\eqref{eq:Hrepres}  with $C$ bounded and bounded $A \colon \ell^q \to \ell^q$ satisfying $\rho(A)<1$, which guarantees the existence of a positive integer $k _0\in \mathbb{N} $ such that  $\vertiii{A^k}<1 $, for all $k\geq k _0 $. Fix $\lambda \in \left(\vertiii{A^{k _0}}^{1/k _0},1\right)$. Then there exists a (Borel) probability measure $\bar{\mu}$ on $\mathcal{X}$ satisfying $\int_{\mathcal{X}} |w| (\|{\bf a}\|_p + \|{\bf c}\|  +  |b|) \bar{\mu}(dw,d{\bf a},d {\bf c},d b) < \infty$
such that $H$ has the  alternative representation 
\begin{equation} 
\label{eq:HrepresCanonical} H({\bf z}) = \int_{\mathcal{X}} w \sigma_2({\bf a} \cdot (\bar{\Lambda} {\bf z}) + {\bf c} \cdot {\bf z}_0 +  b) \bar{\mu}(dw,d{\bf a},d {\bf c},d b), \quad  {\bf z} \in  \mathcal{I}_d,
\end{equation}
where $\bar{\Lambda} \colon (D_d)^{\Z_-} \to \ell^q$, $(\bar{\Lambda} {\bf z})_{d(k-1)+j} = \lambda^{k-1} z_{-k,j}$ for $k \in \N$, $j=1,\ldots,d$. 
\end{proposition}

\begin{remark}
\normalfont
The assumption that $D_d$ is bounded in the previous statement could be relaxed to the requirement that $\sum_{k \in \N} \|{\bf z}_{-k}\| \lambda^k < \infty$.  
\end{remark}

\begin{proof}
First, we recall that by Proposition \ref{usp infinite-dimensional}, the hypotheses in the statement guarantee that the system \eqref{eq:system} admits a unique solution $(\mathbf{x}_t({\bf z}))_{t \in \Z_-} \in \ell^\infty_-(\ell^q)$ for each ${\bf z} \in  \mathcal{I}_d$, and that $\mathbf{x}_t({\bf z}) $ is explicitly given by the expression \eqref{eq:auxEq10}.  
The existence of a positive integer $k _0\in \mathbb{N} $ such that  $\vertiii{A^k}<1 $ for all $k\geq k _0 $ is guaranteed by Gelfand's formula for the expression of the spectral radius $\rho(A) $.

Next, we define $\tilde{A} = \lambda^{-1} A $ and note that the choice $\lambda \in \left(\vertiii{A^{k _0}}^{1/k _0},1\right)$ guarantees that $\vertiii{\tilde{A}^{k _0}}<1 $. Moreover, 
for each $k \in \N_0$ the map $\tilde{A}^k C \colon \R^d \to \ell^q$ is a bounded linear map between two Banach spaces and hence its adjoint is a bounded linear map from $(\ell^q)^* \to (\R^d)^*$ which we can identify with $(\tilde{A}^k C)^* \colon \ell^p \to \R^d$. More specifically, for $y \in (\ell^q)^*, {\bf u} \in \R^d$ the adjoint is defined by $[(\tilde{A}^k C)^*(y)]({\bf u}) = [y \circ (\tilde{A}^k C)]({\bf u}) = y(\tilde{A}^k C {\bf u})$ and so with the identification $y({\bf b}) = {\bf a_y} \cdot {\bf b} $ for some ${\bf a_y} \in \ell^p$ and all ${\bf b} \in \ell^q$ we obtain that the adjoint property translates to $(\tilde{A}^k C)^*({\bf a_y}) \cdot {\bf u} = {\bf a_y} \cdot (\tilde{A}^k C) {\bf u}$.

Thus, we may 
 consider the map $\mathcal{L} \colon \ell^p \to ((\R^d)^{\N_0})$ given by $[\mathcal{L}({\bf a})]_k = (\tilde{A}^k C)^* {\bf a} $ for $k \in \N_0$, ${\bf a} \in \ell^p$. We claim that the image $\mathcal{L}(\ell^p)$ can be identified with a subset of $\ell^p$. 
Indeed,
\begin{multline} 
\label{eq:auxEq11} 
\!\!\!\!\!\!\sum_{k \in \N_0} \|[\mathcal{L}({\bf a})]_k\|^p = \sum_{k \in \N_0} \|(\tilde{A}^k C)^* {\bf a}\|^p \leq \sum_{k \in \N_0} \vertiii{(\tilde{A}^k C)^*}^p \| {\bf a}\|_p^p 
= \sum_{k \in \N_0} \vertiii{\tilde{A}^k C}^p \| {\bf a}\|_p^p
 \leq  (\vertiii{C}\| {\bf a}\|_p)^p \sum_{k \in \N_0} \vertiii{\tilde{A}^k}^{p}\\
\leq(\vertiii{C}\| {\bf a}\|_p)^p \sum_{j \in \N_0}\sum_{i=0}^{k _0-1} \vertiii{\tilde{A}^{k_0}}^{jp}\vertiii{\tilde{A}^{i}}^{p}
=(\vertiii{C}\| {\bf a}\|_p)^p\sum_{i=0}^{k _0-1} \frac{\vertiii{\tilde{A}^{i}}^{p}}{1-\vertiii{\tilde{A}^{k_0}}^{p}} < \infty,
\end{multline}
since $\vertiii{\tilde{A}^{k _0}} = \lambda^{-k _0} \vertiii{A^{k _0}} < 1$.  
Thus, the map $\bar{\mathcal{L}} \colon \ell^p \to \ell^p$ given by $[\bar{\mathcal{L}}({\bf a})]_{d(k-1)+j} = [(\tilde{A}^{k-1} C)^* {\bf a}]_j $ for $k \in \N$, $j=1,\ldots,d$ is well-defined and from \eqref{eq:auxEq11} we deduce that
\begin{equation} \label{eq:auxEq15} 
\begin{aligned}
\|\bar{\mathcal{L}}({\bf a})\|_p = \left(\sum_{k \in \N} \sum_{j=1}^d [(\tilde{A}^{k-1} C)^* {\bf a}]_j^p\right)^{1/p}
= \left(\sum_{k \in \N_0} \| [\mathcal{L}({\bf a})]_k \|_p^p\right)^{1/p} \leq \left({d} \sum_{k \in \N_0} \| [\mathcal{L}({\bf a})]_k \|^p\right)^{1/p} \leq  c_0 \| {\bf a}\|_p
\end{aligned},
\end{equation}
where $c_0 = d^{1/p} \vertiii{C}\left(\sum_{i=0}^{k _0-1} \frac{\vertiii{\tilde{A}^{i}}^{p}}{1-\vertiii{\tilde{A}^{k_0}}^{p}}\right)^{1/p}$ and we have used that $\|\cdot \|_p \leq \|\cdot\|$ for $p\geq 2$ and $\|\cdot \|_p \leq d^{\frac{1}{p}-\frac{1}{2}}\|\cdot\|$ for $p \in [1,2]$. 
We may now use \eqref{eq:auxEq10} to rewrite ${\bf a} \cdot \mathbf{x}_{-1}({\bf z})$  as 
\begin{equation} \label{eq:auxEq12} 
{\bf a} \cdot \mathbf{x}_{-1}({\bf z}) =  {\bf a} \cdot \bar{\bf B}_0 + \sum_{i \in \N} a_i \sum_{k=0}^\infty ({A}^k C {\bf z}_{-1-k})_i  =  {\bf a} \cdot \bar{\bf B}_0 + \sum_{k=0}^\infty {\bf a } \cdot  (\tilde{A}^k C {\bf z}_{-1-k}\lambda^{k}),
\end{equation}
where $\bar{\bf B}_0 = \sum_{k=0}^\infty A^k {\bf B} \in \ell^q$ and the series can be interchanged by Fubini's theorem, as we see by estimating 
\begin{multline*}
\sum_{k=0}^\infty \sum_{i \in \N}  |a_i| |(A^k C {\bf z}_{-1-k})_i|\leq \sum_{k=0}^\infty \|\mathbf{a}\|_p \|{A}^k C {\bf z}_{-1-k}\|_q \leq \sum_{k=0}^\infty \|\mathbf{a}\|_p \vertiii{A^k} \vertiii{C} \|{\bf z}_{-1-k}\|_q \\
\leq \|\mathbf{a}\|_p \vertiii{C} \sum_{j \in \N_0}\sum_{i=0}^{k _0-1} \vertiii{A^{k_0}}^j  \vertiii{A^i}\|{\bf z}_{-1-k}\|_q
< \infty, 
\end{multline*}
since $\vertiii{A^{k_0}} <1 $ and $D_d$ is bounded. The second term in \eqref{eq:auxEq12} can be rewritten as  
\begin{equation} \label{eq:auxEq13} 
\begin{aligned}
\sum_{k=0}^\infty {\bf a } \cdot  (\tilde{A}^k C {\bf z}_{-1-k} \lambda^k) & =   \sum_{k=0}^\infty ((\tilde{A}^k C)^*{\bf a }) \cdot   ({\bf z}_{-1-k}\lambda^k) = \sum_{k=0}^\infty \sum_{j=1}^d [(\tilde{A}^k C)^*{\bf a }]_j   [{z}_{-1-k}]_j \lambda^k
\\ & = \sum_{k=1}^\infty \sum_{j=1}^d [(\tilde{A}^{k-1} C)^*{\bf a }]_j   [{z}_{-k}]_j \lambda^{k-1}  = \sum_{k=1}^\infty \sum_{j=1}^d [\bar{\mathcal{L}}({\bf a})]_{d(k-1)+j} (\bar{\Lambda} {\bf z})_{d(k-1)+j}  \\
	&=  \bar{\mathcal{L}}({\bf a}) \cdot (\bar{\Lambda} {\bf z}).
\end{aligned}
\end{equation}
Combining \eqref{eq:auxEq13} and \eqref{eq:auxEq12} and inserting in \eqref{eq:Hrepres} then yields
\begin{equation} \label{eq:auxEq14} 
\begin{aligned}
H({\bf z}) = \int_{\mathcal{X}} w \sigma_2({\bf a} \cdot \bar{\bf B}_0 + \bar{\mathcal{L}}({\bf a}) \cdot (\bar{\Lambda} {\bf z}) + {\bf c} \cdot {\bf z}_0 +  b) \mu(dw,d{\bf a},d {\bf c},d b), \quad  {\bf z} \in  \mathcal{I}_d.
\end{aligned}
\end{equation}
Let $\bar{\phi} \colon \mathcal{X} \to \mathcal{X}$ be given as $\bar{\phi}(w,{\bf a},{\bf c},b)=(w,\bar{\mathcal{L}}({\bf a}),{\bf c},{\bf a} \cdot \bar{\bf B}_0+ b)$ and denote by $\bar{\mu} = \mu \circ \bar{\phi}^{-1}$ the pushforward measure of $\mu$ under $\bar{\phi}$. Then the change-of-variables formula and \eqref{eq:auxEq15} show that
\[ \begin{aligned}
I_{\bar{\mu},p}  & = \int_{\mathcal{X}} |w| (\|\bar{\mathcal{L}}({\bf a})\|_p + \|{\bf c}\|  +  |{\bf a} \cdot \bar{\bf B}_0+ b|) \mu(dw,d{\bf a},d {\bf c},d b) 
\\ & \leq \int_{\mathcal{X}} |w| (c_0 \| {\bf a}\|_p + \|{\bf c}\|  +  \|{\bf a}\|_p \|\bar{\bf B}_0\|_q + |b|) \mu(dw,d{\bf a},d {\bf c},d b) \\ &  \leq \max(c_0+\|\bar{\bf B}_0\|_q,1) I_{\mu,p}  < \infty
\end{aligned}
\]
and similarly we obtain from \eqref{eq:auxEq14} that for any ${\bf z} \in  \mathcal{I}_d$,
\begin{equation} \label{eq:auxEq16} 
\begin{aligned}
H({\bf z}) & = \int_{\mathcal{X}} w \sigma_2({\bf a} \cdot \bar{\bf B}_0 + \bar{\mathcal{L}}({\bf a}) \cdot (\bar{\Lambda} {\bf z}) + {\bf c} \cdot {\bf z}_0 +  b) \mu(dw,d{\bf a},d {\bf c},d b) \\ &  = \int_{\mathcal{X}} w \sigma_2({\bf a} \cdot (\bar{\Lambda} {\bf z}) + {\bf c} \cdot {\bf z}_0 +  b) \bar{\mu}(dw,d{\bf a},d {\bf c},d b).
\end{aligned}
\end{equation}
\end{proof}

\subsection{Echo state network approximation of normalized, truncated functionals}

Consider now the setting of reservoir computing, where we aim to approximate an unknown element in our class $ \mathcal{C}$  using a finite-dimensional ELM-ESN pair, that is, a randomly generated echo state network-like state equation with a neural network as readout in which only the output layer is trained. As a first step, we consider in this section the situation when the target functional is in the class $ \mathcal{C}_N$  of finite-dimensional recurrent Barron functionals.

 More specifically, consider the echo state network
\begin{equation}
\label{eq:ESN}
\mathbf{x}_t^{\mathrm{ESN}} = \sigma_1({A}^{\mathrm{ESN}} \mathbf{x}_{t-1}^{\mathrm{ESN}} + {C}^{\mathrm{ESN}} {\bf z}_t + {\bf B}^{\mathrm{ESN}}), \quad t \in \Z_-,
\end{equation}
 with randomly generated matrices ${A}^{\mathrm{ESN}} \in \mathbb{M}_N$,  ${C}^{\mathrm{ESN}}  \in \mathbb{M}_{N, d}$, and ${\bf B}^{\mathrm{ESN}} \in \R^N$, which is used to capture the dynamics and then plugged into a random neural network readout. Thus, the element in $ \mathcal{C}_N$ is approximated by
\begin{equation}
\label{eq:ELM}
\widehat{H}({\bf z}) = \sum_{i=1}^N w^{(i)}  \sigma_2({\bf a}^{(i)} \cdot  \mathbf{x}_{-1}^{\mathrm{ESN}}({\bf z}) + {\bf c}^{(i)} \cdot {\bf z}_0 + b_i)
\end{equation}
with randomly generated coefficients ${\bf a}^{(i)}$, ${\bf c}^{(i)} $, $b_i$ valued in $\R^N$, $\R^d$, and $\R$, respectively, and where ${\bf W}= \left(w^{(1)}, \ldots, w^{(N)}\right)^{\top} \in \R^N$ is trainable. 

Let $T = \left\lceil \frac{N}{d} \right\rceil $ and define the {\bfi  Kalman controlability matrix} 
\begin{equation} \label{eq:Kdef} 
K = \pi_{N\times N}({C}^{\mathrm{ESN}}|{A}^{\mathrm{ESN}}{C}^{\mathrm{ESN}}|\cdots|({A}^{\mathrm{ESN}})^{T-2} {C}^{\mathrm{ESN}}|({A}^{\mathrm{ESN}})^{T-1} {C}^{\mathrm{ESN}}),
\end{equation} where, in case $T > \frac{N}{d}$, the map $\pi_{N\times N}$ removes the last $d T - N$ columns.

\begin{proposition} 
\label{lem:ESNtransformation}
Suppose  $H^N \colon (D_d)^{\Z_-} \to \R$ admits a normalized realization of the type
\begin{equation} \label{eq:HNcanonical} H^N({\bf z}) = \int_{\mathcal{X}^N} w \sigma_2({\bf a} \cdot (\Lambda_N {\bf z}) + {\bf c} \cdot {\bf z}_0 +  b) \mu^N(dw,d{\bf a},d {\bf c},d b), \quad  {\bf z} \in  (D_d)^{\Z_-},
\end{equation}
for some $\mu^N$ on $\mathcal{X}^N$ with $I_{\mu^N} = \int_{\mathcal{X}^N} |w| (\|{\bf a}\| + \|{\bf c}\|  +  |b|) \mu^N(dw,d{\bf a},d {\bf c},d b) < \infty$ and where $\Lambda_N \colon (D_d)^{\Z_-} \to \R^N$ satisfies $(\Lambda_N {\bf z})_{d(k-1)+j} = \lambda^{k-1} (z_{-k})_j$ for all $k \in \N$, $j=1,\ldots,d$ with $d(k-1)+j \leq N$. Assume that $\sigma_1(x)=x$, $D_d$ is bounded, and consider an arbitrary system of the type \eqref{eq:ESN} such that $\rho \left({A}^{\mathrm{ESN}}\right)<1$, and that the matrix $K$ in \eqref{eq:Kdef} is invertible. Then there exists a Borel measure $\tilde{\mu}^N$ on $\mathcal{X}^N$ such that $I_{\tilde{\mu}^N} < \infty$ and the mapping
\begin{equation} 
\label{eq:HNESN} \bar{H}^N({\bf z}) =  \int_{\mathcal{X}^N} w \sigma_2({\bf a} \cdot \mathbf{x}_{-1}^{\mathrm{ESN}}( {\bf z}) + {\bf c} \cdot {\bf z}_0 +  b) \tilde{\mu}^N(dw,d{\bf a},d {\bf c},d b), \quad  {\bf z} \in  (D_d)^{\Z_-}
\end{equation} 
satisfies for all ${\bf z} \in  (D_d)^{\Z_-}$ the bound
\begin{align}
| H^N({\bf z})-\bar{H}^N({\bf z})| &
\leq L_{\sigma_2} \left(\vertiii{{C}^{\mathrm{ESN}}} M + \|{\bf B}^{\mathrm{ESN}}\|\right) \vertiii{\left({A}^{\mathrm{ESN}}\right)^{k _0} }^{\lfloor\frac{T}{k _0}\rfloor}\vertiii{{A}^{\mathrm{ESN}}}^{k _0-1}\nonumber \\
& \times \left(\sum_{i=0}^{k _0-1}\frac{ \vertiii{\left({A}^{\mathrm{ESN}} \right) ^{i}}}{1-\vertiii{\left({A}^{\mathrm{ESN}} \right) ^{k _0}}} \right)\vertiii{K^{-\top}\Lambda}  \int_{\mathcal{X}^N} |w| \|{\bf a}\| \mu^N(dw,d{\bf a},d {\bf c},d b),\label{eq:bound1}
\end{align}
where $k _0 \in \mathbb{N} $ is the smallest integer such that $\vertiii{\left({A}^{\mathrm{ESN}}\right)^{k _0} } <1$, $M>0$ is chosen such that $\|{\bf u}\| \leq M$ for all ${\bf u} \in D_d$, and $\Lambda \in \R^{N \times N}$ is the diagonal matrix with entries $\Lambda_{d(k-1)+j,d(k-1)+j} = \lambda^{k-1}$ for all $k \in \N$, $j=1,\ldots,d$ with $d(k-1)+j \leq N$.
\end{proposition}

\begin{remark}
\normalfont
A particular case of the previous statement is when ${A}^{\mathrm{ESN}}  $ is chosen such that  $\vertiii{{A}^{\mathrm{ESN}}}<1$. In that situation, instead of the bound \eqref{eq:bound1} one obtains
\begin{equation}
\label{eq:bound1bis}
| H^N({\bf z})-\bar{H}^N({\bf z})| 
\leq 
L_{\sigma_2} \left(\vertiii{{C}^{\mathrm{ESN}}} M + \|{\bf B}^{\mathrm{ESN}}\|\right) \frac{ \vertiii{{A}^{\mathrm{ESN}}  }^T}{1-\vertiii{{A}^{\mathrm{ESN}} }} \ \vertiii{K^{-\top}\Lambda}  \int_{\mathcal{X}^N} |w| \|{\bf a}\| \mu^N(dw,d{\bf a},d {\bf c},d b).
\end{equation} 
\end{remark}

\begin{remark}
\normalfont
It can be shown (see \cite[Proposition 4.4]{RC21}) that if ${A}^{\mathrm{ESN}}$  and ${C}^{\mathrm{ESN}} $  are randomly drawn using regular random variables with values in the corresponding spaces, then the hypothesis on the invertibility of $K$ holds almost surely.
\end{remark}

\begin{proof}
First notice that by our hypotheses and \eqref{eq:auxEq10}, the equation  \eqref{eq:ESN} has a unique solution in $\ell^\infty_-(\R^N)$ given by
\begin{equation}
\label{eq:auxEq18}
\mathbf{x}_t^{\mathrm{ESN}} = \sum_{k=0}^\infty ({A}^{\mathrm{ESN}})^k ({C}^{\mathrm{ESN}} {\bf z}_{t-k} + {\bf B}^{\mathrm{ESN}}), \quad \mbox{for any $t \in \Z_-$.}
\end{equation}
Let $\mathbf{x}_t^N = \sum_{k=0}^{T-1} ({A}^{\mathrm{ESN}})^k ({C}^{\mathrm{ESN}} {\bf z}_{t-k} + {\bf B}^{\mathrm{ESN}})$. Then 
\begin{align} 
\| \mathbf{x}_t^{\mathrm{ESN}} - \mathbf{x}_t^N \| & \leq \sum_{k=T}^{\infty} \vertiii{\left({A}^{\mathrm{ESN}}\right)^k} (\vertiii{{C}^{\mathrm{ESN}}} \|{\bf z}_{t-k}\| + \|{\bf B}^{\mathrm{ESN}}\|) \nonumber \\
&\leq (\vertiii{{C}^{\mathrm{ESN}}} M + \|{\bf B}^{\mathrm{ESN}}\|) 
 \vertiii{\left({A}^{\mathrm{ESN}}\right)^T} \sum_{k=0}^{\infty} \vertiii{\left({A}^{\mathrm{ESN}}\right)^k} 
\nonumber \\
&\leq (\vertiii{{C}^{\mathrm{ESN}}} M + \|{\bf B}^{\mathrm{ESN}}\|) \vertiii{\left({A}^{\mathrm{ESN}}\right)^{k _0} }^{\lfloor\frac{T}{k _0}\rfloor}\vertiii{{A}^{\mathrm{ESN}}}^{k _0-1}\left(\sum_{i=0}^{k _0-1}\frac{ \vertiii{\left({A}^{\mathrm{ESN}} \right) ^{i}}}{1-\vertiii{\left({A}^{\mathrm{ESN}} \right) ^{k _0}}} \right).\label{eq:auxEq19} 
\end{align}
In order to obtain the last inequality in the previous expression, we decomposed $T= \lfloor\frac{T}{k _0}\rfloor k _0+r $, with $0\leq r< k _0$, which allowed us to write
\begin{equation*}
\vertiii{\left({A}^{\mathrm{ESN}}\right)^T}= 
\vertiii{\left({A}^{\mathrm{ESN}}\right)^{\lfloor\frac{T}{k _0}\rfloor k _0+r}}\leq
\vertiii{\left({A}^{\mathrm{ESN}}\right)^{ k _0}}^{\lfloor\frac{T}{k _0}\rfloor}\vertiii{\left({A}^{\mathrm{ESN}}\right)^{r}}\leq
\vertiii{\left({A}^{\mathrm{ESN}}\right)^{ k _0}}^{\lfloor\frac{T}{k _0}\rfloor} \vertiii{{A}^{\mathrm{ESN}}}^{k _0-1}.
\end{equation*}
In the last inequality we used the fact that $k _0 \in \mathbb{N} $ is the smallest integer such that $\vertiii{\left({A}^{\mathrm{ESN}}\right)^{k _0} } <1$ and hence $\vertiii{\left({A}^{\mathrm{ESN}}\right)^{r}}\geq 1 $ necessarily if $r<k_0 $.

Denote by $\tilde{{\bf z}} \in \R^N$ the vector with entries $\tilde{\bf z}_{d(k-1)+j} = (z_{-k})_j$ and by $\Lambda \in \R^{N \times N}$ the diagonal matrix  such that  $\Lambda_{d(k-1)+j,d(k-1)+j} = \lambda^{k-1}$, for all $k \in \N$, $j=1,\ldots,d$ with $d(k-1)+j \leq N$. Then $\mathbf{x}_{-1}^N = \sum_{k=1}^{T} ({A}^{\mathrm{ESN}})^{k-1} ({C}^{\mathrm{ESN}} {\bf z}_{-k} + {\bf B}^{\mathrm{ESN}}) = K \tilde{{\bf z}} + \tilde{\bf B}$, where $\tilde{\bf B} = \sum_{k=1}^{T} ({A}^{\mathrm{ESN}})^{k-1} {\bf B}^{\mathrm{ESN}} $. Let $\tilde{\phi} \colon \mathcal{X}^N \to \mathcal{X}^N$ be given as $\tilde{\phi}(w,{\bf a},{\bf c},b)=(w,K^{-\top}\Lambda{\bf a},{\bf c},-(K^{-\top}\Lambda{\bf a}) \cdot \tilde{\bf B} +b)$ and denote by $\tilde{\mu}^N = \mu^N \circ \tilde{\phi}^{-1}$ the pushforward measure of $\mu^N$ under $\tilde{\phi}$. Then the change-of-variables formula yields that
\[\begin{aligned}
I_{\tilde{\mu}^N} & = \int_{\mathcal{X}^N} |w| (\|K^{-\top}\Lambda{\bf a}\| + \|{\bf c}\|  +  |-(K^{-\top}\Lambda{\bf a}) \cdot \tilde{\bf B} +b|) \mu^N(dw,d{\bf a},d {\bf c},d b) 
\\ & \leq \max(\vertiii{K^{-\top}\Lambda},1)(1+ \|\tilde{\bf B}\|) I_{\mu^N} < \infty
\end{aligned}\]
and 
\begin{equation} \label{eq:auxEq17} 
\begin{aligned}
H^N({\bf z}) & = \int_{\mathcal{X}^N} w \sigma_2((\Lambda{\bf a}) \cdot \tilde{\bf z} + {\bf c} \cdot {\bf z}_0 +  b) \mu^N(dw,d{\bf a},d {\bf c},d b)
\\ 
& = \int_{\mathcal{X}^N} w \sigma_2((K^{-\top}\Lambda{\bf a}) \cdot K \tilde{\bf z} + {\bf c} \cdot {\bf z}_0 +  b) \mu^N(dw,d{\bf a},d {\bf c},d b)
\\ 
& = \int_{\mathcal{X}^N} w \sigma_2((K^{-\top}\Lambda{\bf a}) \cdot \mathbf{x}_{-1}^N - (K^{-\top}\Lambda{\bf a}) \cdot \tilde{\bf B}  + {\bf c} \cdot {\bf z}_0 +  b) \mu^N(dw,d{\bf a},d {\bf c},d b)
\\ 
& = \int_{\mathcal{X}^N} w \sigma_2({\bf a} \cdot \mathbf{x}_{-1}^N  + {\bf c} \cdot {\bf z}_0 +  b) \tilde{\mu}^N(dw,d{\bf a},d {\bf c},d b).
\end{aligned}
\end{equation}
Combining \eqref{eq:auxEq17} and \eqref{eq:auxEq19} then yields
\begin{equation}
\label{eq:auxEq20}
\begin{aligned}
| H^N({\bf z})-\bar{H}^N({\bf z})| & \leq  \int_{\mathcal{X}^N} w |\sigma_2({\bf a} \cdot \mathbf{x}_{-1}^N  + {\bf c} \cdot {\bf z}_0 +  b) - \sigma_2({\bf a} \cdot \mathbf{x}_{-1}^{\mathrm{ESN}}( {\bf z}) + {\bf c} \cdot {\bf z}_0 +  b)| \tilde{\mu}^N(dw,d{\bf a},d {\bf c},d b)
\\ & \leq 
\int_{\mathcal{X}^N} w L_{\sigma_2} |{\bf a} \cdot (\mathbf{x}_{-1}^N - \mathbf{x}_{-1}^{\mathrm{ESN}}) | \tilde{\mu}^N(dw,d{\bf a},d {\bf c},d b)
\\ & \leq 
\|\mathbf{x}_{-1}^N - \mathbf{x}_{-1}^{\mathrm{ESN}} \| L_{\sigma_2} \int_{\mathcal{X}^N} w  \|K^{-\top}\Lambda{\bf a}\|  \mu^N(dw,d{\bf a},d {\bf c},d b)
\\ & \leq 
L_{\sigma_2} (\vertiii{{C}^{\mathrm{ESN}}} M + \|{\bf B}^{\mathrm{ESN}}\|) \vertiii{\left({A}^{\mathrm{ESN}}\right)^{k _0} }^{\lfloor\frac{T}{k _0}\rfloor}\vertiii{{A}^{\mathrm{ESN}}}^{k _0-1}
\\ & \quad \cdot
\left(\sum_{i=0}^{k _0-1}\frac{ \vertiii{\left({A}^{\mathrm{ESN}} \right) ^{i}}}{1-\vertiii{\left({A}^{\mathrm{ESN}} \right) ^{k _0}}} \right) \vertiii{K^{-\top}\Lambda} I_{\mu^N}.
\end{aligned}
\end{equation} 
\end{proof}

\subsection{Main approximation result}

This section contains our main approximation result, in which we extend the possibility of approximating the elements in the class $ \mathcal{C}_N$ of finite-dimensional recurrent Barron functionals, using finite-dimensional ELM-ESN pairs, to the full class $\mathcal{C} $ of recurrent Barron functionals.

In this case, we shall assume that the random variables used in the construction of the ELM layer \eqref{eq:ELM}, that is, the initialization  $w^{(i)}$ for the readout weight and the hidden weights  ${\bf a}^{(i)}$, ${\bf c}^{(i)} $, $b_i$,  valued in $\mathbb{R}  $, $\R^N$, $\R^d$, and $\R$, respectively, are such that $(w^{(1)},{\bf a}^{(1)},{\bf c}^{(1)},b_1), \ldots,(w^{(N)},{\bf a}^{(N)},{\bf c}^{(N)},b_N)$  are independent and identically distributed (IID); denote by $\nu$ their common distribution. 

Let $\Lambda \in \R^{N \times N}$ be the diagonal matrix introduced in Proposition~\ref{lem:ESNtransformation}, let $\pi \colon \ell^p \to \R^N $ be the projection map $\pi({\bf a})=(a_i)_{i=1,\ldots,N}$ and for given Borel measure $\mu$ on $\mathcal{X}$, ${\bf B} \in \ell^q$, bounded linear maps $C \colon \R^d \to \ell^q$, $A \colon \ell^q \to \ell^q$ with $\rho{(A)}<1$ let  ${\mu}_{A,B,C} = \mu \circ \psi_{A,B,C}^{-1}$, where $\psi_{A,B,C} \colon \mathcal{X} \to \mathcal{X}^N$ is the map
\[
\psi_{A,B,C}(w,{\bf a},{\bf c},b) = \Bigg(w,K^{-\top}\Lambda\pi(\bar{\mathcal{L}}({\bf a})),{\bf c},-(K^{-\top}\Lambda\pi(\bar{\mathcal{L}}({\bf a}))) \cdot \Bigg[\sum_{k=1}^{T} ({A}^{\mathrm{ESN}})^{k-1} {\bf B}^{\mathrm{ESN}}\Bigg] +{\bf a} \cdot \left[\sum_{k=0}^\infty A^k {\bf B}\right]+ b\Bigg)
\]  
with $[\bar{\mathcal{L}}({\bf a})]_{d(k-1)+j} = [(\tilde{A}^{k-1} C)^* {\bf a}]_j $ for $k \in \N$, $j=1,\ldots,d$ .
Furthermore, let 
\[
I_{\mu,p}^{(2)} = \left(\int_{\mathcal{X}} w^2  \left(\|{\bf a}\|_p^2  + \|{\bf c}\|^2  + |b|^2 + 1 \right)  \mu(dw,d{\bf a},d {\bf c},d b) \right)^{1/2}.
\] 

Using this notation, we shall now formulate a result that shows that the elements in $H \in \mathcal{C}$ that admit a representation in which the state equation \eqref{eq:systemi} is linear and has the unique solution property, can be approximated arbitrarily well by randomly generated ESN-ELM pairs in which only the output layer of the ELM is trained. We provide an approximation bound where the dependence on the dimensions $N$ of the ESN state space and $d  $ of the input is explicitly spelled out.

\begin{theorem} 
\label{thm:approx}
In this statement assume that $\sigma_1(x)=x$, $p \in (1,\infty)$, and that the input set $D_d$ is bounded. Let $H \in \mathcal{C}$ be such that it has a realization of the type \eqref{eq:Hrepres},  with  ${\bf B} \in \ell^q$, bounded linear maps $C \colon \R^d \to \ell^q$, $A \colon \ell^q \to \ell^q$ satisfying $\vertiii{A}<1$ and $I_{\mu,p}^{(2)} < \infty$. Using the notation introduced before this statement, suppose that $ \mu_{A,B,C} \ll \nu $ and that ${\displaystyle \frac{d \mu_{A,B,C}}{d \nu}}$ is bounded. Let $\lambda \in \left(\vertiii{A},1\right)$. Consider now a randomly constructed ESN-ELM pair such that $\vertiii{{ A}^{\mathrm{ESN}} }<1$ and for which the controllability matrix $K$ in \eqref{eq:Kdef} is invertible. 

Then, there exists a measurable function $f \colon (\mathcal{X}^N)^N \to \R^N$ such that the ESN-ELM pair $\widehat{H}$ with the same parameters and new readout ${\bf W} = f((w^{(i)},{\bf a}^{(i)},{\bf c}^{(i)},b_i)_{i=1,\ldots,N}) $ satisfies, for any ${\bf z} \in \mathcal{I}_d$, the approximation error bound
\begin{equation}
\begin{aligned}
\label{eq:approxBound}
\E[|H({\bf z}) - \widehat{H}({\bf z})|^2]^{1/2} & \leq  C_{H,\mathrm{ESN}} \left[  \lambda^{\frac{N}{d}}  + \vertiii{{A}^{\mathrm{ESN}} }^T  + \frac{1}{N^{\frac{1}{2}}}  \left\|\frac{d \mu_{A,B,C}}{d \nu}\right\|_{\infty}^{\frac{1}{2}}\right]
\end{aligned}
\end{equation}	
with 
\begin{multline} 
\label{eq:CHESN}
C_{H,\mathrm{ESN}} = \tilde{c}_1 \max\left( \frac{\vertiii{C}}{\left(1-\vertiii{\lambda^{-1}{A}}^{p}\right)^{1/p}},\frac{\|{\bf B}\|_q}{1-\vertiii{A}},1\right) \\
\times  \max\left(I_{\mu,p},I_{\mu,p}^{(2)}\right)\cdot  \left(1+\frac{\vertiii{{C}^{\mathrm{ESN}}}  + \|{\bf B}^{\mathrm{ESN}}\|}{1-\vertiii{{A}^{\mathrm{ESN}} }} \vertiii{ K^{-\top}\Lambda} \right) 
\end{multline}
for $\tilde{c}_1$ only depending on $d$, $\sigma_2$, $p$, $\mathrm{diam}(D_d)$ and $\lambda$ (see \eqref{eq:c1tilde}). 

\end{theorem}

\begin{proof}
Let $\bar{\Lambda} \colon (D_d)^{\Z_-} \to \ell^q$, $(\bar{\Lambda} {\bf z})_{d(k-1)+j} = \lambda^{k-1} (z_{-k})_j$ for $k \in \N$, $j=1,\ldots,d$. Then Proposition~\ref{lem:canonical} implies that there exists a (Borel) probability measure $\bar{\mu}$ on $\mathcal{X}$ satisfying $I_{\bar{\mu},p} < \infty$
and such that the normalized representation \begin{equation} \label{eq:auxEq21} H({\bf z}) = \int_{\mathcal{X}} w \sigma_2({\bf a} \cdot (\bar{\Lambda} {\bf z}) + {\bf c} \cdot {\bf z}_0 +  b) \bar{\mu}(dw,d{\bf a},d {\bf c},d b), \quad  {\bf z} \in  \mathcal{I}_d,
\end{equation} holds. Notice that from the proof of Proposition~\ref{lem:linear} we obtain that
$(\bar{\Lambda} {\bf z})_{d(k-1)+j} = \lambda^{k-1} (z_{-k})_j = \bar{{x}}_{-1,d(k-1)+j}$, where $\bar{\mathbf{x}}$ is the unique solution to 
  \begin{equation}\label{eq:auxEq25}
\bar{\mathbf{x}}_t = \sigma_1(\bar{A} \bar{\mathbf{x}}_{t-1} + \bar{C} {\bf z}_t), \quad t \in \Z_-,
\end{equation}
 with  $\bar{A} \colon \ell^q \to \ell^q$, $\bar{C} \colon \R^d \to \ell^q$ given by $(\bar{A} {\bf x})_{i} = \mathbbm{1}_{\{i> d\}} \lambda x_{i-d}$, $(\bar{C} {\bf z})_{i} = \mathbbm{1}_{\{i\leq d\}} {z}_i$ for ${\bf x} \in \ell^q$, ${\bf z} \in \R^d$, $i \in \N$. 
Let $\hat{A} \in \R^{N \times N}$, $\hat{C}  \in \R^{N \times d}$ by given as $\hat{A}_{ij} = (\bar{A}\boldsymbol{\epsilon}^j)_i = \mathbbm{1}_{\{i> d\}} \lambda \delta_{i-d,j}$, $\hat{C}_{ik} = (\bar{C}{\bf e}^k)_i = \mathbbm{1}_{\{i\leq d\}} \delta_{i,k}$ for  $i,j=1,\ldots,N$, $k=1,\ldots,d$. Proposition~\ref{prop:truncation} then implies that for each ${\bf z} \in  \mathcal{I}_d$ the system 
\begin{equation}\label{eq:auxEq22}
\mathbf{x}_t^N = \sigma_1(\hat{A} \mathbf{x}_{t-1}^N + \hat{C} {\bf z}_t), \quad t \in \Z_-,
\end{equation}
admits a unique solution $(\mathbf{x}_t^N)_{t \in \Z_-} \in \ell^\infty_-(\R^N)$ and  the functional $H^N \colon \mathcal{I}_d \to \R$,
\begin{equation} \label{eq:auxEq23}\begin{aligned} H^N({\bf z}) = \int_{\mathcal{X}} w \sigma_2({\bf a} \cdot \iota(\mathbf{x}_{-1}^N({\bf z})) + {\bf c} \cdot {\bf z}_0 +  b) \bar{\mu}(dw,d{\bf a},d {\bf c},d b), \quad  {\bf z} \in  \mathcal{I}_d,
\end{aligned}	\end{equation}
satisfies $H^N \in \mathcal{C}_N$ and that 
\begin{equation}
	\label{eq:auxEq24}
	|H({\bf z})-H^N({\bf z})| \leq  C_{\mathrm{fin}} \sum_{l=0}^\infty \vertiii{\bar{A}}^l \left(\sum_{i=N+1}^\infty |\bar{{x}}_{-1-l,i}|^q\right)^{1/q}
\end{equation}
for all ${\bf z} \in  \mathcal{I}_d$ and by \eqref{eq:auxEq15}
\[ \begin{aligned} C_{\mathrm{fin}} & = L_{\sigma_2} \int_{\mathcal{X}} |w| \| {\bf a} \|_p   \bar{\mu}(dw,d{\bf a},d {\bf c},d b) = L_{\sigma_2} \int_{\mathcal{X}} |w| \|\bar{\mathcal{L}}({\bf a}) \|_p  {\mu}(dw,d{\bf a},d {\bf c},d b) \leq c_0 L_{\sigma_2} I_{\mu,p}.
\end{aligned}
\]
Choose $M>0$ such that $\|{\bf u}\| \leq M$ for all ${\bf u} \in D_d$. Furthermore, note that $\vertiii{\bar{A}} \leq \lambda$ and $\bar{{x}}_{-1-l,(k-1)d+j}= \lambda^{k-1}[z_{-l-k}]_j$. Hence, using $\|\cdot\|_q^q \leq \max(d^{1-\frac{q}{2}},1) \|\cdot\|^q$ and $\|{\bf u}\| \leq M$ for all ${\bf u} \in D_d$ we obtain from \eqref{eq:auxEq24} 
\begin{equation}
\label{eq:auxEq27}
\begin{aligned}
|H({\bf z})-H^N({\bf z})| &  \leq  c_0 L_{\sigma_2} I_{\mu,p} \sum_{l=0}^\infty (\vertiii{\bar{A}})^l \left(\sum_{k=\lfloor \frac{N}{d}\rfloor+1}^\infty [\lambda^{k-1}]^q \sum_{j=1}^d |[z_{-l-k}]_j|^q\right)^{1/q}
\\ & \leq  c_0 L_{\sigma_2} I_{\mu,p} \sum_{l=0}^\infty (\vertiii{\bar{A}})^l \left(\sum_{k=\lfloor \frac{N}{d}\rfloor+1}^\infty[\lambda^{k-1}]^q \max(d^{1-\frac{q}{2}},1) M^q \right)^{1/q}
\\ & \leq  c_0 L_{\sigma_2} I_{\mu,p} M \max(d^{\frac{1}{q}-\frac{1}{2}},1) \frac{\lambda^{\lfloor \frac{N}{d}\rfloor}}{(1-\lambda)(1-\lambda^{q})^{1/q}} .
\end{aligned}
\end{equation}
 From  \eqref{eq:auxEq22} we obtain for $i=1,\ldots,N$, $t \in \Z_-$ that
$({x}_t^N)_i = \mathbbm{1}_{\{i> d\}} \lambda  ({x}_{t-1}^N)_{i-d} + \mathbbm{1}_{\{i\leq d\}} ({z}_t)_i $. Therefore, $({x}_t^N)_{d(k-1)+j} =  \lambda^{k-1} (z_{t-k+1})_j$ for all $k \in \N$, $j=1,\ldots,d$ with $d(k-1)+j \leq N$. In particular, letting $\Lambda_N \colon (D_d)^{\Z_-} \to \R^N$ satisfy $(\Lambda_N {\bf z})_{d(k-1)+j} = \lambda^{k-1} (z_{-k})_j$ for all $k \in \N$, $j=1,\ldots,d$ with $d(k-1)+j \leq N$ we obtain $({x}_{-1}^N)_{d(k-1)+j} = \lambda^{k-1} (z_{-k})_j = (\Lambda_N {\bf z})_{d(k-1)+j}$. Let  $\phi \colon \mathcal{X} \to \mathcal{X}^N $ be the projection map $\phi(w,{\bf a},{\bf c},b)=(w,\pi({\bf a}),{\bf c},b)$, with $\pi({\bf a})=(a_i)_{i=1,\ldots,N}$, and denote by $\mu^N = \bar{\mu} \circ \phi^{-1}$ the pushforward measure of $\bar{\mu}$ under $\phi$. Then $I_{\mu^N} < \infty$ and from \eqref{eq:auxEq23} we get
\begin{equation} \begin{aligned} H^N({\bf z}) =
\int_{\mathcal{X}^N} w \sigma_2({\bf a} \cdot (\Lambda_N {\bf z}) + {\bf c} \cdot {\bf z}_0 +  b) \mu^N(dw,d{\bf a},d {\bf c},d b), \quad  {\bf z} \in  \mathcal{I}_d.
\end{aligned}
\end{equation}
Proposition~\ref{lem:ESNtransformation} hence implies that there exists a Borel measure $\tilde{\mu}^N$ on $\mathcal{X}^N$ such that $I_{\tilde{\mu}^N} < \infty$ and the mapping
\begin{equation} \label{eq:auxEq33} \bar{H}^N({\bf z}) =  \int_{\mathcal{X}^N} w \sigma_2({\bf a} \cdot \mathbf{x}_{-1}^{\mathrm{ESN}}( {\bf z}) + {\bf c} \cdot {\bf z}_0 +  b) \tilde{\mu}^N(dw,d{\bf a},d {\bf c},d b), \quad  {\bf z} \in  (D_d)^{\Z_-}
\end{equation} 
satisfies for all ${\bf z} \in  \mathcal{I}_d$ the bound \eqref{eq:bound1bis}, that is,
\begin{equation}\label{eq:auxEq26} 
| H^N({\bf z})-\bar{H}^N({\bf z})| \leq L_{\sigma_2} (\vertiii{{C}^{\mathrm{ESN}}} M + \|{\bf B}^{\mathrm{ESN}}\|) \frac{\vertiii{{A}^{\mathrm{ESN}} }^T}{1-\vertiii{{A}^{\mathrm{ESN}} }} \vertiii{K^{-\top}\Lambda}  \int_{\mathcal{X}^N} |w| \|{\bf a}\| \mu^N(dw,d{\bf a},d {\bf c},d b),
\end{equation} 
where $\Lambda \in \R^{N \times N}$ is the diagonal matrix with entries $\Lambda_{d(k-1)+j,d(k-1)+j} = \lambda^{k-1}$ for all $k \in \N$, $j=1,\ldots,d$ with $d(k-1)+j \leq N$. Furthermore, $\| \cdot \| \leq \max(1,d^{\frac{1}{2}-\frac{1}{p}})\| \cdot \|_p$ and \eqref{eq:auxEq15} imply
\begin{equation} \label{eq:auxEq28}
\begin{aligned}
  \int_{\mathcal{X}^N} |w| \|{\bf a}\| \mu^N(dw,d{\bf a},d {\bf c},d b) & = \int_{\mathcal{X}} |w| \|\pi(\bar{\mathcal{L}}({\bf a}))\| \mu(dw,d{\bf a},d {\bf c},d b)
 \\ & \leq \max(1,d^{\frac{1}{2}-\frac{1}{p}}) \int_{\mathcal{X}} |w| \|\bar{\mathcal{L}}({\bf a})\|_p \mu(dw,d{\bf a},d {\bf c},d b)
  \\ & \leq \max(1,d^{\frac{1}{2}-\frac{1}{p}}) c_0 I_{\mu,p}.
\end{aligned}
\end{equation}

Note that the measure $\tilde{\mu}^N$ in \eqref{eq:auxEq33} is given by $\tilde{\mu}^N = \mu^N \circ \tilde{\phi}^{-1}$ with $\tilde{\phi}$ as in the proof of Proposition~\ref{lem:ESNtransformation} and $\bar{\mu} = \mu \circ \bar{\phi}^{-1}$ with $\bar{\phi}$ as in the proof of Proposition~\ref{lem:canonical}. Then we verify that $\tilde{\mu}^N = \mu_{A,B,C}$ and the change-of-variables theorem shows that 
\begin{equation} 
\label{eq:auxEq32} 
\begin{aligned}
\int_{\mathcal{X}^N}& f(w,{\bf a},{\bf c},b) \tilde{\mu}^N(dw,d{\bf a},d {\bf c},d b)  = \int_{\mathcal{X}^N} f(w,K^{-\top}\Lambda{\bf a},{\bf c},-(K^{-\top}\Lambda{\bf a}) \cdot \tilde{\bf B} +b) \mu^N(dw,d{\bf a},d {\bf c},d b)
\\ & = \int_{\mathcal{X}} f(w,K^{-\top}\Lambda\pi({\bf a}),{\bf c},-(K^{-\top}\Lambda\pi({\bf a})) \cdot \tilde{\bf B} +b) \bar{\mu}(dw,d{\bf a},d {\bf c},d b)
\\ & = \int_{\mathcal{X}} f(w,K^{-\top}\Lambda\pi(\bar{\mathcal{L}}({\bf a})),{\bf c},-(K^{-\top}\Lambda\pi(\bar{\mathcal{L}}({\bf a}))) \cdot \tilde{\bf B} +{\bf a} \cdot \bar{\bf B}_0+ b) \mu(dw,d{\bf a},d {\bf c},d b)
\end{aligned}
\end{equation}
for any function $f$ for which the integrand in the last line is integrable with respect to $\mu$ and where $\tilde{\bf B} = \sum_{k=1}^{T} ({A}^{\mathrm{ESN}})^{k-1} {\bf B}^{\mathrm{ESN}} $,  $\bar{\bf B}_0 = \sum_{k=0}^\infty A^k {\bf B} \in \ell^q$ and $\bar{\mathcal{L}} \colon \ell^p \to \ell^p$ is linear and satisfies \eqref{eq:auxEq15}.

Set $W_i = \dfrac{w^{(i)}}{N}  \dfrac{d \tilde{\mu}^N}{d \nu}(w^{(i)},{\bf a}^{(i)},{\bf c}^{(i)},b_{i})$ and notice that 
\[ \begin{aligned}
 \E\left[\sum_{i=1}^N W_i \sigma_2({\bf a}^{(i)} \cdot  \mathbf{x}_{-1}^{\mathrm{ESN}} + {\bf c}^{(i)} \cdot {\bf z}_0 + b_i)\right] & = \int_{\mathcal{X}^N} w \frac{d \tilde{\mu}^N}{d \nu}(w,{\bf a},{\bf c},b) \sigma_2({\bf a} \cdot  \mathbf{x}_{-1}^{\mathrm{ESN}} + {\bf c} \cdot {\bf z}_0 + b) \nu(dw,d{\bf a},d {\bf c},d b)
\\ & = \int_{\mathcal{X}^N} w  \sigma_2({\bf a} \cdot  \mathbf{x}_{-1}^{\mathrm{ESN}} + {\bf c} \cdot {\bf z}_0 + b) \tilde{\mu}^N(dw,d{\bf a},d {\bf c},d b)= \bar{H}^N({\bf z}).
\end{aligned}
\]
Therefore, 
\begin{equation} \label{eq:auxEq29} \begin{aligned}
\E[|\bar{H}^N({\bf z}) - \widehat{H}({\bf z})|^2] & = \E[|\bar{H}^N({\bf z}) - \sum_{i=1}^N W_i \sigma_2({\bf a}^{(i)} \cdot  \mathbf{x}_{-1}^{\mathrm{ESN}} + {\bf c}^{(i)} \cdot {\bf z}_0 + b_i)|^2]
\\ & = \mathrm{Var}\left(\sum_{i=1}^N W_i \sigma_2({\bf a}^{(i)} \cdot  \mathbf{x}_{-1}^{\mathrm{ESN}} + {\bf c}^{(i)} \cdot {\bf z}_0 + b_i)\right)
\\ & = \sum_{i=1}^N \mathrm{Var}( W_i \sigma_2({\bf a}^{(i)} \cdot  \mathbf{x}_{-1}^{\mathrm{ESN}} + {\bf c}^{(i)} \cdot {\bf z}_0 + b_i))
\\ & \leq  N \E[W_1^2 \sigma_2({\bf a}^{(1)} \cdot  \mathbf{x}_{-1}^{\mathrm{ESN}} + {\bf c}^{(1)} \cdot {\bf z}_0 + b_1)^2].
\end{aligned}
\end{equation}
As noted above,  $\tilde{\mu}^N = \mu_{A,B,C}$ and by assumption ${\displaystyle \frac{d \mu_{A,B,C}}{d \nu}}$ is bounded, hence 
{\scriptsize
\begin{align}
\E&[W_1^2 \sigma_2({\bf a}^{(1)} \cdot  \mathbf{x}_{-1}^{\mathrm{ESN}} + {\bf c}^{(1)} \cdot {\bf z}_0 + b_1)^2]  \nonumber  \\&= \int_{\mathcal{X}^N} \frac{w^2}{N^2}  \frac{d \tilde{\mu}^N}{d \nu}(w,{\bf a},{\bf c},b) \sigma_2({\bf a} \cdot  \mathbf{x}_{-1}^{\mathrm{ESN}} + {\bf c} \cdot {\bf z}_0 + b)^2  \tilde{\mu}^N(dw,d{\bf a},d {\bf c},d b)\nonumber 
\\ & \leq 2 \left\|\frac{d \mu_{A,B,C}}{d \nu}\right\|_{\infty} \int_{\mathcal{X}^N} \frac{w^2}{N^2}  [L_{\sigma_2}^2|{\bf a} \cdot  \mathbf{x}_{-1}^{\mathrm{ESN}} + {\bf c} \cdot {\bf z}_0 + b|^2 + |\sigma_2(0)|^2]  \tilde{\mu}^N(dw,d{\bf a},d {\bf c},d b) \nonumber 
\\ & = 2 \left\|\frac{d \mu_{A,B,C}}{d \nu}\right\|_{\infty} \int_{\mathcal{X}} \frac{w^2}{N^2}  [L_{\sigma_2}^2|K^{-\top}\Lambda\pi(\bar{\mathcal{L}}({\bf a})) \cdot  \mathbf{x}_{-1}^{\mathrm{ESN}} + {\bf c} \cdot {\bf z}_0  -(K^{-\top}\Lambda\pi(\bar{\mathcal{L}}({\bf a}))) \cdot \tilde{\bf B} +{\bf a} \cdot \bar{\bf B}_0+ b |^2 + |\sigma_2(0)|^2]  
\mu(dw,d{\bf a},d {\bf c},d b)\nonumber 
\\ & \leq 2 \left\|\frac{d \mu_{A,B,C}}{d \nu}\right\|_{\infty} \int_{\mathcal{X}} \frac{w^2}{N^2}  [L_{\sigma_2}^2|\|\pi(\bar{\mathcal{L}}({\bf a}))\| \| \Lambda K^{-1}(\mathbf{x}_{-1}^{\mathrm{ESN}}-\tilde{\bf B})\| + \|{\bf c}\| \| {\bf z}_0 \| + \|{\bf a}\|_p \|\bar{\bf B}_0\|_q+ |b| |^2 + |\sigma_2(0)|^2]  
\mu(dw,d{\bf a},d {\bf c},d b)\nonumber 
\\ & \leq  \frac{c_1}{N^2} \left\|\frac{d \mu_{A,B,C}}{d \nu}\right\|_{\infty} \max(c_0^2,\|\bar{\bf B}_0\|_q^2,1) (1+\vertiii{ \Lambda K^{-1}}^2 \| \mathbf{x}_{-1}^{\mathrm{ESN}}-\tilde{\bf B}\|^2)\int_{\mathcal{X}} w^2  [\|{\bf a}\|_p^2  + \|{\bf c}\|^2  + |b|^2 + 1 ]  
\mu(dw,d{\bf a},d {\bf c},d b)\nonumber 
\end{align}}
where we used \eqref{eq:auxEq15},  $\| \cdot \| \leq \max(1,d^{\frac{1}{2}-\frac{1}{p}})\| \cdot \|_p$,  $c_0 = d^{1/(2p)} \vertiii{C} (1-\vertiii{\tilde{A}}^{p})^{-1/p}$ and set
\begin{equation} 
\label{eq:Cmu} c_1 = 8 \max(1,d^{\frac{1}{2}-\frac{1}{p}})^2  \max(L_{\sigma_2},|\sigma_2(0)|)^2 \max(M^2,1).
\end{equation}
Combining this with \eqref{eq:auxEq29}, \eqref{eq:auxEq27}, \eqref{eq:auxEq26} and \eqref{eq:auxEq28} then yields
{\scriptsize
\begin{align} 
 & \E[|H({\bf z}) - \widehat{H}({\bf z})|^2]^{1/2} \nonumber 
 \\ &  \leq \E[|H({\bf z}) - H^N({\bf z})|^2]^{1/2} + \E[|H^N({\bf z}) - \bar{H}^N({\bf z})|^2]^{1/2} + \E[|\bar{H}^N({\bf z}) - \widehat{H}({\bf z})|^2]^{1/2}\nonumber 
\\ & \leq  c_0 L_{\sigma_2} I_{\mu,p} M \max(1,d^{\frac{1}{q}-\frac{1}{2}}) \frac{\lambda^{\lfloor \frac{N}{d}\rfloor}}{(1-\lambda)(1-\lambda^{q})^{1/q}}  + L_{\sigma_2} \left(\vertiii{{C}^{\mathrm{ESN}}} M + \|{\bf B}^{\mathrm{ESN}}\|\right) \frac{\vertiii{{ A}^{\mathrm{ESN}} }^T}{1-\vertiii{{A}^{\mathrm{ESN}} }} \vertiii{K^{-\top}\Lambda}  \max\left(1,d^{\frac{1}{2}-\frac{1}{p}}\right) c_0 I_{\mu,p} \nonumber 
\\ & \quad \quad  + \frac{c_1^{1/2}}{N^{1/2}}  \left\|\frac{d \mu_{A,B,C}}{d \nu}\right\|_{\infty}^{1/2} \max(c_0,\|\bar{\bf B}_0\|_q,1) (1+\vertiii{ \Lambda K^{-1}} \|\mathbf{x}_{-1}^{\mathrm{ESN}}-\tilde{\bf B}\|)  I_{\mu,p}^{(2)}\nonumber 
\\ & \leq \tilde{c}_1 \max\left( \frac{\vertiii{C}}{\left(1-\vertiii{\lambda^{-1}{A}}^{p}\right)^{1/p}},\frac{\|{\bf B}\|_q}{1-\vertiii{A}},1\right) [ I_{\mu,p} \lambda^{\frac{N}{d}}  + \left(\vertiii{{C}^{\mathrm{ESN}}}  + \|{\bf B}^{\mathrm{ESN}}\|\right) \frac{\vertiii{{ A}^{\mathrm{ESN}} }^T}{1-\vertiii{{ A}^{\mathrm{ESN}} }} \vertiii{K^{-\top}\Lambda}   I_{\mu,p}  \nonumber 
\\ &  \quad \quad  + \frac{1}{N^{\frac{1}{2}}}  \left\|\frac{d \mu_{A,B,C}}{d \nu}\right\|_{\infty}^{\frac{1}{2}}  (1+\vertiii{ \Lambda K^{-1}} \left(\vertiii{{C}^{\mathrm{ESN}}}  + \|{\bf B}^{\mathrm{ESN}}\|\right) \frac{1}{1-\vertiii{{ A}^{\mathrm{ESN}}}})  I_{\mu,p}^{(2)}]\nonumber 
\\ & \leq \tilde{c}_1 \max\left( \frac{\vertiii{C}}{\left(1-\vertiii{\lambda^{-1}{A}}^{p}\right)^{1/p}},\frac{\|{\bf B}\|_q}{1-\vertiii{A}},1\right) \max(I_{\mu,p},I_{\mu,p}^{(2)})(1+\frac{\vertiii{{ C}^{\mathrm{ESN}}}  + \|{\bf B}^{\mathrm{ESN}}\|}{1-\vertiii{{A}^{\mathrm{ESN}} }} \vertiii{K^{-\top}\Lambda} ) \nonumber 
\\ & \quad \quad \times\left [  \lambda^{\frac{N}{d}}  + \vertiii{{A}^{\mathrm{ESN}} }^T  + \frac{1}{N^{\frac{1}{2}}}  \left\|\frac{d \mu_{A,B,C}}{d \nu}\right\|_{\infty}^{\frac{1}{2}}\right]\label{eq:auxEq35}
\end{align}}
with 
\begin{equation} \begin{aligned}
\label{eq:c1tilde}
\tilde{c}_1 & 
=  
8^{\frac{1}{2}}  \max(L_{\sigma_2},|\sigma_2(0)|,1) \max(1,d^{\frac{1}{2}-\frac{1}{p}}) d^{\frac{1}{2p}} \max(1,d^{\frac{1}{q}-\frac{1}{2}})(1-\lambda)^{-1}(1-\lambda^{q})^{-\frac{1}{q}} \lambda^{-1}\max(M,1)^2.
\end{aligned}
\end{equation}  
\end{proof}

\begin{remark}
\normalfont
A bound similar to the one in \eqref{eq:approxBound} can be obtained if the ESN-ELM used as approximant satisfies the more general condition $\rho({A}^{{\rm ESN}})<1 $. The theorem is proved in that case by replacing in \eqref{eq:auxEq26} the use of the inequality \eqref{eq:bound1bis} by its more general version \eqref{eq:bound1}.
\end{remark}

\begin{corollary}
\label{cor:BoundForSome}
As in Theorem \ref{thm:approx}, suppose the recurrent activation function is $\sigma_1(x)=x$, $p \in (1,\infty)$ and the input set $D_d$ is bounded. Let $H \in \mathcal{C}$ be such that it has a realization of the type \eqref{eq:Hrepres}  with measure $\mu$ and bounded linear maps $C \colon \R^d \to \ell^q$, $A \colon \ell^q \to \ell^q$ satisfying $\vertiii{A}<1$ and $I_{\mu,p}^{(2)} < \infty$. Let $\lambda \in \left(\vertiii{A},1\right)$. Consider now a randomly constructed ESN-ELM pair such that $\vertiii{{ A}^{\mathrm{ESN}} }<1$ and for which the controllability matrix $K$ in \eqref{eq:Kdef} is invertible. 
Then there exists a distribution $\nu$ for the readout hidden layer weights and a readout ${\bf W}$ (an $\R^N$-valued random vector) such that the ESN-ELM $\widehat{H}$ satisfies, for any ${\bf z} \in \mathcal{I}_d$, the approximation error bound
\begin{equation}
\begin{aligned}
\label{eq:approxBoundCor}
\E[|H({\bf z}) - \widehat{H}({\bf z})|^2]^{1/2} & \leq  C_{H,\mathrm{ESN}} \left[  \lambda^{\frac{N}{d}}  + \vertiii{{A}^{\mathrm{ESN}} }^\frac{N}{d}  + \frac{1}{N^{\frac{1}{2}}}\right]
\end{aligned}
\end{equation}	
with $C_{H,\mathrm{ESN}}$ as in \eqref{eq:CHESN}.
\end{corollary}

\begin{proof}
This follows by noting $ \vertiii{{ A}^{\mathrm{ESN}} }^T \leq \vertiii{{A}^{\mathrm{ESN}} }^{\frac{N}{d}}$ due to $T \geq \frac{N}{d}$ and choosing $\nu = \mu_{A,B,C}$, so that $\dfrac{d \mu_{A,B,C}}{d \nu} = 1$ and Theorem~\ref{thm:approx} yields \eqref{eq:approxBoundCor}. 
\end{proof}

\medskip

\begin{remark}
\normalfont
{\bf Curse of dimensionality.}
The error bound \eqref{eq:approxBoundCor} consists of the constant $C_{H,\mathrm{ESN}}$ and three terms depending explicitly on $N$. The constant  $C_{H,\mathrm{ESN}}$ could depend on $N$ through the norms $\vertiii{{ A}^{\mathrm{ESN}} }$,$\|{\bf B}^{\mathrm{ESN}}\|$, $\vertiii{{ C}^{\mathrm{ESN}}}$, but for these norms it is reasonable to assume that they do not depend on $N$ (in practice, the norms of these matrices will be normalized). Similarly, it appears reasonable to assume that the norm $ \vertiii{K^{-\top}\Lambda}$ is bounded in $N$, since the norm $\vertiii{{ A}^{\mathrm{ESN}}}$ is likely to be bounded from below.
This argument indicates that in practically relevant situations, the constant does not depend on $N$ and, as it is apparent from the explicit expression \eqref{eq:c1tilde}, it depends only polynomially on the dimension $d$. To achieve an approximation error of size at most $\varepsilon$, we could thus take 
 $N = \left\lceil \max\left(d \log\left(\dfrac{\varepsilon}{3 C_{H,\mathrm{ESN}}}\right) \dfrac{1}{\log(\lambda)},d \log\left(\dfrac{\varepsilon}{3 C_{H,\mathrm{ESN}}}\right)\dfrac{1}{\log(\vertiii{{ A}^{\mathrm{ESN}} })},9 C_{H,\mathrm{ESN}}^2 \varepsilon^{-2} \right)\right\rceil $,
which grows only polynomially in $d$ and $\varepsilon^{-1}$. Hence, in these circumstances, there is no curse of dimensionality in the bound \eqref{eq:approxBoundCor}.
\end{remark}

\medskip

\begin{remark}
\normalfont
{\bf Implementation.} From a practical point of view, the bounds obtained in Theorem~\ref{thm:approx} and in Corollary \ref{cor:BoundForSome} apply to two different learning procedures: in the first case, when the chosen ELM weight distribution $\nu$ satisfies the absolute continuity assumption in Theorem~\ref{thm:approx}, only the ELM output layer needs to be trained. In the second case, when that condition is not available, all the ELM weights also need to be trained. However, note that these are not recurrent weights; consequently, the vanishing gradient problem does not occur here. In contrast, this problem sometimes occurs for standard recurrent neural networks in which all weights are trained by backpropagation through time.
\end{remark}

We conclude this section by considering another implementation scenario in which we impose the measure that we use for the sampling of the ELM, and show that it is still possible to sample using measures that are arbitrarily close to it in the sense of the $1$-Wasserstein metric in exchange for potentially increasing an error term that can be compensated by increasing the dimension of the ESN.
Let $\mathcal{W}_{1}$ be the $1$-Wasserstein metric. Suppose that we use the measure $\nu_0$ to generate the hidden layer weights. The next corollary shows how the error increases when we sample ``almost'' from the given measure $\nu_0$.

\begin{corollary}
\label{cor:boundClose}
Consider the same situation as in Corollary~\ref{cor:BoundForSome} and assume that $\nu_0$ has finite first moment and $\int_{\mathcal{X}} [|w|+\|{\bf a}\|_p+\|{\bf c}\| + |b|] \mu(dw,d{\bf a},d {\bf c},d b)<\infty$. Then, for any $\delta \in (0,1)$ there exists a probability measure $\nu$ with $\mathcal{W}_{1}(\nu_0,\nu) \leq \delta$ and a readout  ${\bf W}$ (an $\R^N$-valued random vector) such that the ESN-ELM with distribution $\nu$ for the hidden layer weights satisfies  
 for any ${\bf z} \in \mathcal{I}_d$ the approximation error bound
	\begin{equation}
\begin{aligned}
\label{eq:approxBoundCor2}
\E[|H({\bf z}) - \widehat{H}({\bf z})|^2]^{1/2} & \leq  \tilde{C}_{H,\mathrm{ESN}} \left[  \lambda^{\frac{N}{d}}  + \vertiii{{A}^{\mathrm{ESN}} }^\frac{N}{d}  + \frac{1}{N^{\frac{1}{2}}}\min(\delta^{-\frac{1}{2}},\mathcal{J}_{\nu_0}) \right]
\end{aligned}
\end{equation}	
with $\mathcal{J}_{\nu_0} = \left\|\frac{d \mu_{A,B,C}}{d \nu_0}\right\|_{\infty}^{\frac{1}{2}}$, $\tilde{C}_{H,\mathrm{ESN}} = C_{H,\mathrm{ESN}} \max(1,\mathcal{W}_{1}(\nu_0,\mu_{A,B,C}))^{\frac{1}{2}}$ and $C_{H,\mathrm{ESN}}$ as in \eqref{eq:CHESN}.
\end{corollary}
\begin{proof}	
First note that $ \vertiii{{ A}^{\mathrm{ESN}} }^T \leq \vertiii{{ A}^{\mathrm{ESN}} }^{\frac{N}{d}}$, since $T \geq \frac{N}{d}$. 
In the case where $ \mu_{A,B,C} \ll \nu_0 $ and $\frac{d \mu_{A,B,C}}{d \nu_0}$ is bounded, that is,  $\mathcal{J}_{\nu_0} < \infty$,  we may choose $\nu = \nu_0$ and obtain from Theorem~\ref{thm:approx} the bound
\begin{equation}
\begin{aligned}
\label{eq:approxBoundCor2aux}
\E[|H({\bf z}) - \widehat{H}({\bf z})|^2]^{1/2} & \leq  \tilde{C}_{H,\mathrm{ESN}} \left [  \lambda^{\frac{N}{d}}  + \vertiii{{ A}^{\mathrm{ESN}} }^\frac{N}{d}  + \frac{1}{N^{\frac{1}{2}}}\mathcal{J}_{\nu_0}\right].
\end{aligned}
\end{equation}	
We now derive an alternative bound (regardless of whether $\mathcal{J}_{\nu_0}$ is finite or not). Combining both bounds then  yields the bound \eqref{eq:approxBoundCor2}.  

From \eqref{eq:auxEq32} and the estimate $\|\pi(\bar{\mathcal{L}}({\bf a}))\| \leq \max(1,d^{\frac{1}{2}-\frac{1}{p}})c_0\|{\bf a}\|_p$ we obtain
\begin{equation} \label{eq:auxEq46} \begin{aligned}
\int_{\mathcal{X}^N} & \left(|w|+\|{\bf a}\|+\|{\bf c}\|+|b|\right) \mu_{A,B,C}(dw,d{\bf a},d {\bf c},d b) 
\\ & = \int_{\mathcal{X}} \left(|w|+\|K^{-\top}\Lambda\pi(\bar{\mathcal{L}}({\bf a}))\|+\|{\bf c}\|+|-(K^{-\top}\Lambda\pi(\bar{\mathcal{L}}({\bf a}))) \cdot \tilde{\bf B} +{\bf a} \cdot \bar{\bf B}_0+ b| \right)\mu(dw,d{\bf a},d {\bf c},d b)
\\ & \leq  \max(1,\vertiii{ K^{-\top}\Lambda}\max(1,d^{\frac{1}{2}-\frac{1}{p}})c_0)\\
& \times (1+\|\tilde{\bf B}\|+\| \bar{\bf B}_0\|_q)\int_{\mathcal{X}} \left(|w|+\|{\bf a}\|_p+\|{\bf c}\| + |b| \right)\mu(dw,d{\bf a},d {\bf c},d b) < \infty
\end{aligned}
\end{equation}
and thus $\mu_{A,B,C}$ has a finite first moment. Hence, $\tilde{\delta} = \delta [\max(1,\mathcal{W}_{1}(\nu_0,\mu_{A,B,C}))]^{-1} \in (0,1)$ and also $\nu = \tilde{\delta} \mu_{A,B,C} + (1-\tilde{\delta}) \nu_0$ has a finite first moment.
Therefore, we may use the Kantorovich-Rubinshtein duality (see \cite[Theorem~5.10; Particular Case 5.16]{villani2009optimal}) to calculate
\[
\begin{aligned}
\mathcal{W}_{1}(\nu_0,\nu) & = \sup_{f \colon f \text{ is $1$-Lipschitz}} \left\lbrace \int_{\mathcal{X}^N} f(x) (\nu_0-\nu)(dx) \right \rbrace 
\\ & = \tilde{\delta} \sup_{f \colon f \text{ is $1$-Lipschitz}} \left\lbrace \int_{\mathcal{X}^N} f(x) (\nu_0- \mu_{A,B,C})(dx) \right \rbrace 
 = \tilde{\delta} \mathcal{W}_{1}(\nu_0,\mu_{A,B,C}) \leq \delta.
\end{aligned}
\]
In addition, $ \mu_{A,B,C} \ll \nu $  and
\[
\tilde{\delta} \frac{d \mu_{A,B,C}}{d \nu} + (1-\tilde{\delta}) \frac{d \nu_0}{d \nu} = 1,
\]
hence  $\frac{d \nu_0}{d \nu} \geq 0$ implies 
 $\left\|\frac{d \mu_{A,B,C}}{d \nu}\right\|_{\infty} \leq \frac{1}{\tilde{\delta}}$. Therefore,  Theorem~\ref{thm:approx} yields \eqref{eq:approxBoundCor2}. 
\end{proof}

\subsection{Universality}
\label{Universality subsection}

As an additional result, we obtain the following universality result, which shows that any square-integrable functional (not necessarily in the recurrent Barron class) can be well approximated by the ESN-ELM \eqref{eq:ESN}-\eqref{eq:ELM} with suitably chosen readout ${\bf W}$ and hidden weights generated from a distribution $\nu$ arbitrarily close to a prescribed measure $\nu_0$.

\begin{corollary} 
Assume that the input set $D_d$ is bounded, and consider an arbitrary functional $H \colon \mathcal{I}_d \to \R$ that we will approximate with an ESN-ELM built with the following specifications: the activation functions are $\sigma_1(x)=x$ and either $\sigma_2(x)=\max(x,0)$ or $\sigma_2$ is a bounded, Lipschitz-continuous, and non-constant function. Furthermore, assume that there exists $\bar{c}>0$ and $\underline{l},\bar{l} \in (0,1)$, $\underline{l}<\bar{l}$, such that for any choice of $N \in \mathbb{N}$ the ESN parameters satisfy that the controllability matrix $K$ in \eqref{eq:Kdef} is invertible, $\underline{l}<\vertiii{{ A}^{\mathrm{ESN}} }<\bar{l}$, $\|{\bf B}^{\mathrm{ESN}}\|\leq \bar{c}$, $\vertiii{{ C}^{\mathrm{ESN}}} \leq \bar{c}$, and $\vertiii{K^{-1} \Lambda}  \leq \bar{c}$, where  
$\Lambda \in \R^{{N} \times {N}}$ is the diagonal matrix with entries $\Lambda_{d(k-1)+j,d(k-1)+j} = {\underline{l}}^{k-1}$, for all $k \in \N$, $j=1,\ldots,d$ with $d(k-1)+j \leq {N}$.
Let $\nu_0$ be a given hidden weight distribution with finite first moment. Then for any probability measure $\gamma$ on $\mathcal{I}_d \subset (D_d)^{\Z_-}$ such that $H \in L^{2}(\mathcal{I}_d,\gamma)$, there exists  
a probability measure $\nu$ with $\mathcal{W}_{1}(\nu_0,\nu) < \varepsilon$ and 
a readout  ${\bf W}$ (an $\R^N$-valued random vector) such that the ESN-ELM $\widehat{H}$ with readout ${\bf W}$ and distribution $\nu$ for the hidden layer weights satisfies  that
\[
\left(\int_{\mathcal{I}_d} \E[|H({\bf z}) - \widehat{H}({\bf z})|^2] \gamma(d{\bf z})\right)^{1/2} < \varepsilon. 
\]
\end{corollary}

\begin{proof} 	
Firstly, by Proposition~\ref{lem:dense} there exists $H^{RC} \in \mathcal{C} \cap L^{2}(\mathcal{I}_d,\gamma)$ such that $\|H-H^{RC}\|_{L^{2}(\mathcal{I}_d,\gamma)} < \varepsilon/2$.
From the proof of Proposition~\ref{lem:dense} and the construction in \cite{RC8} it follows that $H^{RC}$ is of the form 
\begin{equation} \label{eq:HRC} H^{RC}({\bf z}) = \int_{\mathcal{X}^{\bar{N}}} w \sigma_2({\bf a} \cdot (\pi_{\bar{N}} {\bf z}) + {\bf c} \cdot {\bf z}_0 +  b) \mu^{\bar{N}}(dw,d{\bf a},d {\bf c},d b), \quad  {\bf z} \in  (D_d)^{\Z_-},
\end{equation}
for some $\bar{N} \in \N$, an atomic probability measure $\mu^{\bar{N}}$, and with $\pi_{\bar{N}} \colon (D_d)^{\Z_-} \to \R^{\bar{N}}$ satisfying 
\begin{equation*}
(\pi_{\bar{N}}( {\bf z}))_{d(k-1)+j} =  (z_{-k})_j, \quad \mbox{for all $k \in \N$, $j=1,\ldots,d$ with $d(k-1)+j \leq \bar{N}$.}
\end{equation*}
Now let ${\bar{\lambda}}=\underline{l} \in (0,1)$, $p \in (1,\infty)$, ${\bf B} =0$ and define $A \colon \ell^q \to \ell^q$, $C \colon \R^d \to \ell^q$ by $(A {\bf x})_{i} = \mathbbm{1}_{\{i> d\}} {\bar{\lambda}} x_{i-d}$, $(C {\bf z})_{i} = \mathbbm{1}_{\{i\leq d\}} { z}_i$ for ${\bf x} \in \ell^q$, ${\bf z} \in \R^d$, $i \in \N$. Then, as in the proof of Proposition~\ref{lem:linear} we obtain that \eqref{eq:system} admits a unique solution $(\mathbf{x}_t)_{t \in \Z_-} \in \ell^\infty_-(\ell^q)$ and for $k\in \N$, $j=1,\ldots,d$ we have ${x}_{t,(k-1)d+j} = {\bar{\lambda}}^{k-1} { z}_{t-k+1,j}$. 
Let $\iota \colon \R^{\bar{N}} \to \ell^p$ be the natural injection $(x _1, \ldots, x_{\bar{N}})^{\top} \longmapsto (x _1, \ldots, x_{\bar{N}}, 0, \ldots)$ and let $\pi \colon \ell^q \to  \R^{\bar{N}}$ be the natural projection. Then 
\[{\bf a} \cdot (\pi_{\bar{N}} {\bf z}) = {\bf a} \cdot (\Lambda^{-1} \pi \mathbf{x}_{-1}) = (\iota \Lambda^{-1} {\bf a})\cdot \mathbf{x}_{-1}.
\]
Let $\phi \colon \mathcal{X}^{\bar{N}} \to \mathcal{X} $ be defined by $\phi(w,{\bf a},{\bf c},b)=(w,\iota \Lambda^{-1} {\bf a},{\bf c},b)$ and denote by $\mu = \mu^{\bar{N}} \circ \phi^{-1}$ the pushforward measure of $\mu^{\bar{N}}$ under $\phi$. 
Then 
\begin{equation} \label{eq:HRC2}
\begin{aligned} H^{RC}({\bf z}) & = \int_{\mathcal{X}^{\bar{N}}} w \sigma_2((\iota \Lambda^{-1} {\bf a})\cdot \mathbf{x}_{-1} + {\bf c} \cdot {\bf z}_0 +  b) \mu^{\bar{N}}(dw,d{\bf a},d {\bf c},d b)
\\ & = \int_{\mathcal{X}} w \sigma_2({\bf a}\cdot \mathbf{x}_{-1}({\bf z}) + {\bf c} \cdot {\bf z}_0 +  b) \mu(dw,d{\bf a},d {\bf c},d b).
\end{aligned}
\end{equation}
Thus,  $H^{RC} \in \mathcal{C}$ has a representation of the type \eqref{eq:Hrepres}  with  $\vertiii{A}={\bar{\lambda}} <1$. Furthermore, since $\mu^{\bar{N}}$ is atomic it follows directly that $I_{\mu,p}^{(2)} < \infty$ and $\int_{\mathcal{X}} [|w|+\|{\bf a}\|_p+\|{\bf c}\| + |b|] \mu(dw,d{\bf a},d {\bf c},d b)<\infty$.  

Next, fix  $\lambda \in (\vertiii{A},1)$ arbitrary and note that under the assumptions on the ESN, the constant $C_{H,\mathrm{ESN}}$ in \eqref{eq:CHESN} can be estimated as
\begin{equation} 
C_{H,\mathrm{ESN}} \leq \tilde{c}_1 \max(\frac{1}{(1-(\lambda^{-1}\bar{\lambda})^{p})^{\frac{1}{p}}},1) \max(I_{\mu,p},I_{\mu,p}^{(2)}) \left(1+2\frac{\bar{c}^2 }{1-\bar{l}} \right) =:\bar{C}_{H,0},
\end{equation}	
which does not depend on $N$.
Similarly, the bound
\[
\mathcal{W}_{1}(\nu_0,\mu_{A,B,C}) \leq \int_{\mathcal{X}^N} \|x\| \nu_0(dx)+ \int_{\mathcal{X}^N} \|x\| \mu_{A,B,C}(dx),
\]
the assumptions on the ESN matrices, and \eqref{eq:auxEq46} can be used to obtain an upper bound $\bar{C}_{H,1}$ on $\mathcal{W}_{1}(\nu_0,\mu_{A,B,C})$ that only depends on $\mu$, $d$, $p$, $\lambda$, $\bar{c}$, $\underline{l}$, $\bar{l}$ and $\nu_0$, but not on $N$. Set $\bar{c}_H =\bar{C}_{H,0} \max(1,\bar{C}_{H,1})^{\frac{1}{2}}$.

We now apply Corollary~\ref{cor:boundClose} to $H=H^{RC}$. We select $\delta  \in (0,1)$ with $\delta < \varepsilon$ and 
\[N = \left\lceil \max\left(d \log\left(\frac{\varepsilon}{6\bar{c}_H}\right)\dfrac{1}{\log(\lambda)},d \log\left(\frac{\varepsilon}{6\bar{c}_H}\right)\dfrac{1}{\log(\bar{l})} ,(6\bar{c}_H \varepsilon^{-1})^2\delta^{-1}\right) \right\rceil,\] 
then from Corollary~\ref{cor:boundClose} we  obtain that
there exists a probability measure $\nu$ with $\mathcal{W}_{1}(\nu_0,\nu) < \varepsilon$ and a readout  ${\bf W}$ (an $\R^N$-valued random vector) such that  the ESN-ELM $\widehat{H}$ with readout ${\bf W}$ and distribution $\nu$ for the hidden layer weights satisfies  
for any ${\bf z} \in \mathcal{I}_d$ 
\begin{equation}
\begin{aligned}
\label{eq:approxBoundCor2Proof}
\E[|H^{RC}({\bf z}) - \widehat{H}({\bf z})|^2]^{1/2} & \leq  \bar{c}_H \left [  \lambda^{\frac{N}{d}}  + \vertiii{{ A}^{\mathrm{ESN}} }^\frac{N}{d}  + \frac{1}{\delta^{\frac{1}{2}} N^{\frac{1}{2}}}\right] < \frac{\varepsilon}{2}.
\end{aligned}
\end{equation}	

 
Hence, 
\[
\left(\int_{\mathcal{I}_d} \E[|H({\bf z}) - \widehat{H}({\bf z})|^2] \gamma(d{\bf z})\right)^{1/2} \leq \|H-H^{RC}\|_{L^2(\mathcal{I}_d,\gamma)} + \left(\int_{\mathcal{I}_d} \E[|H^{RC}({\bf z}) - \widehat{H}({\bf z})|^2] \gamma(d{\bf z})\right)^{1/2} < \varepsilon. 
\]
\end{proof}

\subsection{Special case: static situation}
\label{Special case: static situation}

As a special case, we consider the situation without recurrence, that is, when $H \colon D_d \to \R$ is of the form 
\begin{equation} \label{eq:Hstatic} H({\bf u}) = \int_{\R \times \R^d \times \R} w \sigma_2({\bf c} \cdot {\bf u} +  b) \mu_0(dw,d {\bf c},d b), \quad  {\bf u} \in  D_d \subset \R^d,
\end{equation}
for a 
probability measure $\mu_0$ on $\R \times \R^d \times \R$ satisfying $\int_{\R \times \R^d \times \R} |w| (\|{\bf c}\|  +  |b|) \mu_0(dw,d {\bf c},d b) < \infty$.
$H$ is clearly a particular case of a recurrent generalized Barron functional, since it can be written as \eqref{eq:Hrepres} with measure $\mu$ on $\mathcal{X}$
 given by $\mu(dw,d {\bf a}, d {\bf c}, d b) = \delta_0(d{\bf a}) \mu_0(dw,d {\bf c},d b) $, which satisfies $I_{\mu,p} < \infty$ by the integrability assumption on $\mu_{0}$.
 
In this  situation, we  show that such static elements $H$ of the recurrent  Barron class can be approximated by ELMs, that is,  by 
feedforward neural networks 
\begin{equation}\label{eq:ELMstatic}
\widehat{H}({\bf u}) = \sum_{i=1}^N W_i \sigma_2({\bf c}^{(i)} \cdot {\bf u} + b_i)
\end{equation}
with randomly generated coefficients ${\bf c}^{(i)} $, $b_i$ valued in $\R^d$ and $\R$, respectively, and ${\bf W} \in \R^N$ trainable. 
$\widehat{H}$ can be viewed as a functional ${\bf z} \mapsto \widehat{H}({\bf z}_0)$, which is  of the form \eqref{eq:ELM} with $ {\bf a}^{(i)} = 0$. 
Let $w^{(1)}, \ldots, w^{(n)}$ be $\R$-valued random variables, assume that the random variables $(w^{(1)},{\bf c}^{(1)},b_1), \ldots,(w^{(N)},{\bf c}^{(N)},b_N)$  are IID and denote by $\nu_0$ their common distribution. 

\begin{corollary}
Suppose that  $H$ is as in \eqref{eq:Hstatic} with associated $\mu$ satisfying that $I_{\mu,p}^{(2)} < \infty$.
Assume the input set $D_d$ is bounded. Suppose that $ \mu_0 \ll \nu_0 $ and that ${\displaystyle \frac{d \mu_0}{d \nu_0}}$ is bounded.
	Then there exists a measurable function $f \colon (\R \times \R^d \times \R)^N \to \R^N$ such that the ELM $\widehat{H}$ with readout ${\bf W} = f((w^{(i)},{\bf c}^{(i)},b_i)_{i=1,\ldots,N}) \in \mathbb{R}^N$ satisfies for any ${\bf u} \in D_d$ the approximation error bound
	\begin{equation}
	\begin{aligned}
	\label{eq:approxBoundstatic}
	\E[|H({\bf u}) - \widehat{H}({\bf u})|^2]^{1/2} & \leq   \frac{C_{H}}{N^{\frac{1}{2}}}  
	\end{aligned}
	\end{equation}	
	with 
	\begin{equation} \label{eq:CH}
	C_{H} = (2 \max(2L_{\sigma_2}^2,|\sigma_2(0)|^2) \max(1,\sup_{{\bf v} \in D_d} \|{\bf v}\|^2))^{\frac{1}{2}} I_{\mu,p}^{(2)} \left\|\frac{d \mu_{0}}{d \nu_0}\right \|_{\infty}^{\frac{1}{2}}.
	\end{equation}
\end{corollary}

\begin{proof}
As noted above, $H$ is a recurrent generalized Barron functional with 
\[\mu(dw,d {\bf a}, d {\bf c}, d b) = \delta_0(d{\bf a}) \mu_0(dw,d {\bf c},d b)\] 
and we can choose the maps $A$, $C$ and the sequence ${\bf B}$ as $0$. With these choices, the map $\bar{\mathcal{L}}$ and the sequence $\bar{\bf B}_0$ appearing in the proof of Theorem~\ref{thm:approx} are both equal to $0$. Consequently, from \eqref{eq:auxEq32} we obtain that the measure $\tilde{\mu}^N$ in the proof of Theorem~\ref{thm:approx} satisfies 
\[
\int_{\mathcal{X}^N} f(w,{\bf a},{\bf c},b) \tilde{\mu}^N(dw,d{\bf a},d {\bf c},d b) = \int_{\R \times \R^d \times \R} f(w,0,{\bf c},b) \mu_0(dw,d {\bf c},d b)
\]
for any $f$ for which the integrand in the last line is integrable with respect to $\mu_0$. This shows that $\bar{H}^N$ in \eqref{eq:auxEq33} coincides with the map 
\begin{equation} \label{eq:auxEq34} \bar{H}^N({\bf z}) =  \int_{\R \times \R^d \times \R} w \sigma_2({\bf c} \cdot {\bf z}_0 +  b) \mu_0(dw,d {\bf c},d b) = H({\bf z}_0), \quad  {\bf z} \in  (D_d)^{\Z_-}
\end{equation} 
and furthermore $\bar{H}^N = H^N$. Therefore, denoting by $H$ also the functional ${\bf z} \mapsto H({\bf z}_0)$, in the first step of the error estimate \eqref{eq:auxEq35} only the last term is non-zero.

Furthermore, $
\psi_{A,B,C}(w,{\bf a},{\bf c},b) = (w,0,{\bf c},  b)$ and thus  
 $\mu_{A,B,C}(dw,d {\bf a}, d {\bf c}, d b) = \delta_0(d{\bf a}) \mu_0(dw,d {\bf c},d b) $. Therefore, the hypothesis $ \mu_0 \ll \nu_0 $  implies $ \mu_{A,B,C} \ll \delta_0(d{\bf a}) \nu_0 $. 
 
Choosing $W_i = {\displaystyle  \frac{w^{(i)}}{N}  \frac{d \mu_0}{d \nu_0}(w^{(i)},{\bf c}^{(i)},b_{i})}$, from \eqref{eq:auxEq29} we thus obtain
 \begin{equation} \label{eq:auxEq36} \begin{aligned}
 \E[|H({\bf u}) - \widehat{H}({\bf u})|^2]   \leq  N \E[W_1^2 \sigma_2({\bf c}^{(1)} \cdot {\bf u} + b_1)^2].
 \end{aligned}
 \end{equation}
The expectation in \eqref{eq:auxEq36} can be estimated by 
\begin{eqnarray*}
 \E[W_1^2 \sigma_2({\bf c}^{(1)} \cdot {\bf u} + b_1)^2]  & = & \int_{\R \times \R^d \times \R} \frac{w^2}{N^2}  \frac{d \mu_0}{d \nu_0}(w,{\bf c},b) \sigma_2({\bf c} \cdot {\bf u} + b)^2  \mu_0(dw,d {\bf c},d b)
\\ & \leq & 2 \left \|\frac{d \mu_{0}}{d \nu_0}\right\|_{\infty} \int_{\R \times \R^d \times \R} \frac{w^2}{N^2}  [L_{\sigma_2}^2|{\bf c} \cdot {\bf u} + b|^2 + |\sigma_2(0)|^2]  {\mu}_0(dw,d {\bf c},d b)
\\ & \leq &\frac{c_1}{N^2} \left \|\frac{d \mu_{0}}{d \nu_0}\right\|_{\infty} \int_{\R \times \R^d \times \R} w^2  [\|{\bf c}\|^2 + |b|^2 + 1]  {\mu}_0(dw,d {\bf c},d b)
\end{eqnarray*}
with $c_1 = 2 \max(2L_{\sigma_2}^2,|\sigma_2(0)|^2) \max(1,\sup_{{\bf v} \in D_d} \|{\bf v}\|^2)$. 
\end{proof}

\section{Learning from a single trajectory}

Consider now the situation in which we aim to learn a functional $H$ from a single trajectory of input/output pairs. We observe $n \in \N$ data points $({\bf z}_t,{\bf y}_t)$ for $t \in \{0,-1,\ldots,-(n-1)\}$ and aim to learn the input/output relation $H$ from them. In contrast to static situations, these data points are not IID; instead, they constitute a single realization of a discrete-time stochastic process. Similar problems were considered, for instance, in  \cite{pmlr-v178-ziemann22a}. 

More formally, we consider a stationary stochastic process $({\bf Z}_t,{\bf Y}_t)_{t \in\Z_-}$ and assume that $n$ sequential observations $({\bf Z}_t,{\bf Y}_t)$, $t \in \{0,-1,\ldots,-(n-1)\}$, coming from a single realization
of this process are available. Suppose that ${\bf Z}$ takes values in $\mathcal{I}_d$. Let $H \in \mathcal{C}$ be the unknown functional and assume that the input/output relation between the data is given as  
$H({\bf Z}) = \E[{\bf Y}_0 |{\bf Z}]$. 
For example, this is satisfied if ${\bf Z}$ is any stationary process and ${\bf Y}_t = H(\ldots,{\bf Z}_{t-2},{\bf Z}_{t-1},{\bf Z}_{t}) + \bm{\varepsilon}_t$ for a stationary process $(\bm{\varepsilon}_t)_{t \in \Z_-}$ independent of ${\bf Z}$ and with $\E[\bm{\varepsilon}_0] = 0$. 

The goal is to learn the functional $H$ from the observations $({\bf Z}_t,{\bf Y}_t)_{t=0,-1,\ldots,-n+1}$ of the underlying input/output process. Recall that $H({\bf Z}) = \E[{\bf Y}_0 |{\bf Z}]$ minimizes $\mathcal{R}(G)=\E[\|G({\bf Z})-{\bf Y}_0\|^2 ]$ over all measurable maps $G \colon \mathcal{I}_d \to \R^m$. To learn $H$ from the data, one thus aims to find a minimizer of 
\begin{equation}
\label{eq:empiricalRisk}
\mathcal{R}_n(G) = \frac{1}{n} \sum_{i=0}^{n-1} \|G({\bf Z}_{-i}^{-n+1})-{\bf Y}_{-i}\|^2,
\end{equation}
where we denote ${\bf Z}_{-i}^{-n+1} = (\ldots,0,0,{\bf Z}_{-n+1},\ldots,{\bf Z}_{-i-1},{\bf Z}_{-i})$. 

To learn $H$ from data, we use an approximant of the type $\widehat{H}$ introduced in \eqref{eq:ELM} and write $\widehat{H}_W = \widehat{H}$ to emphasize that $W$ are the trainable parameters. Note that $\widehat{H}_W $ is now $\R^m$-valued and is constructed by simply using a readout $W \in \mathbb{M}_{m,N}$ that collects $m$ readout vectors in a matrix or, equivalently, by making the $w^{(i)}$ in \eqref{eq:ELM} $\R^m$-valued. The results in Section~\ref{sec:approximation}, in particular the universality statement in Section \ref{Universality subsection}, indicate that $H$ can be approximated well by such systems. Hence, we expect that a minimizer of $\mathcal{R}_n(\cdot)$ over such systems should be a good approximation of $H$. Therefore, we set to solve the minimization problem 
\begin{equation}
\label{eq:ERM}
\widehat{W} =\underset{W \in \mathcal{W}_R}{\mbox{{\rm arg min}}}  \frac{1}{n} \sum_{i=0}^{n-1} \|\widehat{H}_W({\bf Z}_{-i}^{-n+1})-{\bf Y}_{-i}\|^2
\end{equation}
for $\mathcal{W}_R$ given by the set of all random matrices $W \colon \Omega \to \R^{m \times N}$ which satisfy for each row $W_{i,\cdot}$ the bound $\|W_{i,\cdot}\|\leq R$ for some regularization constant $R>0$ 
and which are measurable with respect to the data $({\bf Z}_t,{\bf Y}_t)_{t=0,-1,\ldots,-n+1}$ and the randomly generated parameters $(w_i,{\bf a}^{(i)},{\bf c}^{(i)},b_i)_{i=1,\ldots,N}$, ${A}^{\mathrm{ESN}}$, ${\bf B}^{\mathrm{ESN}}$, ${C}^{\mathrm{ESN}}$.
The latter requirement accounts for the fact that the randomly generated parameters are fixed after their initialization and hence the trainable weights may depend on them. 
The choice of $R$ will be specified later on in the statement of Theorem \ref{thm:learning}.

The problem \eqref{eq:ERM} admits an explicit solution (see, for instance, \cite[Section~4.3]{Gonon2021}). Once computed, the learned functional is $\widehat{H}_{\widehat{W}}$. The learning performance of the algorithm can now be evaluated by assessing its learning error (or generalization error)
\[
\E[\|H(\bar{\bf Z})-\widehat{H}_{\widehat{W}}(\bar{\bf Z})\|^2],
\]
where $\bar{\bf Z}$ is an IID\ copy of ${\bf Z}$ and $\bar{\bf Z}$  is independent of all other random variables introduced so far. An analysis of the learning error is carried out below in Theorem \ref{thm:learning}, in which we assume, inspired by the developments in \cite{RC10}, that the input and the output processes are of the type introduced in the following definition.

\begin{definition}
An $\R^k$-valued random process ${\bf U}$ is said to have a causal {\bfi  Bernoulli shift} structure, if there exist $q \in \N$, $G \colon (\R^q)^{\Z_-} \to \R^k$ measurable, and an IID\ collection $(\bm{\xi}_t)_{t \in \Z_-}$ of $\R^q$-valued random variables such that 
\[
{\bf U}_t = G(\ldots,\bm{\xi}_{t-1},\bm{\xi}_t), \quad t \in \Z_-. 
\]
The process ${\bf U} $ is said to have {\bfi geometric decay} if there exist $C_{\mathrm{dep}}>0$, $\lambda_{\mathrm{dep}} \in (0,1)$ such that the weak dependence coefficient $\theta(\tau):=\E[\|{\bf U}_0-\tilde{{\bf U}}^\tau_0\|]$ satisfies $\theta(\tau) \leq C_{\mathrm{dep}} \lambda_{\mathrm{dep}}^\tau$ for all $\tau \in \N$, where $\tilde{{\bf U}}^\tau_0 = G(\ldots,\tilde{\bm{\xi}}_{-\tau-1},\tilde{\bm{\xi}}_{-\tau},\bm{\xi}_{-\tau+1},\ldots,\bm{\xi}_{0})$ for $\tilde{\bm{\xi}}$ an independent copy of $\bm{\xi}$. 
\end{definition}

\begin{theorem} 
\label{thm:learning}
Assume $\sigma_1(x)=x$, $m=1$, $p \in (1,\infty)$, and that the input set $D_d$ is bounded. Consider a functional $H \in \mathcal{C}$ that has a representation of the type \eqref{eq:Hrepres}  with  ${\bf B} \in \ell^q$, bounded linear maps $C \colon \R^d \to \ell^q$, $A \colon \ell^q \to \ell^q$ satisfying $\vertiii{A}<1$ and $I_{\mu,p}^{(2)} < \infty$. Let $\lambda \in \left(\vertiii{A},1\right)$. Consider now as approximant an ESN-ELM system such that  $\vertiii{{ A}^{\mathrm{ESN}} }<1$ and that the matrix $K$ in \eqref{eq:Kdef} is invertible. Suppose that $w^{(1)}$ is bounded, $ \mu_{A,B,C} \ll \nu $, and $\frac{d \mu_{A,B,C}}{d \nu} $ is bounded by a constant $\kappa>0$. Let $R \geq \frac{\kappa }{\sqrt{N}} \|w^{(1)}\|_\infty$, $r \in (\vertiii{A^{\mathrm{ESN}}},1)$. Assume ${ Y}$ is bounded by a constant $\widetilde{M}$ and $({ Y},{\bf Z})$  has a causal Bernoulli shift structure with geometric decay and $\log(n)<n\log(\lambda_{max}^{-1})$, where $\lambda_{max} = \max( r,\lambda_{\mathrm{dep}})$.
Then the trained ESN-ELM $\widehat{H}_{\widehat{\bf W}}$ satisfies the learning error bound
\begin{equation} \label{eq:learningErrorBound}
\begin{aligned}
\E[|H(\bar{\bf Z})-\widehat{H}_{\widehat{\bf W}}(\bar{\bf Z})|^2]^{1/2}  \leq  C_{\mathrm{approx}} \left(  \lambda^{\frac{N}{d}}  + \vertiii{{A}^{\mathrm{ESN}} }^T  + \frac{1}{N^{\frac{1}{2}}} \right) + C_{\mathrm{est}} \left(     R N^{\frac{1}{2}}  \frac{\sqrt{\log(n)}}{\sqrt{n}} \right)^{\frac{1}{2}}
\end{aligned}
\end{equation}
with 
\begin{multline}
\label{eq:Capprox}
C_{\mathrm{approx}}  = \tilde{c}_1 \max(\kappa^{\frac{1}{2}},1) \max\left( \frac{\vertiii{C}}{\left(1-\vertiii{\lambda^{-1}{A}}^{p}\right)^{1/p}},\frac{\|{\bf B}\|_q}{1-\vertiii{A}},1\right) \\
\times  \max\left(I_{\mu,p},I_{\mu,p}^{(2)}\right)\cdot  \left(1+\frac{\vertiii{{C}^{\mathrm{ESN}}}  + \|{\bf B}^{\mathrm{ESN}}\|}{1-\vertiii{{A}^{\mathrm{ESN}} }} \vertiii{ K^{-\top}\Lambda} \right) , 
\end{multline}
\begin{multline}
\label{eq:Cest}
C_{\mathrm{est}}  = \tilde{c}_2 \Bigg[\frac{(\vertiii{{ C}^{\mathrm{ESN}}} +\|{\bf B}^{\mathrm{ESN}}\|+1)^3}{(1-\vertiii{{ A}^{\mathrm{ESN}} })^3 \sqrt{\log(\lambda_{max}^{-1})} }
\max(R,1)   (\E[\|{\bf a}^{(1)}\|^2]^{\frac{1}{2}} + \E[\|{\bf c}^{(1)}\|^2]^{\frac{1}{2}}  + \E[|b_1|^2]^{\frac{1}{2}}+1)  \\  
 \times  (r^2-\vertiii{{ A}^{\mathrm{ESN}}}^2)^{-1/2} \max([\kappa \|w^{(1)}\|_\infty]^{-1},1) \Bigg( \frac{ r}{1-r} +   \frac{1}{\lambda_{max}} + \frac{ (\vertiii{{ C}^{\mathrm{ESN}} }^2+1)^{\frac{1}{2}} \lambda_{max}^{-1}}{\log(\lambda_{max}^{-1})} \Bigg)  \Bigg]^{\frac{1}{2}},
\end{multline}
for $\tilde{c}_1$ only depending on $d$, $\sigma_2$, $p$, $\mathrm{diam}(D_d)$, and $\lambda$ (see \eqref{eq:c1tilde}) and $\tilde{c}_2$ only depending on 
$\sigma_2$, $\widetilde{M}$, $\mathrm{diam}(D_d)$, and $C_{\mathrm{dep}}$ (see \eqref{eq:ctilde2}). 
\end{theorem}
\begin{proof} 
Let $(\bar{\bf Z},\bar{{ Y}})$ be an IID\ copy of $({\bf Z},{ Y})$, independent of all other random variables introduced so far. Independence and the assumption $H({\bf Z}) = \E[{ Y}_0 |{\bf Z}]$ imply that 
$\E[\widehat{H}_{\widehat{\bf W}}(\bar{\bf Z})H(\bar{\bf Z})] = \E[\widehat{H}_{\widehat{\bf W}}(\bar{\bf Z})\bar{ Y}_0]$ and $\E[\widehat{H}_{\bf W}(\bar{\bf Z})H(\bar{\bf Z})] = \E[\widehat{H}_{{\bf W}}(\bar{\bf Z})\bar{ Y}_0]$ for any ${\bf W} \in \mathcal{W}_R$. Therefore, for any ${\bf W} \in \mathcal{W}_R$ we obtain
\begin{equation} \label{eq:auxEq37}
\begin{aligned}
\E[|H(\bar{\bf Z})-\widehat{H}_{\widehat{\bf W}}(\bar{\bf Z})|^2] & =
\E[|H(\bar{\bf Z})-\widehat{H}_{\bf W}(\bar{\bf Z})|^2] + \E[|\bar{ Y}_0-\widehat{H}_{\widehat{\bf W}}(\bar{\bf Z})|^2] - \E[|\bar{ Y}_0-\widehat{H}_{\bf W}(\bar{\bf Z})|^2]
\\ & \leq  \E[|H(\bar{\bf Z})-\widehat{H}_{\bf W}(\bar{\bf Z})|^2] + \E[\mathcal{R}(\widehat{H}_{\widehat{\bf W}})-\mathcal{R}_n(\widehat{H}_{\widehat{\bf W}})+\mathcal{R}_n(\widehat{H}_{\bf W})-\mathcal{R}(\widehat{H}_{\bf W})],
\end{aligned}
\end{equation}
where we used in the last step that \eqref{eq:ERM} implies $\mathcal{R}_n(\widehat{H}_{\widehat{\bf W}})\leq \mathcal{R}_n(\widehat{H}_{\bf W})$.

The first term in the right hand side of \eqref{eq:auxEq37} can be bounded using Theorem~\ref{thm:approx}. Indeed, Theorem~\ref{thm:approx} proves that there exists a measurable function $f \colon (\mathcal{X}^N)^N \to \R^N$ such that the ESN-ELM $\widehat{H}$ with readout ${\bf W} = f((w^{(i)},{\bf a}^{(i)},{\bf c}^{(i)},b_i)_{i=1,\ldots,N}) $ satisfies 
\begin{equation}
\begin{aligned}
\label{eq:approxBound2}
\E[|H(\bar{{\bf Z}}) - \widehat{H}_{\bf W}(\bar{{\bf Z}})|^2]^{1/2} & \leq  C_{H,\mathrm{ESN}} \left[  \lambda^{\frac{N}{d}}  + \vertiii{{ A}^{\mathrm{ESN}} }^T  + \frac{1}{N^{\frac{1}{2}}}  \left\|\frac{d \mu_{A,B,C}}{d \nu}\right\|_{\infty}^{\frac{1}{2}}\right]
\end{aligned}
\end{equation}	
with $C_{H,\mathrm{ESN}}$ given in \eqref{eq:CHESN}. Here we used independence and that $\bar{{\bf Z}}$ takes values in $\mathcal{I}_d$. Furthermore, in the proof of Theorem~\ref{thm:approx} $f$ and ${\bf W}$ are explicitly chosen as ${\displaystyle W_i = \frac{w^{(i)}}{N}  \frac{d  \mu_{A,B,C}}{d \nu}(w^{(i)},{\bf a}^{(i)},{\bf c}^{(i)},b_{i})}$. Therefore, $\P$-a.s.\ the random vector ${\bf W}$ satisfies $\|{\bf W}\| \leq \sqrt{N} \max_{i=1}^N \|W_i\|_\infty \leq \frac{\kappa }{\sqrt{N}} \|w^{(1)}\|_\infty \leq R$ and consequently ${\bf W} \in \mathcal{W}_R$.  

It remains to analyze the second term in the right hand side of \eqref{eq:auxEq37}. For notational simplicity we now consider $(w_i,{\bf a}^{(i)},{\bf c}^{(i)},b_i)_{i=1,\ldots,N}$, ${A}^{\mathrm{ESN}}$, ${\bf B}^{\mathrm{ESN}}$, ${C}^{\mathrm{ESN}}$ as fixed; formally this can be justified by performing the subsequent computations conditionally on these random variables as, for instance, in \cite[Proof of Theorem~4.3]{Gonon2021}. Then 
\begin{equation}
\label{eq:auxEq40}
\E[\mathcal{R}(\widehat{H}_{\widehat{\bf W}})-\mathcal{R}_n(\widehat{H}_{\widehat{\bf W}})+\mathcal{R}_n(\widehat{H}_{\bf W})-\mathcal{R}(\widehat{H}_{\bf W})] \leq 2 \E\left[\sup_{{\bf w} \in \R^N \,:\, \|{\bf w}\|\leq R}|\mathcal{R}_n(\widehat{H}_{\bf w})-\mathcal{R}(\widehat{H}_{\bf w})| \right].
\end{equation}
Let $\bar{\lambda} = (r^2-\vertiii{{ A}^{\mathrm{ESN}}}^2)^{1/2}$, 
denote $\mathcal{C}^{RC}= \{\widehat{H}_{\bf w} \,:\, {\bf w} \in \R^N, \|{\bf w}\|\leq R\}$ and note that each $\widehat{H}_{\bf w}$ can be written as $\widehat{H}_{\bf w} = h_{\bf w} \circ H^F$ for $H^F({\bf z}) = (\mathbf{x}_{0}^{\mathrm{ESN}}({\bf z}),{\bf z}_0,\bar{\lambda}\mathbf{x}_{-1}^{\mathrm{ESN}}({\bf z}))$, $h_{\bf w}(({\bf x}_0,{\bf z}_0,{\bf x}_{-1})) = \sum_{i=1}^N w_i \sigma_2(\bar{\lambda}^{-1}{\bf a}^{(i)} \cdot  \mathbf{x}_{-1} + {\bf c}^{(i)} \cdot {\bf z}_0 + b_i)$. Furthermore, $H^F \colon (D_d)^{\Z_-} \to \R^{N+d+N}$ is the reservoir functional associated to the map $F \colon \R^{N+d+N} \times \R^d \to \R^{N+d+N}$, $F(({\bf x}_0,{\bf x}_1,{\bf x}_2),{\bf z}) = (\sigma_1({ A}^{\mathrm{ESN}} \mathbf{x}_{0} + { C}^{\mathrm{ESN}} {\bf z} + {\bf B}^{\mathrm{ESN}}),{\bf z},\bar{\lambda}{\bf x}_0)$, which is just the reservoir map associated to the ESN determined by ${ A}^{\mathrm{ESN}}$, ${\bf B}^{\mathrm{ESN}}$, ${ C}^{\mathrm{ESN}}$ augmented by the current input and the (scaled) previous state. 

Then for each ${\bf w} \in \R^N$ with $\|{\bf w}\|\leq R$ we have 
\begin{multline*}
|h_{\bf w}(({\bf x}_0,{\bf z}_0,{\bf x}_{-1}))-h_{\bf w}((\bar{{\bf x}}_0,\bar{{\bf z}}_0,\bar{{\bf x}}_{-1}))|  \leq \sum_{i=1}^N |w_i| L_{\sigma_2} |\bar{\lambda}^{-1}{\bf a}^{(i)} \cdot  (\mathbf{x}_{-1}-\bar{{\bf x}}_{-1}) + {\bf c}^{(i)} \cdot ({\bf z}_0-\bar{{\bf z}}_0) |
\\  \leq L_{\sigma_2} R \max((\sum_{i=1}^N \|\bar{\lambda}^{-1}{\bf a}^{(i)} \|^2)^{1/2},(\sum_{i=1}^N \|{\bf c}^{(i)}\|^2)^{1/2}) [\|\mathbf{x}_{-1}-\bar{{\bf x}}_{-1}\| +  \|{\bf z}_0-\bar{{\bf z}}_0\|]
\leq \overline{L_h} \|({\bf x}_0,{\bf z}_0,{\bf x}_{-1})-(\bar{{\bf x}}_0,\bar{{\bf z}}_0,\bar{{\bf x}}_{-1})\|,
\end{multline*}
with $\overline{L_h} = 2 L_{\sigma_2} R \max((\sum_{i=1}^N \|\bar{\lambda}^{-1}{\bf a}^{(i)} \|^2)^{1/2},(\sum_{i=1}^N \|{\bf c}^{(i)}\|^2)^{1/2})$ and $|h_{\bf w}({\bf 0})| \leq L_{h,0}$ with $L_{h,0} =R  (\sum_{i=1}^N \sigma_2(b_i)^2)^{1/2}$.
Furthermore, for any ${\bf z} \in D_d$ the reservoir map $F(\cdot,{\bf z})$ is an $r$-contraction with $r = (\vertiii{{ A}^{\mathrm{ESN}} }^2 + \bar{\lambda}^2)^{1/2}<1$  and for any $({\bf x}_0,{\bf x}_1,{\bf x}_2) \in \R^{N} \times D_d \times \R^N$ and ${\bf z},\bar{{\bf z}} \in \R^d$ we have
\[
\|F(({\bf x}_0,{\bf x}_1,{\bf x}_2),{\bf z})-F(({\bf x}_0,{\bf x}_1,{\bf x}_2),\bar{{\bf z}})\| = \|({C}^{\mathrm{ESN}} ({\bf z}-\bar{{\bf z}}),{\bf z}-\bar{{\bf z}},0)\| = (\vertiii{{ C}^{\mathrm{ESN}} }^2 + 1)^{1/2} \|{\bf z}-\bar{{\bf z}}\|.
\]
In addition, for ${\bf z} \in (D_d)^{\Z_-}$ we have from \eqref{eq:auxEq18}
\[
\|H^F({\bf z})\| = (\|\mathbf{x}_{0}^{\mathrm{ESN}}({\bf z})\|^2+\|{\bf z}_0\|^2+\|\bar{\lambda}\mathbf{x}_{-1}^{\mathrm{ESN}}({\bf z})\|^2)^{1/2} \leq M_{\mathcal{F}},
\]
where $M_{\mathcal{F}} =  (2 [(1-\vertiii{{ A}^\mathrm{ESN}})^{-1}(\vertiii{{C}^{\mathrm{ESN}} } M +\|{\bf B}^{\mathrm{ESN}}\|)]^2+M^2)^{1/2}$ with  $M$ chosen such that $\|{\bf z}\| \leq M$ for all ${\bf z} \in D_d$. 
Finally, let $C_L = R M_{\mathcal{F}} + \widetilde{M} $ and note that 
$L(x,y) = \min(C_L,|x - y|)^2$
satisfies for all $H \in \mathcal{C}^{RC}$, ${\bf z} \in (D_d)^{\Z_-}$ and 
${ y} \in \R$ with $|{ y}| \leq \widetilde{M}$ 
that 
$L(H({\bf z}),{ y}) = |H({\bf z}) - { y}|^2$ 
and $z \mapsto [\min(C_L,|z|)]^2$ is Lipschitz-continuous with constant $2 C_L $. This can be seen by writing \[\begin{aligned}|\min(C_L,|z_1|)^2 - \min(C_L,|z_2|)^2| &  = |\min(C_L,|z_1|)-\min(C_L,|z_2|)| |\min(C_L,|z_1|)+\min(C_L,|z_2|)| 
\\ & \leq 2 C_L \big| |z_2|-|z_1|\big| \leq 2 C_L |z_2-z_1|. 
\end{aligned}\]

Consider the Rademacher-type complexity as defined in \cite{RC10}: let $\varepsilon_0,\varepsilon_1,\ldots$ be IID Rademacher random variables which are independent of the IID copies $\tilde{{\bf Z}}^{(l)}$, $l \in \N$,  of ${\bf Z}$. For each $k \in \N$ we define 
\[
{\mathfrak{R}}_k(\mathcal{C}^{RC}) = \frac{1}{k}\E\left[
\sup_{H \in \mathcal{C}^{RC}} \left| \sum_{l=0}^{k-1} \varepsilon_l H(\tilde{{\bf Z}}^{(l)})\right|
\right].
\]
Denote by $\mathbf{X}^l$ the vector with components ${X}^l_j = \sigma_2({\bf a}^{(j)} \cdot  \mathbf{x}_{-1}^{\mathrm{ESN}}(\tilde{\bf Z}^{(l)}) + {\bf c}^{(j)} \cdot \tilde{\bf Z}_{0}^{(l)} + b_j)$, $j=1,\ldots,N$. Then we can rewrite  $\widehat{H}_{{\bf w}}(\tilde{\bf Z}^{(l)})={\bf w}^\top \mathbf{X}^l$ and estimate using the Cauchy-Schwartz and Jensen's inequalities, independence, and the fact that $\E[ \varepsilon_l  \varepsilon_j] = \delta_{lj}$
\[ \begin{aligned}
{\mathfrak{R}}_k(\mathcal{C}^{RC}) & = \frac{1}{k}\E\left[
\sup_{{\bf w} \in \R^N \,:\, \|{\bf w}\|\leq R} \left| \sum_{l=0}^{k-1} \varepsilon_l \widehat{H}_{\bf w}(\tilde{{\bf Z}}^{(l)})\right|
\right] =  \frac{1}{k}\E\left[
\sup_{{\bf w} \in \R^N \,:\, \|{\bf w}\|\leq R} \left| {\bf w}^\top \sum_{l=0}^{k-1} \varepsilon_l  \mathbf{X}^l \right|
\right]
\\ & \leq  \frac{R}{k}\E\left[ \left\| \sum_{l=0}^{k-1} \varepsilon_l  \mathbf{X}^l \right\|
\right] \leq  \frac{R}{k}\E\left[ \left\| \sum_{l=0}^{k-1} \varepsilon_l  \mathbf{X}^l \right\|^2
\right]^{1/2} =  \frac{R}{k}\left(\E\left[  \sum_{l=0}^{k-1} \sum_{j=0}^{k-1} \varepsilon_l  \varepsilon_j (\mathbf{X}^l)^\top    \mathbf{X}^j  
\right]\right)^{1/2}
\\ & = \frac{R}{k}\left(\sum_{l=0}^{k-1} \E\left[   (\mathbf{X}^l)^\top    \mathbf{X}^l  
\right]\right)^{1/2} = \frac{R}{\sqrt{k}}\left( \E\left[   (\mathbf{X}^1)^\top    \mathbf{X}^1  
\right]\right)^{1/2}.
\end{aligned}
\]
Altogether, the hypotheses of \cite[Corollary~8(ii)]{RC10} are satisfied. Writing $\mathcal{R}_n^\infty(G) = \frac{1}{n} \sum_{i=0}^{n-1} |G({\bf Z}_{-i}^{-\infty})-{ Y}_{-i}|^2$ for the idealized empirical risk and applying this result proves that 
\begin{equation}\label{eq:auxEq39}
\begin{aligned}
\E\left[\sup_{{\bf w} \in \R^N \,:\, \|{\bf w}\|\leq R}|\mathcal{R}_n^\infty(\widehat{H}_{{\bf w}})-\mathcal{R}(\widehat{H}_{{\bf w}})| \right] \leq \frac{C_1}{n} + \frac{C_2 {\log(n)}}{{n}} + \frac{C_{3,abs} \sqrt{\log(n)}}{\sqrt{n}}
\end{aligned}
\end{equation}
 for all $n \in \mathbb{N}$ satisfying $\log(n)<n\log(\lambda_{max}^{-1})$, where $\lambda_{max} = \max( r ,\lambda_{\mathrm{dep}})$ and 
\begin{align} 
\label{eq:constantsdef} 
C_1 & = 2 C_L  
\frac{2 M_\mathcal{F}   \overline{L_h}  + C_{\mathrm{dep}}}{\lambda_{max}}, \quad\quad
C_{3, abs}  =  \frac{8 C_L 
}{\sqrt{\log(\lambda_{max}^{-1})} }\left[  
     R\left( \E\left[   (\mathbf{X}^1)^\top    \mathbf{X}^1  
\right]\right)^{1/2} +  \mathbb{E}\left[ | { Y}_{0} |^2_2 \right]^{1/2} \right],
\\ C_2  & = \frac{4 C_L  
	( M_\mathcal{F} \overline{L_h } +L_{h,0} + (\vertiii{{ C}^{\mathrm{ESN}}}^2  + 1)^{1/2}  \overline{L_h}  C_{\mathrm{dep}} \lambda_{max}^{-1}) + 2\mathbb{E}[|{ Y}_{0}|^2]}{\log(\lambda_{max}^{-1})}.
\label{eq:constantsdef3}  
\end{align}
Furthermore, \cite[Proposition~5]{RC10} yields that $\P$-a.s.
\[
\sup_{{\bf w} \in \R^N \,:\, \|{\bf w}\|\leq R} \left| \mathcal{R}_n(\widehat{H}_{{\bf w}})  - \mathcal{R}_n^\infty(\widehat{H}_{{\bf w}}) \right| \leq \frac{4 C_L  
	r  \overline{L_h} M_\mathcal{F} }{(1-r)n}.
\]
Combining this with \eqref{eq:auxEq39} we obtain
\begin{equation}
\label{eq:auxEq41}
\begin{aligned}
& \E\left[\sup_{{\bf w} \in \R^N \,:\, \|{\bf w}\|\leq R}|\mathcal{R}_n(\widehat{H}_{{\bf w}})-\mathcal{R}(\widehat{H}_{{\bf w}})| \right] \\ &  \leq \E\left[\sup_{{\bf w} \in \R^N \,:\, \|{\bf w}\|\leq R}|\mathcal{R}_n(\widehat{H}_{{\bf w}})-\mathcal{R}_n^\infty(\widehat{H}_{{\bf w}})| \right] + \E\left[\sup_{{\bf w} \in \R^N \,:\, \|{\bf w}\|\leq R}|\mathcal{R}_n^\infty(\widehat{H}_{{\bf w}})-\mathcal{R}(\widehat{H}_{{\bf w}})| \right]
\\ & \leq \frac{4 C_L  r  \overline{L_h} M_\mathcal{F} }{(1-r)n} + \frac{C_1}{n} + \frac{C_2 {\log(n)}}{{n}} + \frac{C_{3,abs} \sqrt{\log(n)}}{\sqrt{n}}
\\ & \leq  \max(R,1) 
\max(\overline{L_h},L_{h,0},1,R\left( \E\left[   (\mathbf{X}^1)^\top    \mathbf{X}^1  
\right]\right)^{1/2} ) C_{est,1}       \frac{\sqrt{\log(n)}}{\sqrt{n}},
\end{aligned}
\end{equation}
where 
\begin{equation}
\begin{aligned} 
C_{est,1} & = \frac{32 \max(1,C_{\mathrm{dep}})}{\sqrt{\log(\lambda_{max}^{-1})} } \max(2 \frac{(\vertiii{{C}^{\mathrm{ESN}}} M +\|{\bf B}^{\mathrm{ESN}}\|)^2}{(1-\vertiii{{A}^{\mathrm{ESN}}})^2}+M^2,1) \max(\widetilde{M},1) 
\\ & \quad \quad  \times \Bigg( \frac{ r }{(1-r)} +   \frac{1}{\lambda_{max}} + \frac{(2 + (\vertiii{{C}^{\mathrm{ESN}}}^2 + 1)^{1/2} \lambda_{max}^{-1}) + \mathbb{E}[|{ Y}_{0}|^2]}{\log(\lambda_{max}^{-1})} +  \mathbb{E}\left[ | { Y}_{0} |^2 \right]^{1/2} \Bigg).
\end{aligned}
\end{equation}
Recall that $(w^{(i)},{\bf a}^{(i)},{\bf c}^{(i)},b_i)_{i=1,\ldots,N}$ was considered fixed; now we take expectation also with respect to its distribution $\nu$ and obtain in the same way as we obtained \eqref{eq:auxEq41}
\begin{equation}
\label{eq:auxEq43}
\begin{aligned}
   \E\left[\sup_{{\bf w} \in \R^N \,:\, \|{\bf w}\| \leq R}|\mathcal{R}_n(\widehat{H}_{{\bf w}})-\mathcal{R}(\widehat{H}_{{\bf w}})| \right]  & \leq  \max(R,1) 
    \max\Bigg(2 R L_{\sigma_2} \E\left[\max\left((\sum_{i=1}^N \|\bar{\lambda}^{-1}{\bf a}^{(i)} \|^2)^{\frac{1}{2}},(\sum_{i=1}^N \|{\bf c}^{(i)}\|^2)^{\frac{1}{2}}\right)\right],  \\ & \quad  R\E\left[(\sum_{i=1}^N \sigma_2(b_i)^2)^{\frac{1}{2}}\right],1,R \left( \E\left[   (\mathbf{X}^1)^\top    \mathbf{X}^1  
\right]\right)^{\frac{1}{2}} \Bigg) C_{est,1}       \frac{\sqrt{\log(n)}}{\sqrt{n}}.
\end{aligned}
\end{equation}
The expectations in \eqref{eq:auxEq43} can be estimated using Jensen's inequality and \eqref{eq:auxEq18}:
\[\begin{aligned}
 \E\left[   (\mathbf{X}^1)^\top    \mathbf{X}^1  
\right]^{\frac{1}{2}}  = \left(\sum_{j=1}^N \E[ |{X}^1_j|^2 ]\right)^{\frac{1}{2}} & \leq \max(2L_{\sigma_2}^2,2|\sigma_2(0)|^2 )^{\frac{1}{2}} N^{\frac{1}{2}} \E[ |{\bf a}^{(1)} \cdot  \mathbf{x}_{-1}^{\mathrm{ESN}}({\bf Z}) + {\bf c}^{(1)} \cdot {\bf Z}_{0} + b_1|^2+1]^{\frac{1}{2}}
\\ & \leq N^{\frac{1}{2}} C_{est,2}
\end{aligned}
\]
with
\begin{equation}
\begin{aligned} 
C_{est,2}  =
&2^{\frac{3}{2}} \max(M,1) \max(L_{\sigma_2},|\sigma_2(0)|,1)\\
&\times  \frac{\vertiii{{C}^{\mathrm{ESN}}}  + \|{\bf B}^{\mathrm{ESN}}\|+1}{1-\vertiii{{A}^{\mathrm{ESN}}}}  (\E[\|\bar{\lambda}^{-1}{\bf a}^{(1)}\|^2]^{\frac{1}{2}} + \E[\|{\bf c}^{(1)}\|^2]^{\frac{1}{2}}  + \E[|b_1|^2]^{\frac{1}{2}}+1)
\end{aligned}
\end{equation}
and consequently
\begin{multline}
\label{eq:auxEq44}
  \max\Bigg(2R L_{\sigma_2} \E\left[\max\left((\sum_{i=1}^N \|\bar{\lambda}^{-1}{\bf a}^{(i)} \|^2)^{\frac{1}{2}},(\sum_{i=1}^N \|{\bf c}^{(i)}\|^2)^{\frac{1}{2}}\right)\right],  R\E\left[(\sum_{i=1}^N \sigma_2(b_i)^2)^{\frac{1}{2}}\right],1,R \left( \E\left[   (\mathbf{X}^1)^\top    \mathbf{X}^1  
\right]\right)^{\frac{1}{2}} \Bigg)
\\  \quad \leq \max\left(R N^{\frac{1}{2}} C_{est,2},1 \right).
\end{multline}
Inserting this in \eqref{eq:auxEq43} yields 
\begin{equation}
\label{eq:auxEq45}
\begin{aligned}
\E\left[\sup_{{\bf w} \in \R^N \,:\, \|{\bf w}\| \leq R}|\mathcal{R}_n(\widehat{H}_{{\bf w}})-\mathcal{R}(\widehat{H}_{{\bf w}})| \right]  & \leq  \max(R,1) 
 \max\left(R N^{\frac{1}{2}} C_{est,2},1 \right) C_{est,1}       \frac{\sqrt{\log(n)}}{\sqrt{n}}.
\end{aligned}
\end{equation}

Combining this with \eqref{eq:auxEq37}, \eqref{eq:approxBound2} and \eqref{eq:auxEq40} we thus obtain 
\begin{equation} \label{eq:auxEq42}
\begin{aligned}
\E[|H(\bar{\bf Z})-\widehat{H}_{\widehat{\bf W}}(\bar{\bf Z})|^2]^{1/2} &
 \leq  \E[|H(\bar{\bf Z})-\widehat{H}_{\bf W}(\bar{\bf Z})|^2]^{1/2} + \E[\mathcal{R}(\widehat{H}_{\widehat{\bf W}})-\mathcal{R}_n(\widehat{H}_{\widehat{\bf W}})+\mathcal{R}_n(\widehat{H}_{\bf W})-\mathcal{R}(\widehat{H}_{\bf W})]^{1/2}
\\ & \leq  C_{H,\mathrm{ESN}} [  \lambda^{\frac{N}{d}}  + \vertiii{{A}^{\mathrm{ESN}} }^T  + \frac{1}{N^{\frac{1}{2}}}  \left\|\frac{d \mu_{A,B,C}}{d \nu}\right\|_{\infty}^{\frac{1}{2}}] \\ & \quad + \sqrt{2} \left(   \max(R,1) 
 \max\left(R N^{\frac{1}{2}} C_{est,2},1 \right) C_{est,1}       \frac{\sqrt{\log(n)}}{\sqrt{n}} \right)^{\frac{1}{2}}
\\ & \leq  C_{\mathrm{approx}} [  \lambda^{\frac{N}{d}}  + \vertiii{{A}^{\mathrm{ESN}} }^T  + \frac{1}{N^{\frac{1}{2}}} ] + \tilde{C}_{\mathrm{est}} \left(     R N^{\frac{1}{2}}  \frac{\sqrt{\log(n)}}{\sqrt{n}} \right)^{\frac{1}{2}},
\end{aligned}
\end{equation}
where we used $\max(R N^{\frac{1}{2}},1)\leq R N^{\frac{1}{2}} \max([\kappa \|w^{(1)}\|_\infty]^{-1},1)$ and set  $C_{\mathrm{approx}} = C_{H,\mathrm{ESN}} \max(\kappa^{\frac{1}{2}},1)$, $\tilde{C}_{\mathrm{est}} = (2 C_{est,1} \max(R,1)C_{est,2}\max([\kappa \|w^{(1)}\|_\infty]^{-1},1)
)^{\frac{1}{2}}$.
Setting
\[\begin{aligned}
C_{\mathrm{est}} & = \tilde{c}_2 \Bigg[\frac{(\vertiii{{ C}^{\mathrm{ESN}}} +\|{\bf B}^{\mathrm{ESN}}\|+1)^3}{(1-\vertiii{{A}^{\mathrm{ESN}}})^3 \sqrt{\log(\lambda_{max}^{-1})} }
  \max(R,1)   (\E[\|{\bf a}^{(1)}\|^2]^{\frac{1}{2}} + \E[\|{\bf c}^{(1)}\|^2]^{\frac{1}{2}}  + \E[|b_1|^2]^{\frac{1}{2}}+1)  \\ & \quad \quad  \times (r^2-\vertiii{{ A}^{\mathrm{ESN}}}^2)^{-1/2}  \max([\kappa \|w^{(1)}\|_\infty]^{-1},1) \Bigg( \frac{ r}{1-r} +   \frac{1}{\lambda_{max}} + \frac{ (\vertiii{{ C}^{\mathrm{ESN}}} ^2+1)^{\frac{1}{2}} \lambda_{max}^{-1}}{\log(\lambda_{max}^{-1})} \Bigg)  \Bigg]^{\frac{1}{2}}
\end{aligned}\]
with 
\begin{equation}
\label{eq:ctilde2}
\tilde{c}_2 = 2^6( \max(1,C_{\mathrm{dep}}) \max(M,1)^5 \max(\widetilde{M},1)^3 
 \max(L_{\sigma_2},|\sigma_2(0)|,1))^{\frac{1}{2}} 
\end{equation}
and inserting \eqref{eq:CHESN} then shows \eqref{eq:learningErrorBound} with \eqref{eq:Capprox} and \eqref{eq:Cest}, as claimed.  
\end{proof}

\begin{remark}
The bound in Theorem~\ref{thm:learning} can be directly extended to the multivariate case $m>1$ as follows: if $H=(H_1,\ldots,H_m)$ and each $H_i$ satisfies the hypotheses formulated in Theorem~\ref{thm:learning}, then 
\begin{equation} \label{eq:learningErrorBoundmultivariate}
\begin{aligned}
\E[\|H(\bar{\bf Z})-\widehat{H}_{\widehat{W}}(\bar{\bf Z})\|^2]^{1/2}  & = \E\left[\sum_{i=1}^m |H_i(\bar{\bf Z})-\widehat{H}_{\widehat{W}_i}(\bar{\bf Z})|^2\right]^{1/2}
 \leq \sum_{i=1}^m \E\left[ |H_i(\bar{\bf Z})-\widehat{H}_{\widehat{W}_i}(\bar{\bf Z})|^2\right]^{1/2}
\\ & \leq \sum_{i=1}^m C_{\mathrm{approx},i} \left(  \lambda_i^{\frac{N}{d}}  + \vertiii{{A}^{\mathrm{ESN}} }^T  + \frac{1}{N^{\frac{1}{2}}} \right) + \sum_{i=1}^m C_{\mathrm{est},i} \left(     R N^{\frac{1}{2}}  \frac{\sqrt{\log(n)}}{\sqrt{n}} \right)^{\frac{1}{2}}
\end{aligned}
\end{equation}
with $C_{\mathrm{approx},i}$, $C_{\mathrm{est},i}$ the constants from Theorem~\ref{thm:learning} corresponding to $H_i$. 
\end{remark}

\section{Conclusions}

In this paper we have provided reservoir computing approximation and generalization bounds  for a new concept class of input/output systems that extends to a dynamical context the so-called generalized Barron functionals. We call the elements of the new class recurrent generalized Barron functionals. Its elements are characterized by the availability of readouts with a certain integral representation built on infinite-dimensional state-space systems. We have also shown that they form a vector space and that the class contains as a subset systems of this type with finite-dimensional state-space representations. This class has been shown to be very rich: it contains all sufficiently smooth finite-memory functionals, linear and convolutional filters, as well as any input-output system built using separable Hilbert state spaces with readouts of integral Barron type. Moreover, this class is dense in the $L^2$-sense.

The reservoir architectures used for the approximation/estimation of elements in the new class are randomly generated echo state networks with either linear (for approximation) or ReLU (estimation) activation functions, and with readouts built using randomly generated neural networks, in which only the output layer is trained (extreme learning machines or random feature neural networks). The results in the paper yield a fully implementable recurrent neural network-based learning algorithm with provable convergence guarantees that does not suffer from the curse of dimensionality.

\medskip

\addcontentsline{toc}{section}{Acknowledgments}
\noindent {\bf Acknowledgments.} The authors acknowledge partial financial support coming from the Swiss National Science Foundation (grant number 200021\_175801/1). L. Grigoryeva and J.-P. Ortega wish to thank the hospitality of the University of Munich and L. Gonon that of the Nanyang Technological University during the academic visits in which some of this work was developed.  We thank Christa Cuchiero and Josef Teichmann for interesting exchanges that inspired some of the results in this paper.

\addcontentsline{toc}{section}{Bibliography}

\end{document}